\documentclass[letterpaper]{article}

\usepackage{PRIMEarxiv}

\PassOptionsToPackage{hyphens}{url}

\usepackage[
backend=biber,
citestyle=authoryear,
style=authoryear,
sorting=nyt,
maxbibnames=99,
natbib
]{biblatex} 

\renewbibmacro{in:}{} 

\DeclareFieldFormat{pages}{#1} 

\DeclareFieldFormat{journaltitle}{\mkbibemph{#1}\isdot}

\renewbibmacro*{journal+issuetitle}{
  \usebibmacro{journal}%
  \setunit*{\addcomma\space}%
  \iffieldundef{series}
    {}
    {\newunit
     \printfield{series}%
     \setunit{\addspace}}%
  \usebibmacro{volume+number+eid}%
  \setunit{\addspace}%
  \usebibmacro{issue+date}%
  \setunit{\addcolon\space}%
  \usebibmacro{issue}%
  \newunit}

\renewbibmacro*{volume+number+eid}{%
  \printfield{volume}%
  \setunit*{\addnbspace}
  \printfield{number}%
  \setunit{\addcomma\space}%
  \printfield{eid}}

\DeclareFieldFormat[article]{number}{\mkbibparens{#1} }

\date{}            

\usepackage{dcolumn} 
\newcolumntype{d}[1]{D{.}{.}{#1}}
\usepackage{siunitx} 
\usepackage{calc,setspace,amsmath,pictex,amssymb,hhline,warpcol,rotating}
\providecommand{\mathbbm}[1]{\mathbb{#1}}
\usepackage{subcaption}
\usepackage{algorithm}
\usepackage{mathtools}
\usepackage{algpseudocode}
\usepackage{pifont}
\usepackage{graphicx}
\usepackage[table]{xcolor}
\usepackage{threeparttable}
\usepackage{colortbl}
\usepackage[hyphens]{url}
\usepackage{endnotes}			
\usepackage{tabularx}
\usepackage{array} 
\usepackage{multicol}
\usepackage{multirow}
\usepackage[justification=centering]{caption}
\usepackage[toc,page]{appendix}
\usepackage{float}
\usepackage{hyperref}
\usepackage{hypcap} 
\usepackage{longtable} 
\usepackage{dsfont}

\usepackage{booktabs}
\usepackage{makecell}
\usepackage{etoolbox}
\usepackage{footnote}
\usepackage{pdflscape}
\usepackage{afterpage}
\usepackage{adjustbox}
\usepackage{enumitem}
\usepackage{capt-of}

\hypersetup{
  hidelinks,
  hypertexnames=false,
  pdftitle={Forecasting Revenue with Its Customer-Base Drivers: When and Why Coordination Helps},
  pdfauthor={Kyeongbin Kim, Daniel M. McCarthy, and Dokyun Lee},
  pdfsubject={Machine learning for coordinated customer-base and revenue forecasting}
}

\makeatletter
\def\BState{\State\hskip-\ALG@thistlm}
\makeatother

\let\footnote=\endnote

\makeatletter
\@fpsep\textheight
\makeatother

\title{Forecasting Revenue with Its Customer-Base Drivers: When and Why Coordination Helps%
\thanks{Author ordering reflects contributions. The data used in this study were provided by Consumer Edge.}}
\author{%
  \textbf{Kyeongbin Kim} \\
  University of Wisconsin--Madison \\
  \texttt{kyeongbin.kim@wisc.edu}
  \And
  \textbf{Daniel M. McCarthy} \\
  University of Maryland \\
  \texttt{dmccar@umd.edu}
  \And
  \textbf{Dokyun Lee} \\
  Boston University \\
  \texttt{dokyun@bu.edu}
}

\begin{document}

\maketitle

\begin{abstract} 
Revenue forecasts guide acquisition budgets, demand planning, and customer-based valuations, yet an aggregate forecast does not show whether change reflects acquisition, repeat purchasing, spending per order, or offsetting movements. Using weekly transaction panels for 966 companies in 25 industries, the authors develop the Customer-Based Multi-task Transformer (CBMT), which learns shared structure, retains separate primitive forecasts, and aligns their combination with downstream revenue. CBMT's mean total-sales error is 30\% below the strongest representative established customer-base benchmark. It is also 2.65\% below a Transformer that forecasts total sales directly, although the paired difference is not statistically significant ($p=.222$), and it beats separately estimated single-task forecasts for 74.3\% of firms. CBMT's source MAE is lower in 23 of 24 benchmark-by-outcome comparisons, with the remaining difference not statistically distinguishable from zero. Firms whose primitives co-move more strongly are more likely to benefit from joint forecasting; selected-family scenario-3 comparisons are consistent with gains from shared representation and revenue alignment but remain diagnostic rather than causal. Accuracy deteriorates for all models when customer-base dynamics are highly volatile, and CBMT's advantage narrows there. Calibration-period routing rules do not improve average accuracy over always deploying CBMT. The results show how coordinated customer-base forecasts support revenue planning and when they warrant greater caution.
\end{abstract}



\section{Introduction}

Revenue forecasts carry considerable weight inside firms. They set acquisition budgets, shape retention and demand planning, anchor internal guidance, and feed customer-based valuations. Despite all that influence, an aggregate forecast says little about why revenue is expected to change. A projected shortfall may mean that the firm is acquiring fewer customers, that existing customers are ordering less often, that order values are shrinking, or that several of these forces are at work together. These explanations carry different implications for the future. Fewer new customers shrink the base available to generate revenue in later periods, weaker repeat purchasing suggests waning interest in the product among cohorts the firm has already acquired, and softer order values often reflect pricing and promotion choices that remain at least partly within the firm's control. The same projected sales path can therefore arise from very different customer-base states, and each points managers toward different actions. That makes the components themselves---customer acquisition, repeat orders per customer (ROPC), and average order value (AOV)---worth forecasting in their own right. We refer to these three components as customer-base primitives.

Recent episodes at two prominent subscription businesses illustrate why the source of a revenue change matters. After Netflix reported subscriber losses in April 2022, management introduced a lower-priced ad-supported tier to bring in new members and moved to convert password-sharing households into paying ones \citep{reuters2022nflx351,reuters2022nflx_response}. Peloton, facing slowing growth and weaker guidance, cut hardware prices and expanded distribution to support acquisition while raising subscription prices to earn more from its installed base \citep{reuters2024peloton_guidance,verge2022peloton_pricing}. Both firms used different customer-base levers, supporting acquisition where growth was weak while generating more revenue from existing customer relationships that appeared durable.

These primitives are usually forecast separately, with each series modeled on its own history. That practice can discard useful information. Promotions, product launches, seasonal demand, pricing changes, and competitive shocks rarely touch a single primitive in isolation; they tend to move acquisition, repeat purchasing, and order values together or with short lags. When such shared drivers are present, the recent path of one primitive carries news about the conditions shaping the others, and a model that forecasts the primitives jointly can put that information to use. Joint forecasting is not guaranteed to help, however. When one primitive is dominated by idiosyncratic noise, or when customer-base dynamics are unusually volatile, pooling can transmit noise instead of signal. The question we study is therefore conditional: when does jointly forecasting customer-base primitives improve revenue projections, and why?

\begin{figure}[!htbp]
    \centering
    \caption{Decomposing Revenue into Customer Behavioral Components (Wayfair)}
    \includegraphics[width=\textwidth]{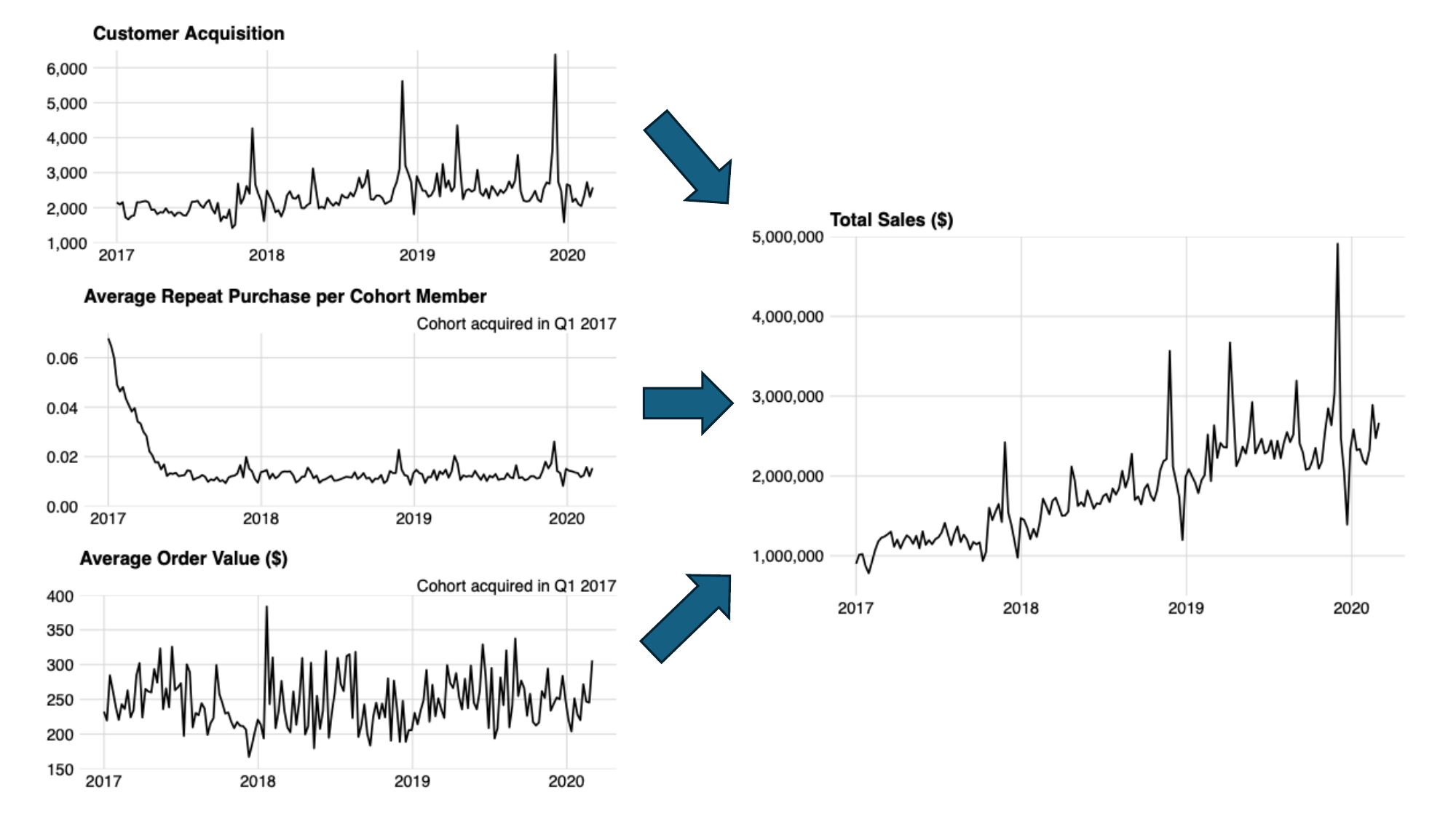}
    \label{fig:substantive_goal}
    \begin{minipage}{\textwidth}
        \footnotesize \textit{Note:} Customer acquisitions (top-left) represent new customers acquired each week. Each acquisition cohort generates repeat orders over calendar time (center-left), with associated average order values (bottom-left). The example cohort displayed in this figure consists of customers acquired in Q1 2017. The combination of these three customer-base primitives determines overall company-level weekly sales (right).
    \end{minipage}
\end{figure}

We study this question using weekly cohort-level transaction panels covering 966 companies in 25 industries. Cohorts offer a practical middle ground between individual transaction histories, which are sparse because most customers purchase infrequently, and firm-level aggregates, which obscure the behavior that generates revenue. They are also a unit managers already work with when they ask whether acquisition spending is attracting valuable customers and whether recent cohorts are repeating and spending as expected. Figure~\ref{fig:substantive_goal} illustrates the resulting decomposition for Wayfair: customers enter through weekly acquisition cohorts, those cohorts generate repeat orders at an associated average order value, and the cohort contributions sum to weekly company sales. Forecasting at this level delivers more than a sales number: the model projects the acquisition, repeat purchasing, and spending behaviors that generate the sales path, so managers can see not only how much revenue is expected to change but where that change is expected to come from.

To examine when and why joint forecasting pays off, we develop the Customer-Based Multi-task Transformer (CBMT), a cohort-level forecasting framework with two main ingredients. The model learns a shared representation from the joint history of acquisition, ROPC, and AOV, so that signals appearing in one primitive can sharpen forecasts of the others, while separate prediction layers preserve each primitive's distinct dynamics. During training, an auxiliary revenue-alignment regularizer penalizes combinations of primitive forecasts that imply implausible downstream revenue, holding the components accountable to the sales they generate. We implement the shared layer with a Transformer, but the framework's substance is joint learning with primitive-specific predictions; the section on the joint customer-base prediction framework presents the details.

The empirical analyses yield three main results. First, CBMT is consistently competitive across the three customer-base primitives, and its advantage is largest where coordination matters most. Relative to the strongest representative established customer-base benchmark, it reduces total-sales forecasting error by approximately 30\%. Relative to a tuned Transformer trained on aggregate-sales history to predict aggregate sales alone---a benchmark that can devote all of its capacity to the topline because it is never asked to explain where sales come from---its mean total-sales SMAPE is 2.65\% lower, although the paired difference is not statistically significant ($p=.222$). It has lower total-sales SMAPE than separately estimated single-task forecasts for 718 of 966 companies (74.3\%). The decomposition itself is also forecast accurately. When we break each firm's realized change in sales into the customer behaviors that produced it, CBMT has lower mean absolute error in 23 of the 24 benchmark-by-outcome comparisons; in the remaining comparison, the benchmark's 0.47-point edge is not statistically distinguishable from zero. Second, the gains line up with the shared-driver account. Firms whose primitives co-move more strongly during the calibration period are more likely to benefit from joint forecasting in the holdout period. A scenario-3 comparison of selected model families is also consistent with downstream gains from shared representation and revenue alignment, but it is diagnostic rather than a causal ablation. Third, the benefits have boundary conditions. Forecast accuracy deteriorates in high-volatility settings for every model we examine, and CBMT's edge over flexible alternatives narrows there, a pattern managers can anticipate using observable calibration-period characteristics.

This research makes three contributions. First, we provide evidence on why jointly forecasting customer-base primitives improves revenue projections. The three primitives often serve as multiple noisy readouts of the same business conditions, and joint-model wins concentrate among firms whose primitives share more temporal structure. Second, we show how to forecast these primitives jointly. The cohort-level accounting that links acquisition, repeat purchasing, and spending to revenue is well established in customer-based valuation research; CBMT adds shared learning across the three primitives and a revenue-alignment regularizer that disciplines their combination, while preserving separate forecasts that managers can translate into the sources of a projected sales change: repeat-order volume from the existing customer base, repeat spend per order, and net customer-base replenishment. Third, we provide large-scale evidence on where joint forecasting is most valuable and where customer-base forecasting remains difficult. The forecast-difficulty estimates turn that evidence into something managers can act on. Smaller and more volatile customer bases warrant wider uncertainty bands, more frequent updating, and more conservative guidance. Together, the results position joint customer-base forecasting as a diagnostic tool for revenue planning, customer-based valuation, and marketing resource allocation.

The remainder of the paper proceeds as follows. We first review related research on customer-based valuation, customer-behavior forecasting, and joint modeling of customer outcomes. We then describe the data, develop the CBMT framework, and present the forecasting results, including how accurately each model forecasts the sources of a sales change, followed by diagnostic evidence on why joint forecasting helps and analyses of when forecasts warrant added caution. We close by discussing implications for marketing planning, revenue guidance, and customer-based valuation.

\section{Related Literature} \label{sec:related-literature}

Our research builds on several connected streams that link customer behavior to aggregate revenue and firm value. Prior research establishes that acquisition, repeat purchasing, and spending are economically meaningful drivers of customer-base value and that these processes need not evolve independently. We organize the literature around a related but less studied question: when is it valuable to forecast these customer-base primitives jointly, rather than separately, for revenue projection and managerial diagnosis? We first review the customer-to-firm-value literature, then summarize work modeling individual customer-base primitives and their interdependencies. We conclude by drawing on multi-task learning research to motivate the conditions under which pooling information across primitives should be more or less useful.

The recognition that customer relationships constitute valuable firm assets has deep roots in marketing strategy \citep{blattberg1996manage,rust2004return}. \citet{gupta2004valuing} showed how customer lifetime value calculations can be used to approximate corporate valuation. \citet{libai2009diffusion} extend diffusion models to services by incorporating company- and category-level attrition, showing that neglecting these dynamics biases growth estimates. A collection of subsequent papers formalized the link to corporate valuation: \citet{schulze2012linking} propose a framework connecting customer equity to shareholder value with capital structure considerations, while \citet{mccarthy2017valuing,mccarthy2018customer} show how models of acquisition, retention, and spending can generate period-by-period revenue forecasts for use in standard discounted cash-flow models. Complementary perspectives have emerged from finance, modeling customer capital as an intangible asset that drives firm value (\citealt{gourio2014customer}) and providing practical frameworks for valuing users and subscribers (\citealt{damodaran2018going}), while accounting research documents the value relevance of the customer base and their financial reporting implications (\citealt{bonacchi2015customer}).

Research on forecasting customer acquisition has deep roots in the diffusion literature \citep{bass1969new}. Important extensions of this work incorporate generational and substitution dynamics \citep{norton1987diffusion}, and marketing-mix variables \citep{bass1994gbm}. See \citet{peres2010innovation} for a review of this literature. This work emphasizes the dynamic nature of customer adoption, typically employing parametric diffusion models to forecast those dynamics. 

An extensive literature has developed around the problem of predicting repeat purchases. The seminal Pareto/NBD model (Schmittlein, Morrison, and Colombo 1987) introduced a framework for businesses characterized by latent attrition; the Beta-Geometric/Negative Binomial Distribution (BG/NBD, \citealt{fader2005rfm}) offers a more parsimonious alternative with comparable accuracy. \citet{platzer2016ticking} extend the recency-frequency paradigm by incorporating regularity in interpurchase timing through the Pareto/GGG model. Other work incorporates covariates to improve fit and provide additional managerial insight \citep[e.g.,][]{schweidel2013incorporating,braun2015transaction,bachmann2021role}. While these parametric models perform well, they impose distributional assumptions on interpurchase timing and dropout (e.g., gamma and beta--geometric mixtures) that can be restrictive when customer acquisition and repeat purchasing dynamics are irregular and/or non-stationary. More flexible approaches relax these assumptions: \citet{dew2018bayesian} use Gaussian process priors to capture nonlinear temporal structure in purchase processes, and \citet{valendin2022customer} employ a long short-term memory (LSTM) model (\citealt{hochreiter1997long}) to model temporal dependencies from transactions directly with few distributional assumptions. Recent work by \citet{lu2025express} demonstrates the usefulness of transformer architectures for predicting customer touchpoints and conversion probabilities in multichannel marketing settings. We treat the Transformer as a useful forecasting architecture rather than as the substantive contribution of this research.

In contrast to the extensive literature on repeat purchasing behavior, the modeling of customer spending conditional on purchase has received comparatively less attention. The amount that is spent when purchases are made is often modeled with a gamma--gamma specification, with individual spending decisions modeled as draws from a gamma distribution, with cross-sectional heterogeneity accounted for through another gamma distribution \citep[]{fader2005rfm}, given its interpretability and the ease with which it can be used alongside the aforementioned repeat purchasing models.

Taken together, these literatures provide increasingly rich approaches for modeling individual customer-base primitives. Their primary objective, however, is generally to model a focal behavioral process accurately or to link separately estimated components to customer value. They provide less systematic guidance on when information contained in one primitive should improve forecasts of the others, when joint forecasting may transmit noise, and how primitive-level forecasts should be disciplined by their downstream revenue implications.

A separate literature stream explicitly recognizes that customer behaviors may be interdependent. \citet{schweidel2008bivariate} develop a bivariate timing model for acquisition and churn in subscription settings, showing that correlation between processes materially affects customer-base projections. \citet{ascarza2013joint} jointly model usage and retention. Related state-space and hidden Markov model (HMM) formulations infer latent relationship states that simultaneously affect purchase incidence and intensity \citep[e.g.,][]{netzer2008hidden}, while other bivariate and hierarchical models treat purchasing and spending together \citep[e.g.,][]{singh2009bayesian}. This work establishes that jointly modeling customer behaviors can be substantively important. At the same time, the empirical settings and modeled relationships are often deliberately narrow: a specific pair of outcomes, a particular business context, or a limited set of cohorts. This leaves open a broader forecasting question. Across heterogeneous firms, when does pooling information across acquisition, repeat purchasing, and spending improve downstream revenue projections, and when does it provide limited benefit?

Research on multi-task learning provides conceptual guidance for this question. Multi-task approaches seek to exploit shared information across related tasks while retaining task-specific structure \citep[e.g.,][]{ruder2017overview}. Pooling is not automatically beneficial. Joint optimization can be harmed when tasks interfere with one another \citep[e.g.,][]{yu2020gradient}. Applied to customer-base analysis, this logic suggests that the value of joint forecasting should vary systematically across firms. Gains should be larger when acquisition, repeat purchasing, and spending exhibit shared temporal structure and more limited when the primitives are weakly related, unusually noisy, or volatile. These are empirical propositions rather than assumptions built into the framework.

The same logic makes decomposition granularity a context-dependent design choice rather than a universal optimum. A more granular decomposition can provide a more detailed diagnosis when its components correspond to distinct managerial levers and can be estimated with sufficiently stable data. For example, repeat purchasing could be decomposed further into the share of customers who remain active and orders per active customer. Additional tasks, however, increase sparsity and create more opportunities for weakly related or noisy signals to enter a shared representation. We therefore use acquisition, ROPC, and AOV as an intermediate-granularity decomposition. The three primitives map directly into revenue, remain interpretable for managers, and can be measured consistently across the heterogeneous firms in our data. Evaluating finer decompositions remains an opportunity for future research.

The literatures above supply two of the building blocks for our study. Customer-based valuation research establishes the cohort-level accounting that links acquisition, repeat purchasing, and spending to revenue, and joint-modeling research establishes that customer behaviors can be interdependent. We add the joint forecasting method itself, a framework that learns a shared representation across the three primitives and an auxiliary revenue-alignment regularizer that discourages combinations of primitive forecasts implying implausible downstream revenue, while retaining the primitive-specific predictions that identify the sources of a projected revenue change.

We also add evidence on when joint forecasting helps. Multi-task learning research implies that pooling should help more in some settings than in others, and transaction panels covering 966 companies in 25 industries allow us to examine this variation directly: where joint forecasting delivers larger gains, and where customer-base forecasting is harder for every model we consider. Web Appendix~\ref{wa:literature_comparison} provides a descriptive comparison of selected studies in these literatures.

\section{Data} \label{sec:data}

Our dataset is cohort-level weekly transactional data provided by Consumer Edge, a leading data analytics company that has access to de-identified credit and debit card transactions for a static panel of approximately 2.3 million panel members across the United States from January 2016 to February 2020 (the `observation period'), covering 966 companies that began commercial operations at or before the start of the observation period, and did not cease operations before the end of the observation period. We have full transaction records for these panel members across these companies over the observation period.

\begin{figure}[!t]
\centering
    \caption{Active Customers, Sales, and Average Order Value Across 966 Companies \label{fig:earnest_visualize}}
    \includegraphics[width=\textwidth]{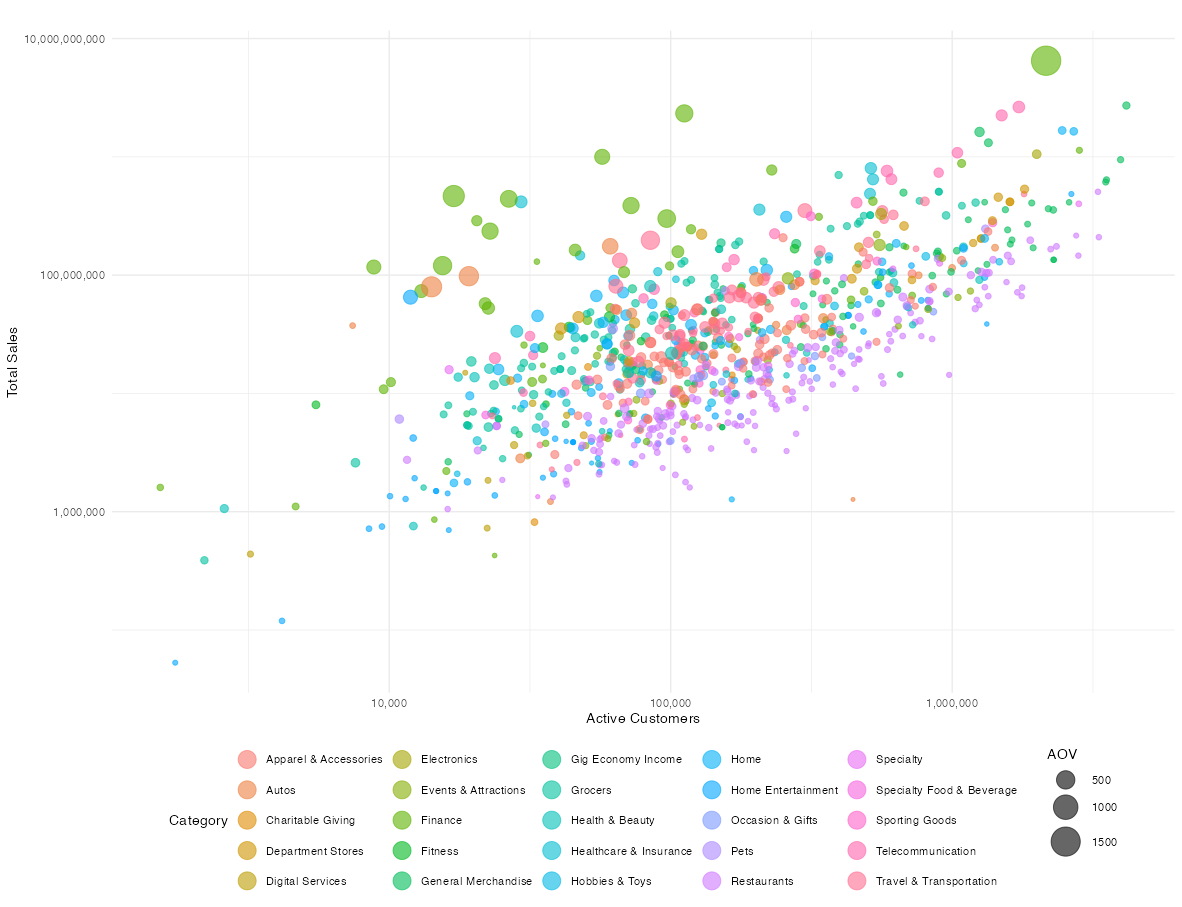}
    \begin{minipage}{\textwidth}{\footnotesize \textit{Note:} The horizontal axis reports average annual active customers, and the vertical axis reports average annual sales. Point size is proportional to median average order value (AOV); the legend identifies the 25 industry categories. All measures use the 2017--2018 credit and debit card data.}\end{minipage}
\end{figure}

Figure \ref{fig:earnest_visualize} summarizes cross-company variation in active customers, sales, order value, and industry. 
The 966 companies in our dataset span 25 industries such as apparel \& accessories (e.g., H\&M), restaurants (e.g., Chipotle), department stores (e.g., Macy's), digital services (e.g., Microsoft 365), events \& attractions (e.g., AMC), general merchandise (e.g., Target), grocers (e.g., Whole Foods), healthcare \& insurance (e.g., State Farm), and home improvement (e.g., Home Depot). In Appendix \ref{append:examplecompany}, we provide a sample of companies from within each industry. 
Restaurants (192 companies), which is the most represented category in our data, are dispersed widely in terms of active customer size but typically have smaller AOVs, leading to relatively lower total sales compared to other categories.
The Grocers category (111 companies), the second-largest category in the dataset, features frequent and regular transactions, and is predominantly situated in the center of the plot. 
The Apparel \& Accessories category (109 companies) features a higher AOV than restaurants leading to higher total sales.

Table \ref{tab:summary_stat} shows summary statistics of customer metrics over the 966 companies in our dataset over the calibration period (i.e., the time period over which we estimate the model) and the holdout period (i.e., the time period we hold aside to assess predictive accuracy). Within the credit and debit card panel\footnote{There is some evidence that trends within the credit card panel are representative of aggregate-level data at the population-level. See \citet{kim2024wheels} for details.}, company size varies widely. For example, customers acquired and total sales per week range from 350 to 1,855 and from \$127.9K to \$936.7K, respectively, over their interquartile ranges. 
In an average week, a typical company acquires approximately 776 new customers, has approximately 4,364 active customers, generates 0.04 repeat orders per acquired customer, with approximately \$50 spent per order.
Over the holdout period, new-customer acquisition slows while the number of weekly active customers increases; repeat orders per customer decline while AOV remains roughly unchanged. The net result is a slight upward trend in total sales, reflecting the growth in active customers. 
\begin{table}[!htbp]
\caption{Summary Statistics Across 966 Companies \label{tab:summary_stat}}
\resizebox{\columnwidth}{!}{%
\begin{threeparttable}
\begin{tabular}{@{}lrrrrrr@{}}
\toprule
Customer Metrics & \multicolumn{3}{c}{Calibration Period} & \multicolumn{3}{c}{Holdout Period} \\ \midrule
 & \multicolumn{1}{c}{25\%} & \multicolumn{1}{c}{Median} & \multicolumn{1}{c}{75\%} & \multicolumn{1}{c}{25\%} & \multicolumn{1}{c}{Median} & \multicolumn{1}{c}{75\%} \\ \cmidrule(l){2-4} \cmidrule(l){5-7} 
Acquisition (per week) & 350.3 & 775.9 & 1,854.9 & 289.4 & 606.6 & 1,310.9 \\
Active Customers (per week) & 2,133.7 & 4,364.2 & 12,858.9 & 2,266.4 & 4,533.9 & 13,396.1 \\
ROPC (per week) & 0.020 & 0.040 & 0.094 & 0.015 & 0.030 & 0.074 \\
AOV (per week) & 26.1 & 49.5 & 99.1 & 26.9 & 49.6 & 99.8 \\
Sales (week) & 127,910.3 & 338,153.7 & 936,679.1 & 137,732.1 & 352,013.6 & 1,046,004.0 \\ \cmidrule(l){2-4} \cmidrule(l){5-7} 
Acquisition (cumulative) & 41,338 & 91,561 & 218,874 & 13,602 & 28,510 & 61,610 \\
Repeat Orders (cumulative) & 234,353 & 524,346 & 1,737,618 & 103,544 & 227,281 & 748,269 \\
Sales (cumulative) & 15,093,417 & 39,902,136 & 110,528,129 & 6,473,411 & 16,544,638 & 49,162,167 \\ \bottomrule
\end{tabular}%
    \begin{tablenotes}
      \small
      \item \textit{Note: } This table presents summary statistics for customer metrics across 966 companies. ROPC denotes repeat orders per acquired customer; AOV denotes average order value. The calibration period spans January 2017 to March 2019 (27 months), while the holdout period covers April 2019 to February 2020 (11 months). The first five rows report weekly averages; the final three rows show cumulative totals over each period. Values represent the 25th percentile, median (50th percentile), and 75th percentile across companies.
    \end{tablenotes}
  \end{threeparttable}
  }
\end{table}

\subsection{Estimands of Interest\label{sec:estimandsofinterests}}

\begin{figure}[!htbp]
    \centering
    \caption{Cohorted Panel Data (Example: Wayfair)}   
    \includegraphics[width=0.95\textwidth]{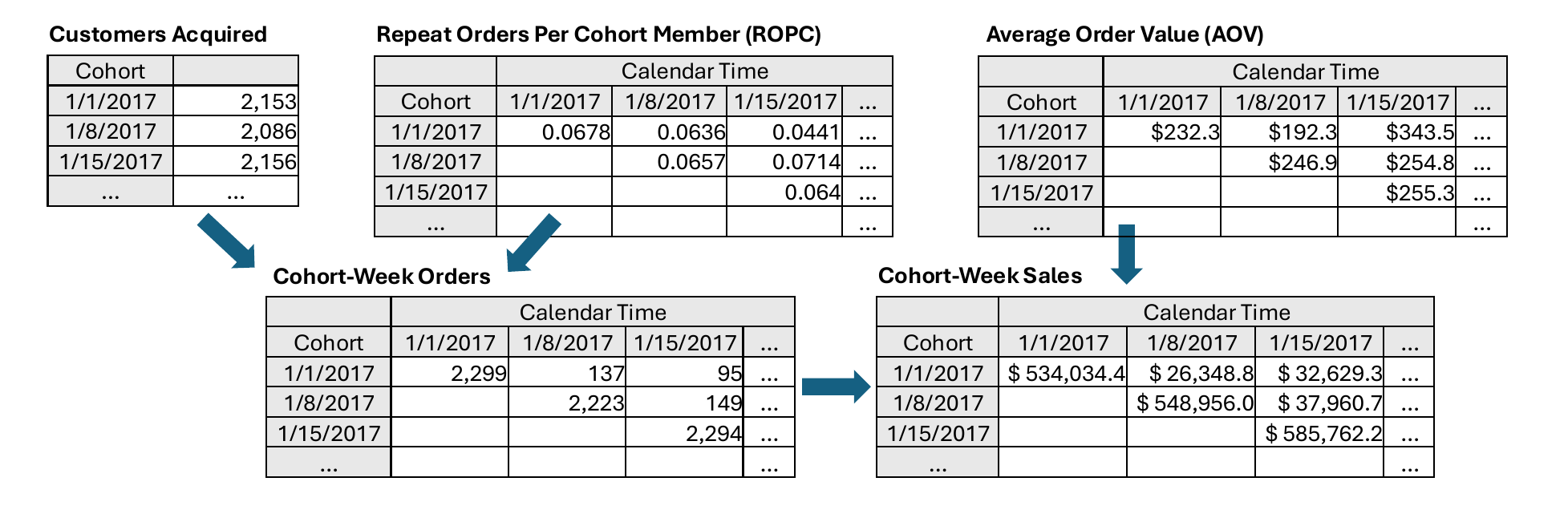}
    \label{fig:cohorted_data_wayfair}
    \begin{minipage}{\textwidth}{\footnotesize \textit{Note:} Cohorted panel data structure using data from Wayfair. Rows index weekly acquisition cohorts; columns represent calendar weeks. Three tables in the first row highlight the primary inputs for customer base analysis: customers acquired, repeat orders per cohort member, and average order value. Arrows indicate the multiplicative relationships that generate cohort-specific orders or sales.}\end{minipage}
\end{figure}

The estimands of interest in customer base analysis and customer-based corporate valuation are cohort-level customer behaviors over time. Figure \ref{fig:cohorted_data_wayfair} illustrates how total sales is constructed as a function of three distinct customer behavioral processes, using data from Wayfair. 
We see, for example, that 2,086 customers were acquired during the week beginning January 8, 2017 (top-left table). A customer is acquired when they place their very first order with the company, implying 2,086 orders coming from initial purchases made by this cohort in this time period. 
These customers placed 0.0657 repeat orders per customer in the week in which they were acquired, and another 0.0714 in the week thereafter (top-middle table). 
The total number of orders placed by this cohort in the week beginning January 8, 2017 is therefore the sum of initial and repeat orders: 2,086 + 137 (0.0657 $\times$ 2,086), or 2,223 (bottom-left table). These orders are then multiplied by the AOV for that cohort in that time period (top-right table), giving us total sales from a given acquisition cohort in a given calendar time period (bottom-right table). For example, the revenue generated by the January 8, 2017 cohort in the week beginning January 8, 2017 is \$548,956, which is equal to the total number of orders of 2,223 multiplied by the AOV associated with that cohort-time period of \$246.9 (displayed figures are rounded; products are computed from unrounded values). In a similar manner, we can obtain all diagonal elements within the tables summarizing total orders and total spend by cohort-time period.

After a cohort is acquired, it can only generate repeat purchases because initial purchases can only happen in the time period in which the cohort was acquired, by construction. Otherwise, however, the way in which we derive total orders from customers acquired and repeat orders per customer, and total sales from total orders and AOV, remains the same. In this way, we obtain all off-diagonal elements of the tables for total orders and total spend by cohort-time period. 

Aggregating such cohort-specific revenues across all cohorts for any given week yields the aggregated total weekly revenue, which is a key input for company valuation calculations. In that way, total revenue is decomposed into these three components, which serve as both the inputs and outputs of the model.

Representing the data in a cohorted format is preferable in our empirical setting for several reasons.
First, cohorting greatly reduces data sparsity, which in turn yields more stable signals for model training and thus improves the model's ability to predict future cohort-level data. Cohorting aggregates individual-level outcomes into cohort--time cells, grouping same-period first purchasers and tracking their collective behavior over subsequent weeks. This substantially reduces zero inflation relative to an individual-by-day panel and as a result, autoregressive deep learning models like the Transformer tend to produce more reliable forecasts in this context. 
Second, cohorted data protects customer privacy. Modelers work with aggregated cohort behavior rather than individual transaction histories.
Third, many marketing and financial decisions are operationalized ex ante at predetermined segment or cohort levels rather than at the individual customer level. Firms routinely define segments by acquisition channel, geography, product category, or business unit based on both operational and strategic considerations. Practitioners can apply our approach by filtering transaction data to the relevant segment and then constructing cohorts within that filtered subset, which gives them the predictive benefits of reduced sparsity while generating forecasts at the strategically appropriate level of granularity.
This segment-and-cohort structure aligns naturally with how company stakeholders use these insights. Managers and investors report acquired customers over time, cohort-level repeat purchase rates and average order values by tenure, customer lifetime value (CLV), and sales at the cohort or segment level. Similarly, discounted-cash-flow models project revenue by period (overall or by segment), and unit-economics dashboards track cohort- or segment-level trajectories. In each of these settings, individual-level predictions would be unnecessarily granular for the decision at hand.

\begin{figure}[!htbp]
    \centering
    \caption{Weekly Customer Acquisition and Cohort-Level Repeat Purchasing (Wayfair) \label{fig:model-free-evidence}}
    \includegraphics[width=\textwidth]{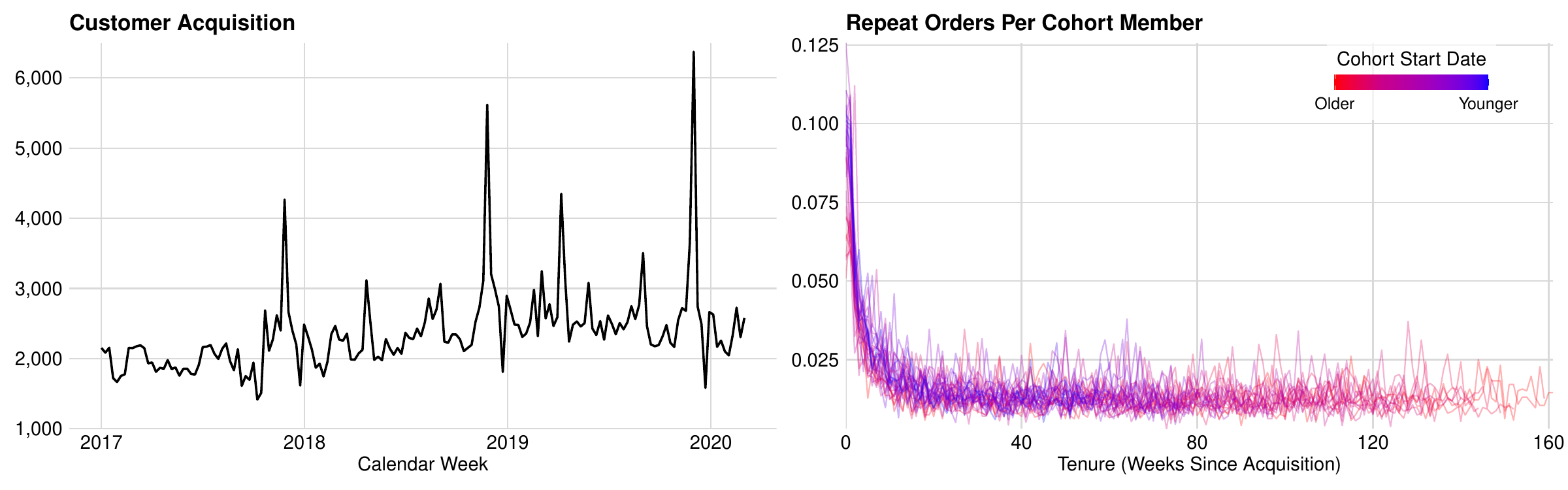}    
        \begin{minipage}{\textwidth}{\footnotesize \textit{Note:} The black line in the left panel shows weekly customer acquisitions for Wayfair. In the right panel, each line represents a cohort's average repeat orders per customer as a function of customer tenure. Cohorts are defined as groups of customers acquired within the same calendar week.}\end{minipage}
\end{figure}

Figure \ref{fig:model-free-evidence} presents descriptive evidence of temporal dynamics in customer acquisition and retention behavior for Wayfair. The left panel displays weekly customer acquisitions, which exhibit substantial volatility around the smoothed trend. Several notable features emerge. Pronounced spikes occur at irregular intervals, suggesting the influence of promotional events or external shocks, and the baseline acquisition rate itself appears to undergo regime shifts rather than following a smooth parametric form.

The right panel illustrates repeat orders per cohort member (ROPC) as a function of customer tenure, with each line representing a distinct acquisition cohort. All cohorts exhibit a sharp initial decay in ROPC during the first 20 to 40 weeks post-acquisition, consistent with standard retention models. However, subsequent behavior reveals considerable heterogeneity. Early cohorts (red lines) display markedly different decay rates compared to more recent cohorts (blue lines), and cohorts stabilize at different levels and display varying degrees of volatility at higher tenure values. Several cohorts exhibit non-monotonic patterns that deviate from smooth exponential decay, suggesting that retention dynamics may be influenced by cohort-specific factors such as acquisition timing or prevailing market conditions.

These patterns suggest that customer acquisition and retention processes may be influenced by time-varying factors, such as marketing mix changes, competitive dynamics, or macroeconomic conditions, that standard parametric models might treat as unexplained variance. While conventional approaches typically fit a smooth baseline, with deviations from that baseline incorporated through handcrafted covariates, they may not fully exploit the information contained in these systematic deviations. The modeling framework we propose in the following section is designed to capture these additional signals across separate customer-base processes.

\section{A Joint Customer-Base Prediction Framework}\label{sec:modeling}

The preceding section decomposes firm revenue into three customer-base primitives: customer acquisition, repeat purchasing, and order value. A forecasting system for customer-based revenue should therefore recover not only the level of future sales, but also the customer-base path through which sales are generated. We develop a joint multi-task prediction framework that learns from the joint history of these primitives, produces primitive-specific forecasts, and uses downstream revenue as an auxiliary alignment signal during training. This section presents the model at the level needed to interpret the empirical analyses; Web Appendix~\ref{wa:model_details} provides the full implementation details.

\subsection{Customer-Base Primitives and Revenue Decomposition}\label{sec:modeling_primitives}

The unit of analysis is a cohort-week. Let $i$ index acquisition cohorts and $t$ index calendar weeks. Let $A_t$ denote calendar-week acquisition. Once cohort $i$ enters in week $t_0(i)$, its size is the fixed historical attribute $A_i=A_{t_0(i)}$. For cohort-week $(i,t)$, let $R_{i,t}$ denote repeat orders per customer (ROPC) and $V_{i,t}$ denote average order value (AOV). Cohort-week sales ($Y_{i,t}$) and aggregate weekly sales ($S_t$) satisfy
\begin{equation}\label{eq:cohort_sales_main_jm}
Y_{i,t}=A_i\left(\mathbbm{1}\{t=t_0(i)\}+R_{i,t}\right)V_{i,t},
\qquad
S_t=\sum_{i\in\mathcal I_t}Y_{i,t},
\end{equation}
where $\mathcal I_t$ is the set of cohorts active in week $t$. Equation~\eqref{eq:cohort_sales_main_jm} defines the customer-base decomposition that links the customer-base primitives to revenue. Sales growth can come from acquiring more customers, increasing repeat purchase intensity, increasing order value, or some combination of the three. Conversely, two firms with similar sales trajectories can have very different customer-base trajectories. Forecasting the primitives directly therefore provides a behavioral interpretation of revenue forecasts.

At the end of week $t$, the model observes recent histories of acquisition, ROPC, AOV, and an auxiliary aggregate-sales channel, along with calendar and cohort-tenure covariates known at the forecast origin. It predicts next week's acquisition, ROPC, and AOV and constructs the revenue forecast implied by those components. Past aggregate sales summarize the realized downstream contribution of all active cohorts, but future sales are unknown. The model therefore predicts this stochastic input channel recursively only to update future input windows; the focal revenue forecast remains the component-implied forecast constructed from predicted acquisition, ROPC, and AOV. Web Appendix~\ref{wa:input_data} and Web Appendix~\ref{wa:output_data} provide the detailed input and output construction.

\subsection{A Joint Multi-Task Prediction Model}\label{sec:modeling_mtl}

Acquisition, ROPC, and AOV are distinct customer-base primitives, but they often reflect overlapping business conditions. A promotion, product launch, seasonal demand peak, pricing change, or competitive shock may appear in more than one primitive. Estimating each primitive separately can discard useful shared information, while forcing the primitives to move identically would ignore their distinct dynamics. CBMT therefore combines shared representation learning, primitive-specific forecasts, and an auxiliary revenue-alignment regularizer.

\begin{figure}[!htbp]
    \centering
    \caption{Customer-Based Multi-task Transformer (CBMT)}
    \includegraphics[width=0.7\textwidth]{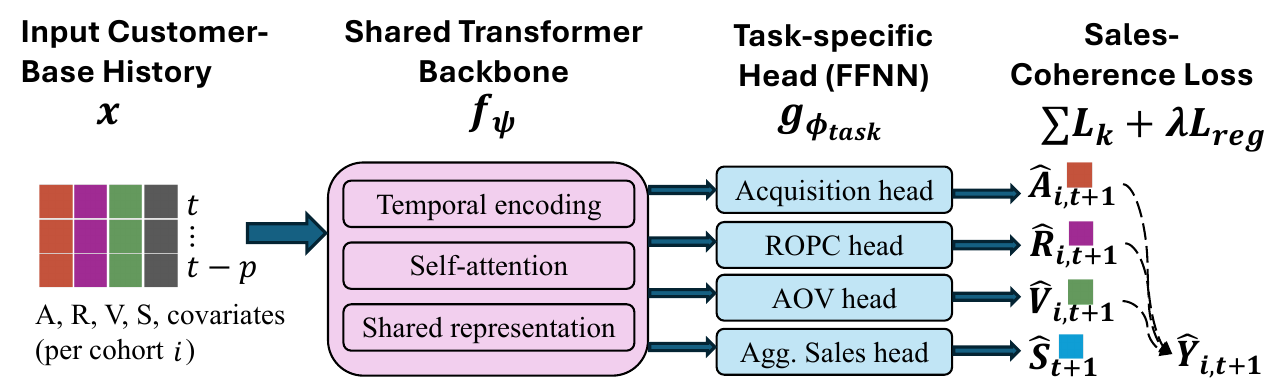}
    \label{fig:simpler_model}
    \vspace{0.3em}
    \begin{minipage}{0.95\textwidth}
        \footnotesize \textit{Note:} The figure summarizes CBMT. For each cohort $i$, the model takes as input a recent 20-week window of customer-base histories, including acquisition (A), repeat orders per customer (R), average order value (V), an auxiliary aggregate-sales channel (S), and known covariates. A shared Transformer backbone learns a common representation from these histories, and task-specific heads translate this representation into forecasts of acquisition, ROPC, AOV, and an auxiliary aggregate-sales update. The auxiliary aggregate-sales update is used to maintain the stochastic covariate input channel during recursive forecasting. The reported cohort-week revenue forecast ($\widehat Y$) is constructed from the predicted customer-base components ($\widehat A, \widehat R, \widehat V$). During training, a revenue-alignment loss ($L_{reg}$) regularizes the component forecasts by penalizing combinations that imply implausible downstream revenue.
    \end{minipage}
\end{figure}

For each cohort-week $(i,t)$, let $D_{i,t}$ denote the recent 20-week input window. The model maps this window into a shared customer-base representation, $h_{i,t}=f_{\psi}(D_{i,t})$, and then uses primitive-specific heads to preserve differences across outcomes. The acquisition head produces cohort-conditioned calendar-week forecasts, $\widehat A^{(i)}_{t+1}$, which are averaged within week to obtain $\widehat A_{t+1}$. The ROPC and AOV heads produce cohort-week forecasts, $\widehat R_{i,t+1}$ and $\widehat V_{i,t+1}$. A fourth head produces an auxiliary aggregate-sales update, $\widehat S^{\mathrm{aux}(i)}_{t+1}$, used only to update the stochastic input channel during walk-forward prediction. We implement the shared layer with a Transformer, but the substantive idea is shared learning with primitive-specific prediction layers rather than the Transformer architecture itself. Web Appendix~\ref{wa:transformer} provides the full technical details.

Training minimizes a weighted objective that combines task-specific prediction losses with an auxiliary revenue-alignment loss:
\begin{equation}\label{eq:loss_main_jm}
\mathcal L(\psi,\phi)=
\sum_{k\in\{A,R,V,S^{\mathrm{aux}}\}}
\omega_k\mathcal L_k(\psi,\phi_k)
+
\omega_C\mathcal L_C(\psi,\phi_A,\phi_R,\phi_V).
\end{equation}
The task losses reward one-step-ahead accuracy for acquisition, ROPC, AOV, and the auxiliary aggregate-sales channel. The last term, $\mathcal L_C$, connects the upstream component forecasts to downstream revenue. Specifically, it combines the predicted acquisition, ROPC, and AOV components into an auxiliary cohort-week revenue object and penalizes combinations that imply implausible revenue magnitudes relative to observed downstream sales.

Because acquisition is forecast as a calendar-week inflow, whereas the size of an existing cohort is a fixed historical attribute, $\mathcal L_C$ is implemented as coherence-oriented predictive regularization rather than as a hard accounting constraint. This distinction is important. The model must forecast calendar-week acquisition to initialize cohorts born during the forecast horizon. The training-time alignment loss therefore uses downstream revenue to regularize the joint scale and interaction of the component forecasts without claiming to reconstruct literal cohort-week sales. At inference, the reported component-implied revenue forecast uses observed cohort size for cohorts already present before the forecast horizon and predicted acquisition for cohorts born during the forecast horizon. Web Appendix~\ref{wa:training} gives the exact construction of $\mathcal L_C$, and Web Appendix~\ref{wa:inference} describes the inference-time revenue construction.

\subsection{Estimation and Forecasting Protocol}\label{sec:modeling_forecast_protocol}

The model is trained on the calibration period and evaluated using a walk-forward forecasting procedure that mirrors the information available to a manager. At the beginning of the holdout period, the model receives the most recent observed histories and known deterministic covariates, forecasts one week ahead, averages the cohort-conditioned acquisition and auxiliary aggregate-sales forecasts within calendar week before appending its own predictions for the stochastic customer-base inputs, and then shifts the input window forward; future values of deterministic covariates are known by construction, and realized holdout outcomes never enter the input window. For cohorts already observed, cohort size is fixed and the model forecasts future ROPC and AOV. For cohorts acquired during the forecast horizon, predicted acquisition in the birth week initializes cohort size for subsequent revenue contributions. The reported aggregate revenue forecast sums these component-implied cohort contributions without using future information.

Joint learning need not always help. If the primitives are weakly related or one is dominated by idiosyncratic shocks, pooling can transmit noise. The empirical analysis therefore benchmarks CBMT against single-task and other non-joint alternatives and examines whether joint forecasting is more valuable when the primitives exhibit stronger calibration-period co-movement. These tests evaluate shared-driver signal extraction and revenue-alignment discipline as empirical implications of the framework rather than assumptions built into the model.

\section{Empirical Results} \label{sec:results}

\subsection{Study Design}
\begin{figure}[!htbp]
    \centering
    \caption{Data Structure: Calibration and Holdout Periods with Cohort Segmentation}
    \includegraphics[width=0.6\textwidth]{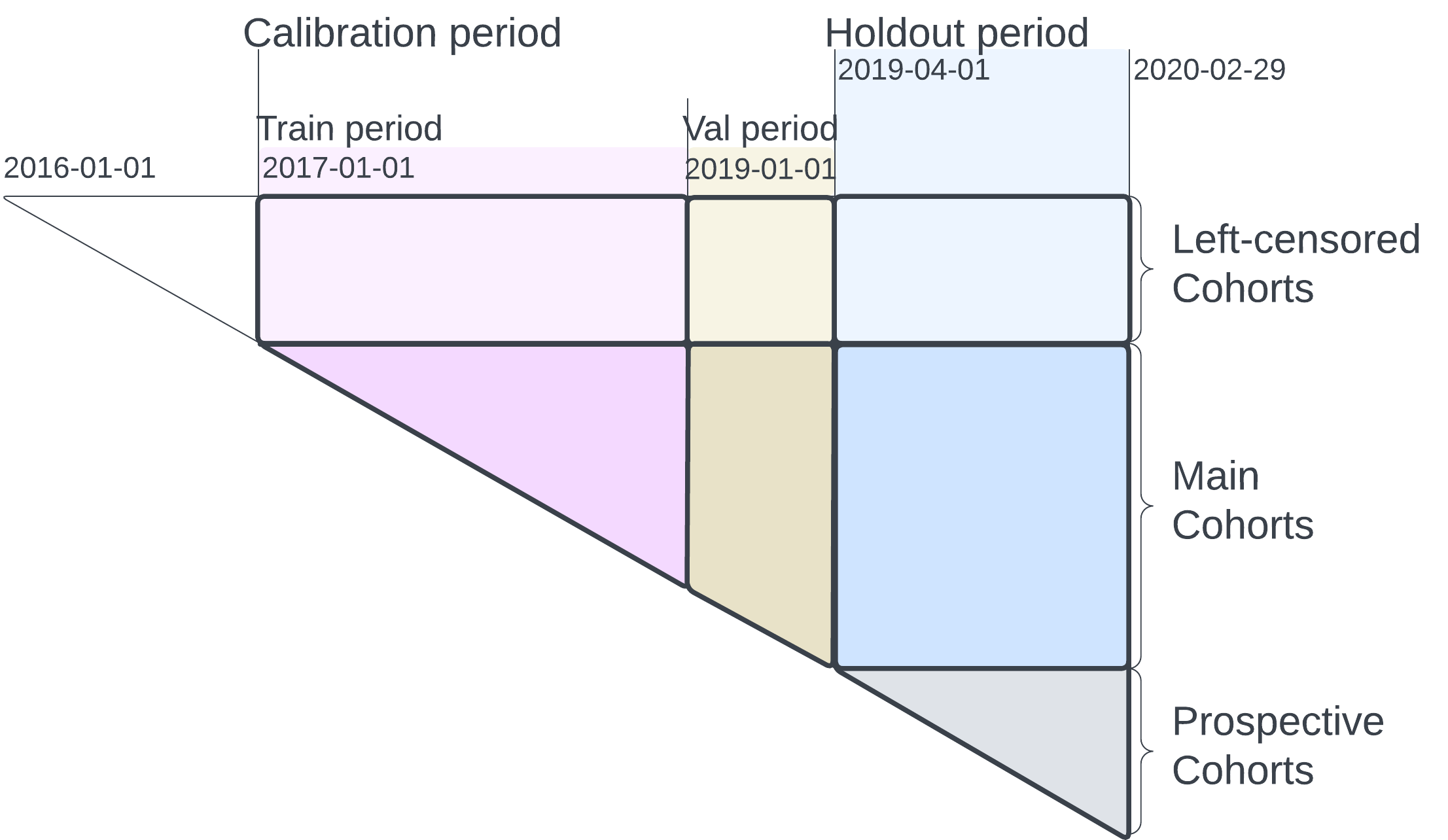}
    \label{fig:cohort-triangle}
    \begin{minipage}{0.9\textwidth}
        \footnotesize \textit{Note:} This figure illustrates the cohorted data structure (tabular form shown in Figure \ref{fig:cohorted_data_wayfair}). Temporal segmentation of the dataset showing three cohort categories: left-censored cohorts (customers acquired before the end of 2016), main cohorts (acquired during calibration), and cold cohorts (to be acquired during holdout).
    \end{minipage}
\end{figure}

Our empirical analysis uses transaction data spanning January 2016 through February 2020. We use the 2016 data to identify customer acquisition timing, with the primary study period extending from January 2017 through February 2020. The dataset has two components: calibration data available for model training and holdout data withheld during training. The calibration period spans 27 months from January 2017 to March 2019, while the holdout period covers 11 months from April 2019 to February 2020, concluding before the COVID-19 pandemic. Figure \ref{fig:cohort-triangle} illustrates this data structure.

To establish customer acquisition cohorts, we designate 2016 as a baseline identification period. Customers with no purchases throughout 2016---indicating at least one year of inactivity---who then make their first observed purchase in, for example, the week of January 29, 2017, are assigned to the `2017-01-29' cohort. This approach allows us to distinguish newly acquired customers from recently active returning ones.\footnote{Acquisition is thus defined operationally as the first observed purchase following at least one year of inactivity. A customer acquired before our observation window who remained inactive throughout 2016 and returned later would be classified as newly acquired, so measured acquisition may include some reactivations of long-dormant customers. The one-year washout limits this misclassification, and to the extent that a return after more than a year of inactivity resembles a fresh acquisition behaviorally, the approximation is innocuous for forecasting.}

Under this framework, our sales predictions for the holdout period aggregate three distinct customer segments: (1) `left-censored cohorts'---customers acquired before the end of 2016 whose exact acquisition timing remains unknown, (2) `main cohorts'---customers acquired during the calibration period, and (3) `cold cohorts'---customers who will be acquired during the holdout period itself. Our calibration data encompasses both left-censored and main cohorts. Established businesses with long-standing customer bases (such as McDonald's or Starbucks) naturally tend to have a larger proportion of their customers in the left-censored cohort.

We evaluate model performance using symmetric Mean Absolute Percentage Error (SMAPE) as our primary metric. SMAPE provides an intuitive percentage-based interpretation of forecast accuracy while addressing the directional bias inherent in traditional MAPE:
\begin{equation*}
\text{SMAPE} = \frac{100\%}{n} \sum_{t=1}^{n} \frac{|F_t - X_t|}{\left(\frac{|X_t| + |F_t|}{2}\right)}
\end{equation*}
where $F_t$ denotes the forecast value and $X_t$ the actual value in period $t$. By using the average of actual and forecast values in the denominator, SMAPE treats over- and under-predictions symmetrically.


To select hyperparameters for the autoregressive deep learning models, including CBMT, we use temporal holdout validation. The calibration period is partitioned into training data (January 2017 to mid-November 2018) and validation data (late November 2018 to March 2019). This split preserves the sequence of the data while enabling out-of-sample validation of hyperparameter choices.

For each company, we optimize model architecture parameters (learning rate, number of layers, hidden layer dimensions, and output layer specification) based on validation loss. 
Our hyperparameter selection follows a multi-stage approach: we begin with exploratory searches to identify promising parameter ranges, then perform systematic grid searches within these ranges, and conclude with automated optimization using the Optuna framework \citep{akiba2019optuna} for final refinement. The stages move from broad exploratory ranges to focused search and final refinement.

For probabilistic benchmark models, we optimize parameters based on calibration period fit, treating operational characteristics (such as the start of commercial operations) as tuning parameters selected for best model performance. Parameters are chosen via exhaustive grid search.

\subsection{Benchmark Models} \label{sec:benchmark_models}

We evaluate CBMT against four categories of established benchmarks from the customer analytics literature: (1) classical customer-based corporate valuation (CBCV) models that link customer metrics to firm value, (2) contemporary customer relationship management (CRM) models that employ probabilistic frameworks for customer behavior, (3) deep learning approaches that have recently entered the customer analytics domain, and (4) traditional tree-based machine learning models.

\textbf{Classical CBCV Models.} The foundational CBCV literature provides three approaches that empirically estimate corporate valuation from customer purchase data: GLS \citep{gupta2004valuing}, LMP \citep{libai2009diffusion}, and SSW \citep{schulze2012linking}. These models link customer equity to firm financial performance through variants of the technological substitution model \citep{bass1969new}, though their implementation approaches differ. The original specifications model the number of active customers over time but do not model revenue directly; they derive revenue projections using average historical revenue per customer. For the benchmark implementation, we instead estimate average order value (AOV) using linear regression with month dummies, making the classical benchmarks more competitive than their original specifications.

\textbf{Probabilistic CRM Models.} More recent CBCV frameworks by \cite{mccarthy2017valuing} and \cite{mccarthy2018customer} model all three customer behavioral components and were specifically designed with period-by-period revenue forecasting accuracy in mind. Given our access to detailed cohort-time panel data, we adapt the underlying statistical models from this CBCV literature---specifically the Weibull-Gamma and Pareto/NBD specifications---and apply them directly to our cohort-time structure. We label these implementations as WG-PNBD models to reflect their statistical components: Weibull-Gamma for acquisition \citep{morrison1980jobs}, Pareto/Negative Binomial for repeat purchases \citep{schmittlein1987counting}, and linear regression for AOV. We also include an extended version (WG-PNBD-E) that incorporates covariates following \cite{bachmann2021role}, enabling both acquisition and repeat purchase models to capture additional dynamics through month dummies and cohort indicators.

\textbf{Deep Learning Benchmark.} We include LSTM (\citealt{hochreiter1997long}) networks as a benchmark model, introduced into CRM literature by \cite{valendin2022customer}. LSTM, like the Transformer, is an autoregressive deep learning model designed for sequential data; as in Markov models, the previous latent state influences the current one through recurrent connections. We fit separate, independently estimated LSTM models for each behavioral process, with covariates as specified in the CBMT configuration for consistency.

\textbf{Traditional Machine Learning Benchmarks.} We include Random Forest (RF) and XGBoost (XGB) as flexible nonparametric benchmarks that can capture nonlinearities and interactions among lagged customer behaviors and covariates. Both models use the same input variables and forecasting protocol as CBMT. We estimate a single multi-output model that predicts acquisition, ROPC, and AOV one step ahead and generates holdout forecasts recursively.

Table \ref{tab:cbcv_benchmark} summarizes the modeling approaches employed by each benchmark for the three customer behavioral processes.

\begin{table}[!htbp]
\caption{Benchmark Models}
\label{tab:cbcv_benchmark}
\centering
\begin{threeparttable}
\begin{tabular}{@{}llll@{}}
\toprule
Model & Acquisition & ROPC & AOV \\ \midrule
GLS        & Bass Diffusion-based       & Regression       & Regression    \\ 
LMP        & Bass Diffusion-based       & Regression       & Regression    \\ 
SSW        & Bass Diffusion-based       & Regression       & Regression    \\ 
WG-PNBD    & Weibull-Gamma             & Pareto/NBD       & Regression     \\ 
WG-PNBD-E  & Weibull-Gamma + Covariates & Pareto/NBD + Covariates & Regression \\ 
LSTM       & LSTM                      & LSTM             & LSTM         \\
RF & RF & RF & RF \\
XGB & XGB & XGB & XGB \\
\bottomrule
\end{tabular}
\begin{tablenotes}
\footnotesize
      \item \textit{Note:} GLS has three variants, LMP has two variants, SSW has three variants, WG-PNBD has two variants, and WG-PNBD-E has two variants. In total, we evaluate CBMT against 15 benchmark variants, yielding 16 models when CBMT is included. Details appear in Web Appendix \ref{sec:wa_benchmark_selection}.
    \end{tablenotes}
    \end{threeparttable}
\end{table}

To make the benchmarks as competitive as possible, we implemented multiple versions of each model family, systematically varying data granularity and covariate specifications. We evaluate CBMT against 15 benchmark variants, yielding 16 models in total when CBMT is included. Through the model-selection procedures detailed in Web Appendix \ref{sec:wa_benchmark_selection}, we identified strong representative benchmarks from each family: GLS v2 (aggregate data with covariates), WG-PNBD-E PP (partially pooled cohort estimation with covariates), LSTM, and RF. For readability, these four models serve as the primary benchmarks in the main text. The broader Web Appendix comparisons report XGB and alternative implementations within the CBCV and probabilistic CRM model families.

\subsection{Absolute and Relative Performance Evaluation}

Table \ref{tab:eval_main} summarizes evaluation results for 966 companies. For readability, the table reports CBMT alongside four representative benchmarks: GLS, WG-PNBD-E, LSTM, and RF. Web Appendix \ref{sec:wa_benchmark_selection} reports the full comparison against 15 benchmark variants, including XGB and alternative implementations within the CBCV and probabilistic CRM model families. In measuring prediction error for holdout data, we report SMAPE, with lower values indicating higher accuracy. Web Appendix \ref{append:main_robustness} reports the corresponding results using MASE as an alternative scale-invariant error metric. The substantive conclusions are unchanged.

\begin{table}[!htbp]
\centering
\small
\caption{Evaluation with SMAPE Measure}
\label{tab:eval_main}
\begin{tabular}{llrrrrr}
\toprule
Processes & & CBMT & GLS & WG-PNBD-E & LSTM & RF \\
\midrule
Acquisition & ($\mu$) & 19.30 & 27.71 & 23.31 & 26.03 & 26.02 \\
 & (\%) &  & (+30\%) & (+17\%) & (+26\%) & (+26\%) \\
 & ($\sigma$) & 20.12 & 23.86 & 23.66 & 29.09 & 20.68 \\
ROPC & ($\mu$) & 35.56 & 178.64 & 50.85 & 42.56 & 37.26 \\
 & (\%) &  & (+80\%) & (+30\%) & (+16\%) & (+5\%) \\
 & ($\sigma$) & 24.68 & 17.70 & 27.94 & 25.90 & 24.30 \\
AOV & ($\mu$) & 35.94 & 39.37 & 38.59 & 41.05 & 38.62 \\
 & (\%) &  & (+9\%) & (+7\%) & (+12\%) & (+7\%) \\
 & ($\sigma$) & 25.00 & 29.92 & 29.33 & 31.05 & 31.39 \\
Cohort-Week Sales & ($\mu$) & 53.91 & 170.93 & 64.47 & 61.06 & 59.29 \\
 & (\%) &  & (+68\%) & (+16\%) & (+12\%) & (+9\%) \\
 & ($\sigma$) & 30.40 & 21.41 & 32.44 & 31.06 & 33.38 \\
Total Sales & ($\mu$) & 15.48 & 42.36 & 22.11 & 28.15 & 23.57 \\
 & (\%) &  & (+63\%) & (+30\%) & (+45\%) & (+34\%) \\
 & ($\sigma$) & 19.14 & 27.82 & 23.60 & 28.77 & 24.94 \\
\bottomrule
\end{tabular}
\end{table}

CBMT is consistently strong across acquisition, ROPC, and AOV, and that consistency matters most when the primitive forecasts are combined into downstream sales. Relative to the closest benchmark for each outcome, CBMT reduces acquisition error by 17\%, ROPC error by 5\%, AOV error by 7\%, cohort-week sales error by 9\%, and weekly aggregate-sales error by 30\%. No single benchmark is the closest competitor across all outcomes, which is why a coordinated joint forecasting system can deliver larger downstream gains than the primitive-level comparisons alone suggest.

A natural question is how CBMT compares with a flexible model built only to forecast aggregate sales. As a practical benchmark, we evaluate a tuned Transformer that predicts weekly company-level sales directly from aggregate-sales history and calendar covariates, without forecasting acquisition, ROPC, or AOV. CBMT's mean total-sales SMAPE is 15.48, compared with 15.90 for the tuned direct model, a reduction of 2.65\%. The paired comparison is not statistically significant ($p=.222$), so the evidence does not establish that CBMT is more accurate than the direct model on average. The comparison nevertheless places CBMT's decomposable forecast on the same accuracy scale as a model that can devote all of its capacity to the topline because it is never asked to explain where sales come from. CBMT must simultaneously recover the acquisition, repeat purchasing, and order-value paths that generate sales, and its slightly lower point estimate indicates that this behavioral decomposition does not produce a detectable topline penalty. This comparison is distinct from the 30\% mean-SMAPE reduction relative to the strongest representative established customer-base benchmark. Web Appendix~\ref{wa:direct_sales_benchmark} reports the model, tuning design, and paired analysis.\footnote{Under the recursive one-step forecasting protocol, acquisition, ROPC, and AOV would be stochastic inputs whose future paths are unknown. Adding them to a sales-only model would therefore require either forecasting those paths recursively or adopting a different direct multi-horizon architecture, either of which defines a substantially different benchmark.}

CBMT also outperforms its single-task counterpart for a substantial majority of firms. It produces lower total-sales SMAPE for 718 of the 966 companies, while single-task forecasting performs better for 248, giving CBMT a 74.3\% win rate. Across all firms, mean total-sales SMAPE is 15.48 for CBMT and 20.54 for single-task forecasting; an oracle that selects the better of the two for each firm reaches 14.79. The resulting 0.69-point oracle gap motivates the deployment analysis, which asks whether calibration-period information can identify the firms that should switch to single-task forecasting.

For the established benchmarks in Table \ref{tab:eval_main}, our results are also robust to mean absolute scaled error (MASE). For total-sales prediction, CBMT's mean MASE is 39\% lower than WG-PNBD-E's and 49\% lower than RF's.

Moving from absolute to relative performance measures, we examine how each model ranks across companies. Table \ref{tab:avg_rank} presents the average rank of each representative main-text model across customer-base primitives and sales outcomes. Among the 16 models tested (including all variants of each benchmark model specified in Web Appendix \ref{sec:wa_benchmark_selection}), we assign a rank of 1 to the model with the smallest SMAPE and a rank of 16 to the model with the largest SMAPE for each company. The direct aggregate-sales model is not included because it does not forecast all focal outcomes.

Table \ref{tab:avg_rank} shows that CBMT achieves the best (lowest) average ranking across all metrics. The model performs particularly well for ROPC, with an average rank of 2.12, indicating it is the best or near-best model for most companies. This strong ROPC performance is accompanied by low average ranks for cohort-week sales (2.15) and total sales (3.06).

\begin{table}[!htbp]
\centering
\caption{Relative Performance Measured by Average Rank}
\label{tab:avg_rank}
\begin{threeparttable}
\begin{tabular}{@{}lrrrrr@{}}
\toprule
Models & Acquisition & ROPC & AOV & Cohort-Week Sales & Total Sales \\
\midrule
\textbf{CBMT} & \textbf{3.50} & \textbf{2.12} & \textbf{6.07} & \textbf{2.15} & \textbf{3.06} \\
LSTM & 5.30 & 3.85 & 9.37 & 4.43 & 7.80 \\
RF & 6.74 & 3.16 & 6.31 & 4.15 & 7.01 \\
WG-PNBD-E & 4.83 & 6.06 & 7.00 & 5.56 & 6.17 \\
GLS & 6.36 & 12.27 & 8.28 & 10.43 & 11.02 \\
\bottomrule
\end{tabular}
\begin{tablenotes}
\footnotesize
\item \textit{Note:} The rank of a method for a given outcome is the average of company-specific ranks for that method. Lower ranks indicate better performance, with 1 being the best and 16 being the worst. Each company-outcome pair is ranked by SMAPE across all 16 models.
\end{tablenotes}
\end{threeparttable}
\end{table}

Web Appendix Tables \ref{tab:append_avg_rank} and \ref{tab:append_p_rank1} report the full average-rank comparison and the corresponding probability of ranking first across all model variants.

In sum, CBMT performs consistently well across the three customer-base primitives and produces its largest improvement for weekly aggregated total sales. Relative to the strongest established benchmark for each outcome, the primitive-level gains are modest, whereas total-sales SMAPE is approximately 30\% lower than the best representative established customer-base benchmark. This pattern, reinforced by the average-rank results in Table \ref{tab:avg_rank} and the fuller relative-performance comparisons in Web Appendix \ref{append:relative_perform}, is consistent with the paper's central argument. The value of joint customer-base forecasting lies less in dominating each primitive-specific task than in producing better coordinated downstream revenue projections.

\subsection{Forecasting the Sources of Sales Change}
\label{subsec:sources_target_sales}

A key benefit of forecasting the customer-base primitives is that managers learn not only how much sales are expected to change but where the change will come from. This subsection asks whether the models deliver on that benefit, using the representation a manager would most naturally reach for: a sales bridge that starts from baseline sales and attributes the change in sales to changes in customer behavior. Comparing seasonally aligned 47-week baseline and target windows, we write the change in sales as the sum of exactly three contributions:
\begin{equation*}
\begin{aligned}
\text{Target sales}-\text{baseline sales}
={}&\text{Existing-base repeat-order volume}\\
&+\text{Existing-base repeat spend per order}\\
&+\text{Net customer-base replenishment}.
\end{aligned}
\end{equation*}
The first two terms capture how much more (or less) the customers a firm already had as of the February 24, 2019 baseline cutoff are ordering, and how much they spend per order when they do. Net customer-base replenishment captures the growth engine: sales from cohorts acquired after the cutoff, net of the baseline initial-order sales that mechanically roll off because a customer makes an initial purchase only once. The three contributions sum exactly to the change in sales, so the bridge always reconciles. Web Appendix~\ref{wa:source_sales} details the construction, reconciliation checks, benchmark eligibility, and common-sample design.

We evaluate each method by the mean absolute error (MAE) of its predicted contributions. To place firms of different sizes on a common scale, each error is expressed in percentage points of the firm's baseline sales, so an MAE of 5 means the predicted contribution misses the realized one by 5\% of baseline sales on average. Table~\ref{tab:source_mae_main} compares the seven methods able to produce the bridge, all on an identical 965-company sample. CBMT's MAE is 7.91 for the total change, 6.43 for repeat-order volume, 5.91 for repeat spend per order, and 5.80 for net customer-base replenishment. CBMT has lower MAE in 23 of the 24 benchmark-by-outcome comparisons. In the sole exception, WG-PNBD-E FP's repeat-spend-per-order MAE is 0.47 points lower than CBMT's, but the paired benchmark-minus-CBMT difference is not statistically distinguishable from zero (95\% CI: $[-1.01, 0.06]$).

\begin{table}[!htbp]
\centering
\scriptsize
\caption{Mean Absolute Error in Total Sales Change and Its Sources}
\label{tab:source_mae_main}
\resizebox{\textwidth}{!}{%
\begin{tabular}{lrrrrrrr}
\toprule
Outcome & CBMT & WG-PNBD-E PP & RF & XGB & WG-PNBD FP & WG-PNBD PP & WG-PNBD-E FP \\
\midrule
Total sales change & 7.91 & 18.17 & 22.60 & 28.87 & 22.70 & 22.13 & 18.62 \\
Existing-base repeat-order volume & 6.43 & 9.60 & 11.11 & 11.18 & 9.98 & 11.07 & 9.81 \\
Existing-base repeat spend per order & 5.91 & 8.27 & 8.70 & 12.65 & 7.83 & 8.22 & 5.43 \\
Net customer-base replenishment & 5.80 & 6.93 & 10.10 & 13.01 & 10.58 & 8.39 & 8.49 \\
\bottomrule
\end{tabular}%
}
\begin{minipage}{\textwidth}
\footnotesize
\noindent\textit{Note:} Entries are MAEs in percentage points of baseline sales; lower values mean forecasts are closer to realized outcomes. Every column uses the same 965 companies. Web Appendix~\ref{wa:source_sales} reports full MAE results, paired uncertainty, eligibility dispositions, and the prespecified sensitivity.
\end{minipage}
\end{table}

\begin{figure}[!htbp]
\centering
\caption{Realized and Predicted Sources of Sales Change: Two Illustrative Firms}
\label{fig:source_cases}
\includegraphics[width=\textwidth]{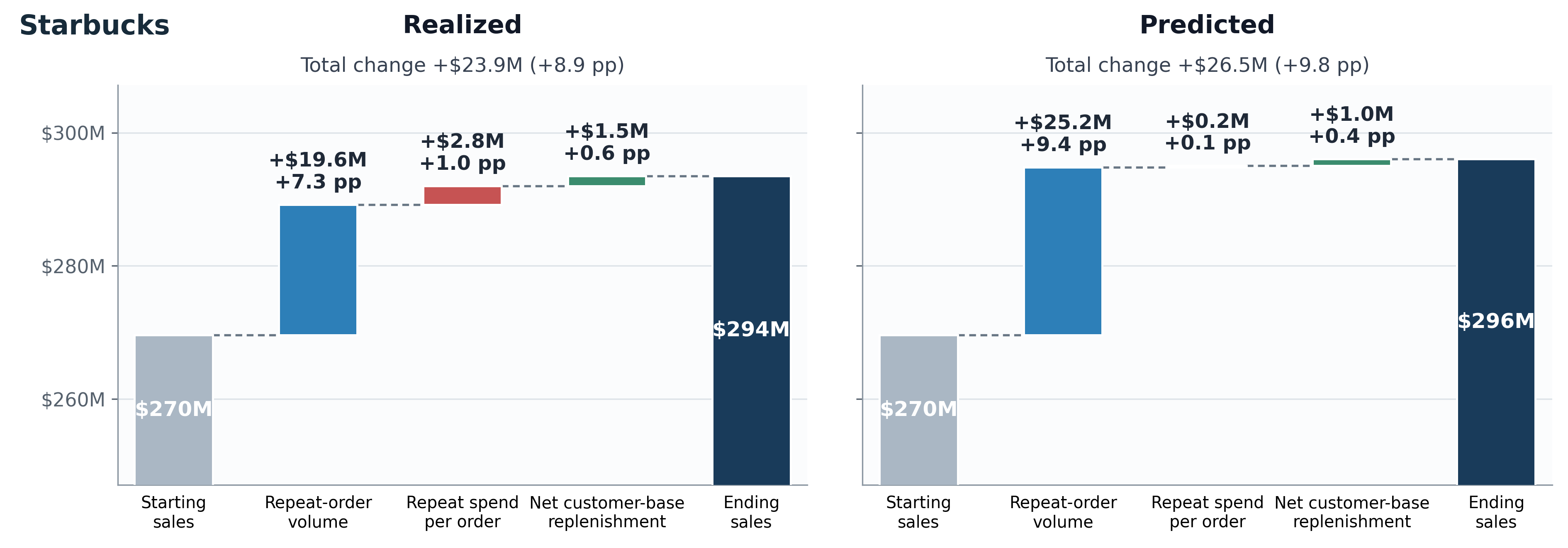}
\vspace{0.75em}

\includegraphics[width=\textwidth]{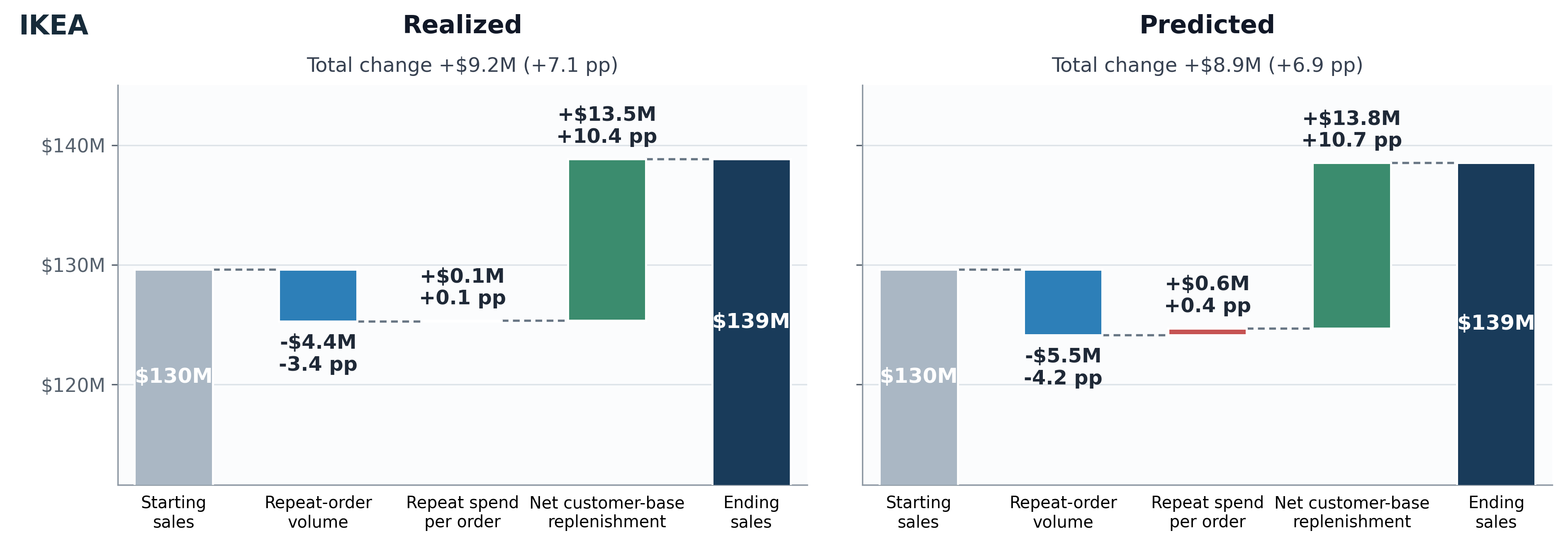}
\begin{minipage}{0.95\textwidth}
\footnotesize
\textit{Note:} Starbucks is shown at top and IKEA at bottom. Within each row, the left panel is the realized bridge and the right panel is the bridge implied by CBMT's predicted primitives; both start from the same realized baseline. Each bridge attributes the change in sales between seasonally aligned 47-week windows (baseline: April 8, 2018 to February 24, 2019; target: April 7, 2019 to February 23, 2020; baseline cohort cutoff: February 24, 2019) to the three contributions: existing-base repeat-order volume, existing-base repeat spend per order, and net customer-base replenishment. Bar labels report dollars and percentage points of baseline sales.
\end{minipage}
\end{figure}

Figure~\ref{fig:source_cases} shows what the bridge looks like in practice for two illustrative firms whose toplines grew similarly but for very different reasons. Starbucks' realized sales rise by 8.9 percentage points of baseline sales, compared with a predicted increase of 9.8 points, and nearly all of the realized growth comes from the existing base ordering more often: repeat-order volume contributes 7.3 points. IKEA's sales rise by a comparable 7.1 points, compared with a predicted increase of 6.9 points, yet its existing base contracts, with repeat-order volume subtracting 3.4 points; all of the realized growth comes from net customer-base replenishment, which contributes 10.4 points. A manager looking only at the toplines would see two similar growth stories. The bridge shows one firm deepening its existing customer relationships and the other outgrowing a shrinking base by adding new customers, and CBMT's predicted bridges recover both patterns. These cases illustrate the managerial view the bridge provides; the systematic evidence is the full-sample comparison in Table~\ref{tab:source_mae_main}.

We next examine the separate question of why coordinating the primitive forecasts improves downstream prediction.

\section{Why Joint Customer-Base Forecasting Helps}
\label{subsec:why_helps}

The sales-bridge analysis establishes how accurately the coordinated forecasts recover the sources of a sales change, but it does not explain why coordination improves forecast accuracy. Why should jointly forecasting customer-base primitives improve revenue prediction? Our argument begins with the simple premise that acquisition, repeat purchasing, and average order value are distinct outcomes but often reflect overlapping business conditions. Customer-equity research treats acquisition, retention or repeat purchasing, and spending as linked contributors to customer value \citep{rust2004return,gupta2004valuing,stahl2012impact}. Related work shows that acquisition conditions can predict subsequent customer quality and that purchase incidence, quantity, and spending may respond jointly to marketing actions and market conditions \citep{thomas2001methodology,lewis2006customer,villanueva2008impact,datta2015effect,gupta1988impact,pauwels2002long}. These relationships create an opportunity for joint forecasting, but they do not guarantee that pooling will always help.

We posit two mechanisms. The first is \textit{shared-driver signal extraction}. Acquisition, ROPC, and AOV are different noisy readouts of latent demand, relationship, and marketing states, such as promotional intensity, seasonality, category demand, product launches, pricing changes, and competitive shocks. When these latent states shape multiple behavioral primitives, modeling each primitive separately leaves useful information unused. A joint forecasting approach instead allows signals observed in one behavior to inform the prediction of the others. This mechanism makes a conditional prediction. Pooling should help when the primitives share temporal structure, including lead--lag patterns in which one primitive registers a change before the others, and it can hurt when one primitive is dominated by idiosyncratic noise, because joint learning then transmits noise rather than signal. These conditional predictions characterize the boundary conditions of joint forecasting's value.
The second mechanism is \textit{revenue-alignment discipline}, and it addresses a different issue. Component forecasts that each look reasonable on their own can still combine into a poor sales forecast, because their errors interact when the components are multiplied and aggregated into revenue. The auxiliary revenue-alignment regularizer discourages combinations of acquisition, ROPC, and AOV forecasts that imply implausible downstream revenue magnitudes. The regularizer is not a hard accounting constraint; it is a predictive discipline that makes primitive forecasts answerable to their downstream revenue implications. The two mechanisms thus carry distinct empirical implications, which the analyses below examine separately. Gains from shared-driver extraction should concentrate among firms whose primitives co-move more strongly, while gains from revenue alignment should appear in downstream sales accuracy even where primitive-level changes are modest.

Table~\ref{tab:mechanism_improvements} reports a diagnostic comparison of the single-task, joint, and full model families selected under scenario 3. The families use family-specific candidate grids and selection strategies rather than a common hyperparameter configuration, so the comparison is not a controlled architectural ablation and does not identify the causal effect of shared representation or revenue alignment. Instead, it asks whether the performance ordering among the scenario-3 selected families is consistent with the mechanisms proposed here. The single-task family forecasts the three primitives separately, the joint family uses a shared representation without revenue alignment, and the full family combines the shared representation with revenue alignment; Web Appendix~\ref{wa:ablation_protocol} details the family-specific selection procedures.

\begin{table}[!htbp]
\centering
\caption{Diagnostic Ablations for Joint Customer-Base Forecasting}
\label{tab:mechanism_improvements}
\begin{threeparttable}
\fontsize{9}{10.8}\selectfont
\setlength{\tabcolsep}{3.1pt}
\renewcommand{\arraystretch}{1.10}
\begin{tabular*}{0.99\textwidth}{@{\extracolsep{\fill}}lccrrrrr@{}}
\toprule
Model family or contrast
& \makecell{Shared\\repr.}
& \makecell{Revenue\\align.}
& Acq.
& \makecell{AOV\\weighted}
& \makecell{ROPC\\weighted}
& \makecell{Cohort-week\\Sales}
& \makecell{Total\\Sales} \\
\midrule
\multicolumn{8}{l}{\textit{Panel A: Mean SMAPE by model family}} \\
Single-task & No & No & 35.69 & 34.93 & 37.33 & 57.67 & 20.54 \\
Joint & Yes & No & 21.11 & 36.95 & 37.55 & 55.49 & 18.57 \\
Full & Yes & Yes & 19.30 & 35.94 & 35.56 & 53.91 & 15.48 \\
\addlinespace
\multicolumn{8}{l}{\textit{Panel B: Improvement relative to baseline}} \\
Joint vs. single-task & No$\rightarrow$Yes & No & +40.84\% & -5.76\% & -0.58\% & +3.77\% & +9.56\% \\
Full vs. joint & Yes & No$\rightarrow$Yes & +8.61\% & +2.73\% & +5.30\% & +2.85\% & +16.67\% \\
Full vs. single-task & No$\rightarrow$Yes & No$\rightarrow$Yes & +45.93\% & -2.87\% & +4.75\% & +6.52\% & +24.64\% \\
\bottomrule
\end{tabular*}
\begin{tablenotes}[flushleft]
\scriptsize
\item \textit{Note:} Panel A reports mean SMAPE across 966 companies. Panel B reports $100(\mathrm{SMAPE}_{B}-\mathrm{SMAPE}_{A})/\mathrm{SMAPE}_{B}$ for the variant $A$ and baseline $B$ named in each contrast, so positive values favor the first model. AOV and ROPC use cohort-size-weighted SMAPE. Full includes shared representation and revenue alignment; Joint includes shared representation only; Single-task fits acquisition, AOV, and ROPC separately. Families use family-specific candidate grids and selection strategies.
\end{tablenotes}
\end{threeparttable}
\end{table}

The selected-family contrasts show a consistent downstream ordering. Relative to the selected single-task family, the selected joint family has 9.56\% lower total-sales SMAPE. The selected full family in turn has 16.67\% lower total-sales SMAPE than the selected joint family and 24.64\% lower total-sales SMAPE than the selected single-task family. These descriptive differences are consistent with value from shared representation and revenue alignment, but family-specific tuning and selection mean that the percentages should not be interpreted as causal component effects.

The selected-family comparison documents a downstream performance ordering, but it does not by itself establish why the joint and full families perform better. We therefore turn next to a more targeted diagnostic implication of the shared-driver account: joint forecasting should help more when the customer-base primitives exhibit stronger shared temporal structure during the calibration period.

\subsection{Diagnostic Evidence on Shared-Driver Signal Extraction}
\label{subsec:shared_driver_test}

We next ask whether CBMT tends to help more in the settings where shared-driver signal extraction should be most valuable. If acquisition, repeat purchasing, and AOV contain overlapping information about common business conditions, then firms whose customer-base primitives exhibit stronger systematic co-variation should receive greater incremental benefit from joint forecasting. The analysis is diagnostic rather than causal. It does not identify the underlying business shock, but it tests whether the value of joint forecasting is larger when the observed primitives contain more shared structure.

The test constructs a firm-level measure of cross-metric co-movement. For each firm, we compute weekly aggregated trajectories of the three upstream primitives over the calibration window, apply a \(\log(1+x)\) transformation, and measure the extent to which the transformed trajectories move together over time.
To avoid relying on any single dependence metric, we construct a broad Cross-Metric Co-Movement Score. The score is the first principal component of a battery of 34 implemented dependence measures, drawn from 28 conceptual families, that capture linear and rank correlation, lead-lag co-movement, nonlinear dependence, predictive dependence, common-factor structure, and cohort-panel dependence. Intuitively, high values of this score indicate that acquisition, ROPC, and AOV contain more shared temporal structure. This need not mean that the primitives always rise and fall together. A promotion, for example, may increase acquisition and repeat orders while lowering AOV through discounting; a pricing change may raise AOV while weakening new-customer acquisition. The relevant feature is systematic co-variation that reveals a common business environment.

Our dependent variable is $ \text{WinJoint}_i = \mathbbm{1}\left[ \text{SMAPE}_{\text{Joint},i} < \text{SMAPE}_{\text{SingleTask},i} \right], $ where $\text{WinJoint}_i=1$ if the full joint model produces lower holdout sales SMAPE than the corresponding single-task baseline for firm \(i\). The binary outcome is less sensitive to extreme percentage errors than a ratio measure, which is useful in a sample containing firms with very different revenue scales. Because $\text{WinJoint}_i=1$ for 718 of 966 firms (74.3\%), the analysis asks whether calibration-period co-movement is associated with where the full joint model outperforms its single-task counterpart.

We estimate the association between calibration-period co-movement and
holdout-period joint-model wins using a controlled linear-probability
specification:
\begin{align*}
    \text{WinJoint}_i ={}&
    \alpha + \beta \text{CoMovement}_i
    + \gamma_1 \text{HistoricalScale}_i \\
    &+ \gamma_2 \text{HistoricalDispersion}_i
    + \delta_{\text{Industry}(i)} + \epsilon_i.
\end{align*}
The co-movement score is expressed in standard-deviation units. Historical
scale is the log of one plus mean positive annual sales over 2016--2018, and
historical dispersion is the coefficient of variation across those annual
sales totals. Both are measured before the April 2019 holdout begins. Web
Appendix \ref{wa:why_mtl} reports baseline and controlled linear-probability
models, a controlled logit, and a permutation placebo that randomly reassigns
firm-level co-movement scores across firms.

\begin{figure}[!htbp]
\centering
\caption{Joint-Model Win Rate by Cross-Metric Co-Movement Tercile}
\label{fig:dependence_tercile}
\includegraphics[width=0.45\linewidth]{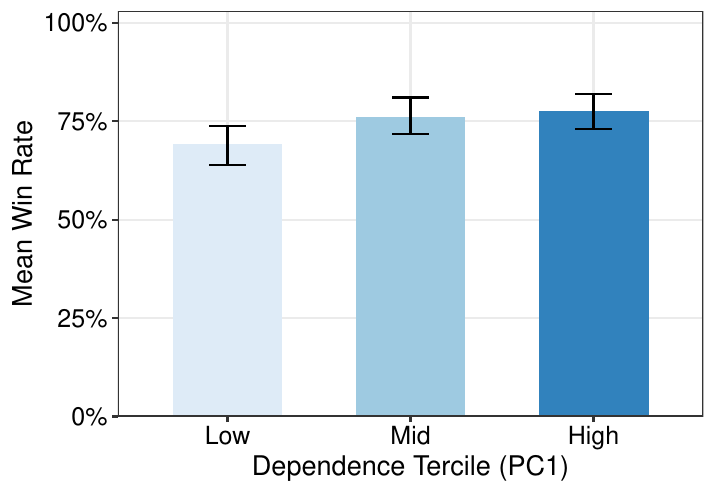}
\begin{minipage}{0.8\linewidth}
\vspace{0.5em}
\footnotesize \textit{Note.} Firms are divided into terciles based on the calibration-period Cross-Metric Co-Movement Score. Bars report the average value of \(\text{WinJoint}\), where \(\text{WinJoint}=1\) if the joint model has lower holdout sales SMAPE than the single-task baseline. Error bars show bootstrapped 95\% confidence intervals.
\end{minipage}
\end{figure}

Figure~\ref{fig:dependence_tercile} presents the main result in its simplest
form. Firms are divided into low, medium, and high terciles based on the
Cross-Metric Co-Movement Score, and the joint-model win rate rises monotonically
across the three groups. Firms in the high co-movement tercile are 8.39
percentage points more likely to exhibit a joint-model win than firms in the
low co-movement tercile in a two-sided comparison (\(p=.016\)). The regression
evidence leads to the same conclusion. In the controlled linear-probability
specification, a one-standard-deviation increase in co-movement is associated
with a 5.03 percentage point increase in joint-model win probability
(\(\beta=.050, p=.002\)). The score is measured before the holdout period, but
the analysis remains an exploratory diagnostic rather than a causal test. It
does not provide a rule for abandoning joint forecasting in low-co-movement
settings. Rather, it identifies where the incremental benefit of joint
forecasting is especially likely to arise.

Web Appendix \ref{wa:why_mtl} provides additional checks. The association is
not driven by one specialized dependence measure, and the appendix reports
logistic and permutation checks. Taken together, these results support the
narrower conclusion that observed cross-metric co-movement is associated with
where joint forecasting performs better. They do not identify the underlying
business shocks or establish a causal mechanism.

\subsection{Diagnostic Evidence on Revenue-Alignment Discipline}
\label{subsec:revenue_alignment_evidence}

The second contrast in Table~\ref{tab:mechanism_improvements} compares the scenario-3 selected full and joint families. The selected full family has 8.61\% lower acquisition SMAPE, 2.73\% lower AOV SMAPE, 5.30\% lower ROPC SMAPE, 2.85\% lower cohort-week sales SMAPE, and 16.67\% lower total-sales SMAPE. This ordering is consistent with revenue-alignment discipline improving the joint configuration of primitive forecasts when they are combined into sales. Because the two families use family-specific candidate grids and selection strategies, however, the contrast is diagnostic and cannot be interpreted as the causal effect of adding the regularizer while holding all else fixed.

The analysis that follows addresses a separate managerial question. The co-movement diagnostic explains when joint forecasting is more likely to add value relative to separate primitive models. The Predicted Holdout Error analysis instead identifies when forecasting is easier or harder in absolute terms. Keeping these questions distinct helps managers determine both whether joint forecasting is worth deploying and how much confidence to place in the resulting forecasts.

\section{Forecasting Difficulty and Contextual Drivers}
\label{sec:difficulty}

The preceding analyses examine where the incremental gains from joint forecasting are more likely to arise, in effect the boundary conditions of its value. We now turn to a distinct managerial question. Before relying on a forecast, can practitioners anticipate whether a firm's customer base will be relatively easy or difficult to predict? This question matters even when CBMT remains the preferred model. Managers may wish to update difficult forecasts more frequently, use more conservative planning ranges, place greater weight on scenario analysis, or subject high-stakes projections to additional review.

To address this question, we develop an ex-ante forecasting difficulty measure using observable customer-base characteristics available during the calibration period. We then examine the contextual drivers associated with absolute forecast accuracy. Performance also varies across the 25 industries in our sample (Web Appendix \ref{append:hetero_industry}), though we focus on structural drivers that generalize across industry contexts.

We construct 21 calibration-period features capturing scale, volatility, growth patterns, retention dynamics, and temporal regularity. We use gradient boosting regression (GBR) and elastic-net meta-models to relate these observable features to each forecasting model's realized holdout SMAPE. Across the outcome, forecasting-model, and meta-model combinations in Web Appendix Table~\ref{tab:cv_r2}, cross-validated $R^2$ ranges from $-0.009$ to $0.627$. The contextual features therefore explain substantial cross-firm variation in some settings but little in others; predicted difficulty should be treated as a diagnostic signal rather than a guarantee of forecast accuracy. Web Appendix~\ref{wa:FeatureImportance} provides feature definitions and additional results.

\subsection{Performance Across Difficulty Levels}

Using the GBR meta-model trained on CBMT's cohort-week sales holdout SMAPE, we generate out-of-fold predictions for each firm that serve as a Predicted Holdout Error (PHE), an ex-ante forecast difficulty measure computed using only calibration-period characteristics. We stratify firms into terciles based on PHE (Low, Medium, High) to examine whether the CBMT model maintains its performance advantage across varying levels of expected forecasting difficulty.

Table~\ref{tab:phe_bands} presents realized holdout SMAPE by PHE tercile for the CBMT model and three benchmark methods. As expected, absolute error increases monotonically across difficulty bands for all models, with close correspondence between actual and expected holdout error for the CBMT model, confirming that PHE serves as a meaningful proxy for forecasting difficulty for CBMT.

\begin{table}[htbp]
\centering
\caption{Average Holdout SMAPE by Predicted Difficulty Bands}
\label{tab:phe_bands}
\begin{threeparttable}
\begin{tabular}{lccccccc}
\toprule
& Avg. & \multicolumn{4}{c}{Holdout SMAPE} & Wins & \\
PHE Band & PHE & CBMT & LSTM & WG-PNBD-E & WG-PNBD & (CBMT) & N \\
\midrule
Low & 29.81 & \textbf{28.84} & 37.15 & 39.87 & 53.44 & 293 & 322 \\
 & & & (22.3\%) & (27.7\%) & (46.0\%) & & \\
Med & 50.71 & \textbf{49.40} & 56.47 & 58.67 & 63.80 & 251 & 322 \\
 & & & (12.5\%) & (15.8\%) & (22.6\%) & & \\
High & 80.84 & \textbf{83.48} & 89.56 & 94.87 & 96.09 & 190 & 322 \\
 & & & (6.8\%) & (12.0\%) & (13.1\%) & & \\
\bottomrule
\end{tabular}
\begin{tablenotes}[flushleft]
\footnotesize
\item \textit{Note:} Bands are equal-size terciles of Predicted Holdout Error (PHE) computed out-of-fold from the GBR meta-model, with deterministic first-rank handling for exact PHE ties. Avg. PHE shows the mean predicted holdout error within each band. Holdout SMAPE reports actual holdout error. Percentages in parentheses indicate relative SMAPE reduction of the CBMT model compared to each benchmark. Wins (CBMT) reports the number of firms for which the CBMT model achieved the lowest SMAPE among the four models. N indicates total firms in each band. CBMT uses the scenario-3 Full results, and LSTM uses the fixed rerun snapshot.
\end{tablenotes}
\end{threeparttable}
\end{table}

CBMT has the lowest average SMAPE in all three difficulty bands, while its relative advantage narrows as predicted difficulty rises. In the low-difficulty tercile, CBMT reduces average SMAPE by 22.3\% relative to LSTM, 27.7\% relative to WG-PNBD-E, and 46.0\% relative to WG-PNBD, and it ranks first for 293 of 322 firms (91.0\%). In the medium tercile, the corresponding reductions are 12.5\%, 15.8\%, and 22.6\%, with CBMT ranking first for 251 firms (78.0\%). In the high-difficulty tercile, the reductions narrow to 6.8\%, 12.0\%, and 13.1\%, and CBMT ranks first for 190 firms (59.0\%). Thus, PHE identifies settings in which all methods face higher error and CBMT's relative margin is smaller, without overturning its average advantage in any tercile.

The practical utility of PHE extends beyond retrospective performance analysis. Because the measure relies solely on calibration-period data, practitioners can estimate expected forecasting difficulty before relying on the resulting projections. Firms with low predicted error may use the forecasts for routine planning with standard monitoring. Firms with higher predicted error may update forecasts more frequently, use more conservative internal planning ranges, place greater weight on scenario analysis, and escalate high-stakes projections for managerial review. PHE should therefore be interpreted as a forecast-governance aid, not as a rule for discarding difficult forecasts altogether.

\subsection{Drivers of Forecast Accuracy}

To identify which firm characteristics most strongly influence forecast accuracy, we examine SHAP (SHapley Additive exPlanations; \cite{lundberg2017unified}) values from the GBR meta-model. SHAP values decompose each firm's predicted SMAPE into feature-level contributions, quantifying the marginal effect of each contextual factor while holding others constant. Web Appendix Figure~\ref{fig:top_features_GBR} presents the full permutation importance analysis across all 21 contextual features. Two features are substantially more influential than the rest: Total Orders (scale) and Cohort Sales CV (volatility) together account for most of the model's explanatory power.

\begin{figure}[!htbp]
\centering
\caption{SHAP Dependence Plots for Firm Scale and Cohort-Sales Volatility}\label{fig:shap_dependence}
\begin{minipage}{0.8\textwidth}
\begin{minipage}{0.48\linewidth}
    \centering
    \includegraphics[width=\linewidth]{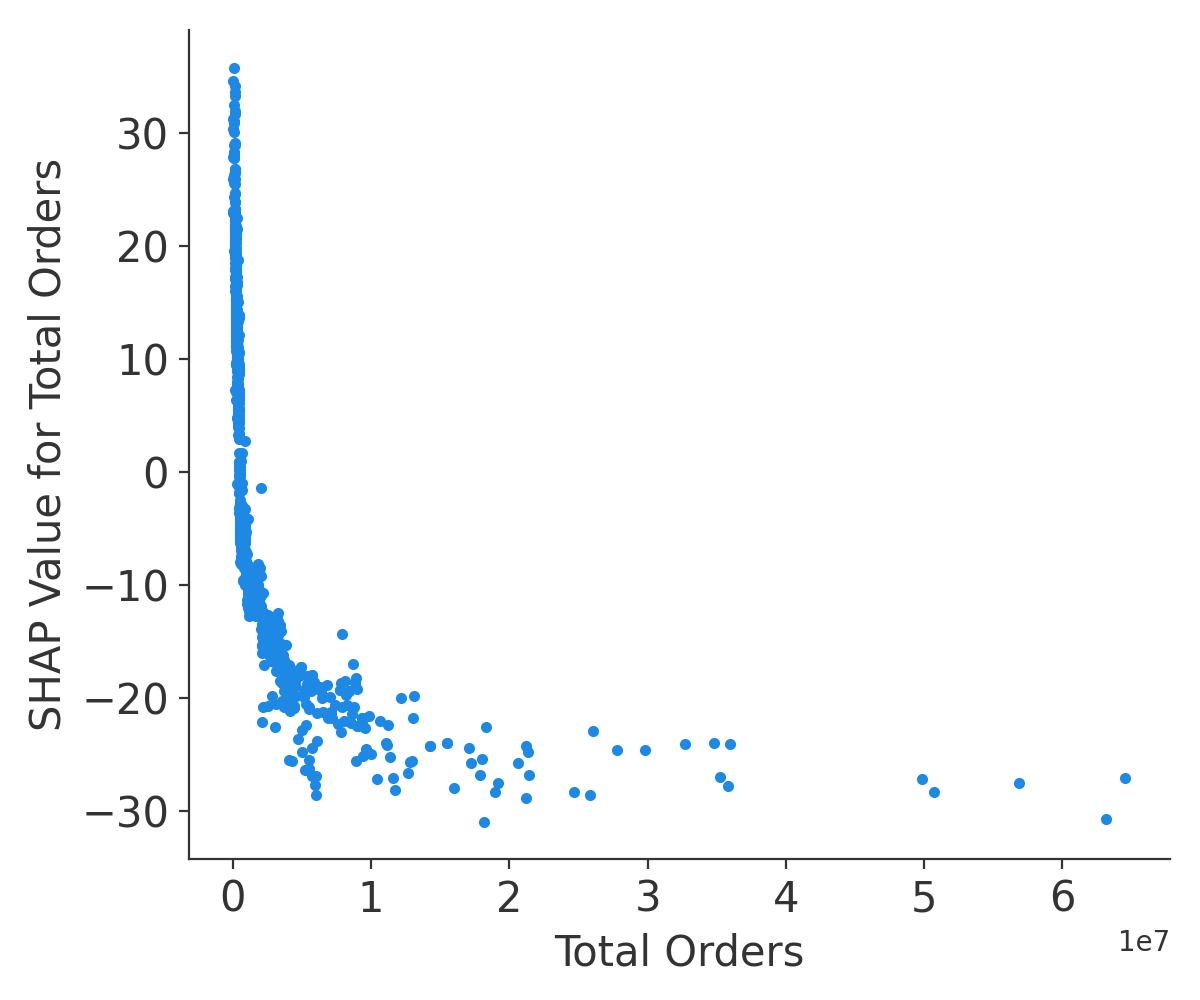}
\end{minipage}
\hfill
\begin{minipage}{0.48\linewidth}
    \centering
    \includegraphics[width=\linewidth]{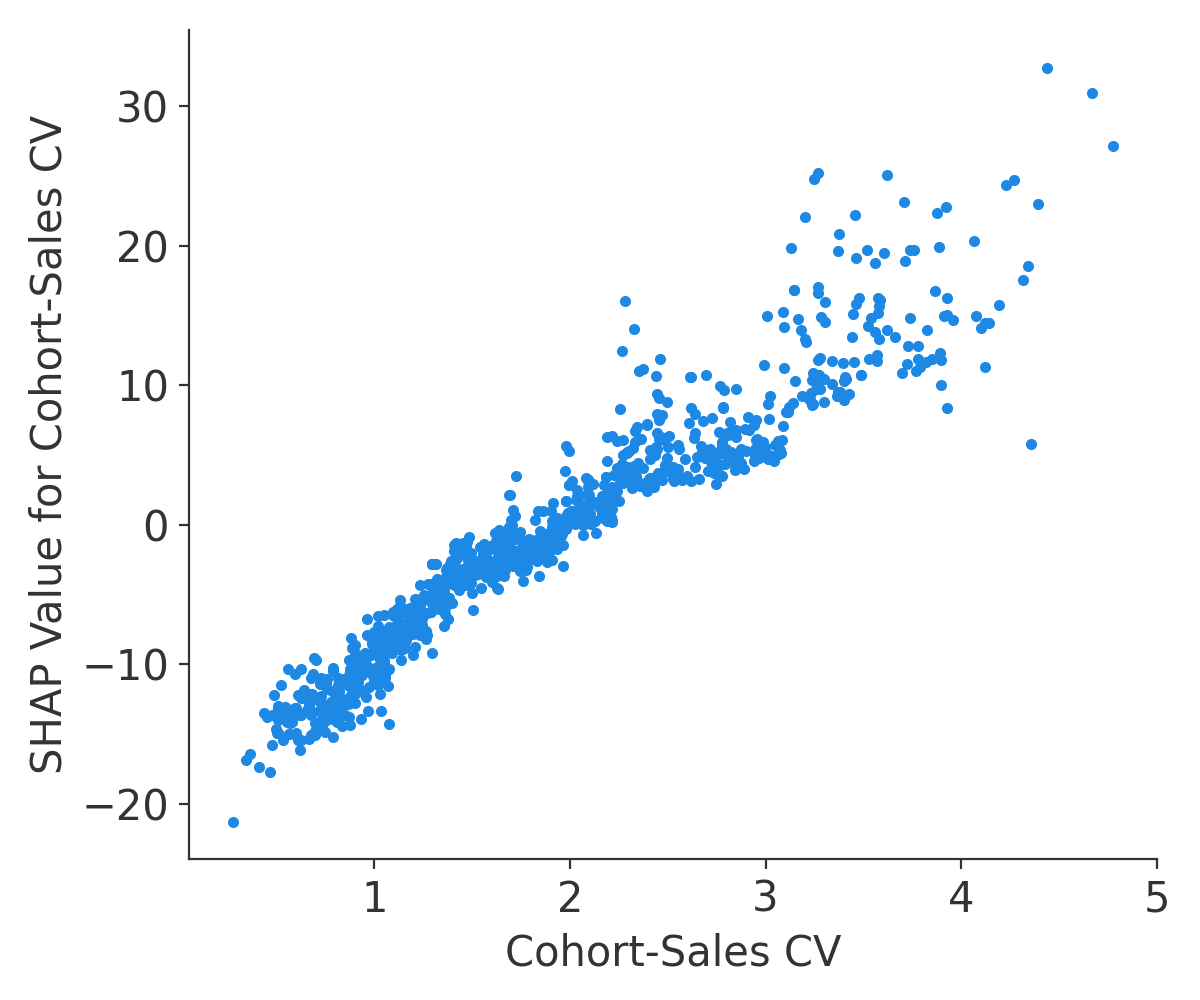}
\end{minipage}
\end{minipage}
\vspace{2mm} 
\footnotesize
\begin{minipage}{\textwidth}{\textit{Note:} These SHAP dependence plots come from the gradient boosting regression (GBR) model of CBMT's cohort-week sales SMAPE. Each point is a company. The horizontal axis reports the observed contextual feature; the vertical axis reports that feature's SHAP contribution to predicted SMAPE, holding the other features constant.}\end{minipage}
\end{figure}

Figure~\ref{fig:shap_dependence} presents SHAP dependence plots for these two dominant features. The left panel reveals a strong negative relationship between firm scale and forecast difficulty. Larger firms, as measured by cumulative order volume, exhibit lower predicted SMAPE because their transaction histories provide more stable signal for learning.

The right panel demonstrates that volatility exerts the opposite effect. Companies with greater week-to-week variation in cohort-level sales, as measured by the coefficient of variation, exhibit substantially higher predicted SMAPE. This is not a limitation unique to CBMT: all models exhibit elevated error rates in high-volatility settings. Volatility therefore signals an intrinsically difficult forecasting environment.

The CBMT model's relative performance also varies systematically with these two structural drivers. We stratify firms into a $3 \times 3$ grid defined by terciles of Total Orders and Cohort Sales CV (Web Appendix Figure~\ref{fig:winrate_tercile_2drivers}). CBMT is the modal winner in all nine cells, with particularly strong performance in low-volatility contexts across all scale levels. Its relative advantage narrows as volatility increases. In the high-volatility, high-scale cell, CBMT remains the most frequent winner and LSTM is the second most frequent, although that cell contains only 24 firms. This pattern yields a practical model-governance rule. In especially volatile customer-base environments, practitioners can retain an LSTM challenger, compare the resulting projections, and investigate material disagreements before using forecasts for high-stakes planning or guidance. Across the segments we examine, classical probabilistic benchmarks rarely achieve first rank, although they may remain useful as transparent reference points. For additional details regarding this analysis, see Web Appendix~\ref{wa:FeatureImportance}.

\subsection{Forecast Accuracy Across Horizons}

\begin{figure}[!htbp]
    \centering
    \caption{Forecast Accuracy Across Projection Horizons}
    \label{fig:hetero_horizon}
    \includegraphics[width=0.9\textwidth]{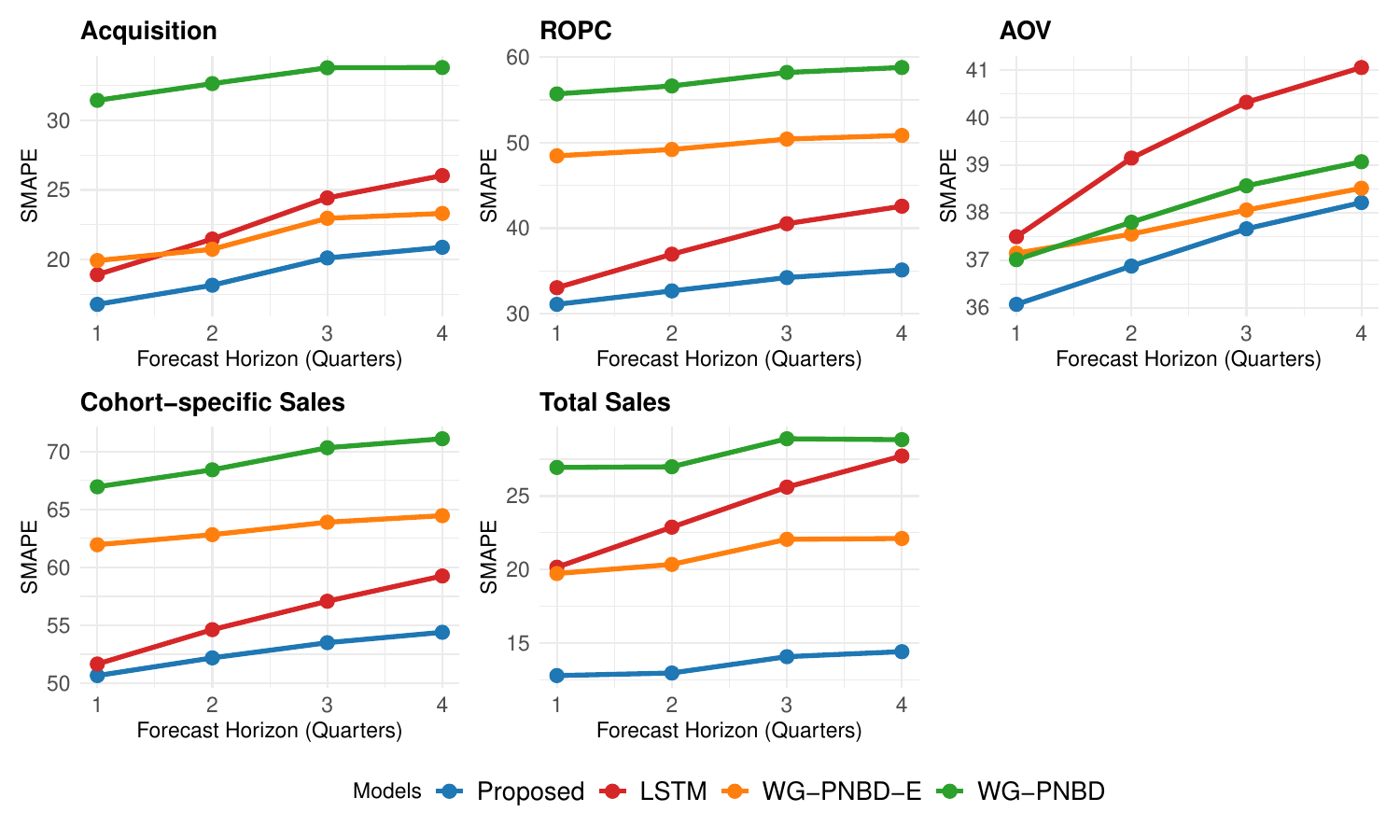}
    \footnotesize
    \begin{minipage}{\textwidth}{\textit{Note:} The horizontal axis denotes the projection horizon: one quarter ahead (April--June 2019), two quarters ahead (April--September 2019), three quarters ahead (April--December 2019), and the full holdout (April 2019--February 2020). The vertical axis reports average SMAPE across 966 companies. The upper panels show acquisition, ROPC, and AOV; the lower panels show cohort-week and total sales. Model-specific lines are identified in the legend.}\end{minipage}
\end{figure}

We examine how prediction accuracy degrades as the forecast horizon extends from one to four quarters ahead.\footnote{The holdout period spans 11 months (April 2019 through February 2020), so the fourth quarterly horizon contains the final two months of the holdout rather than a complete quarter.} Figure~\ref{fig:hetero_horizon} shows that average SMAPE increases monotonically with horizon length across all customer behavioral processes and revenue outcomes. This is expected, as uncertainty compounds as predictions extend further into the future.
The CBMT model maintains its accuracy advantage across all horizons, though the magnitude of improvement varies by behavioral process. For customer acquisition and repeat purchase predictions, the model's relative advantage remains fairly stable as horizons extend. However, the LSTM benchmark shows notably steeper degradation at longer horizons, suggesting that CBMT may be particularly effective at capturing longer-range dependencies in customer behavioral sequences.

\subsection{Model Selection for Deployment}
\label{subsec:deployment}

The heterogeneity documented in this section and in the mechanism analysis raises a practical question. If joint forecasting helps more for some firms than for others, are there firms for which a manager, using only information available through the calibration period, would be better off forecasting each primitive with a separate single-task model? We study this as a cross-firm out-of-sample routing problem in which the decision criterion is forecasting-holdout total-sales SMAPE. The logistic specification uses a Cross-Metric Co-Movement Score, log total orders, and cohort-sales volatility, all measured through March 31, 2019. The decision tree and gradient-boosted classifier also use the 34 individual dependence measures underlying the co-movement score. We split firms 70/30 into policy-training and held-out sets, stratified by industry. Within every inner and outer split, missing-value replacement, standardization, and principal-component estimation use only the corresponding training firms. We select each model's routing threshold by five-fold cross-validation within the policy-training firms, then evaluate the resulting policy once on the held-out firms. Web Appendix~\ref{wa:deployment} reports the full procedure and results.

No tested routing rule improves on always deploying CBMT on average. Across all 966 firms, mean total-sales SMAPE is 20.54 for always single-task, 15.48 for always CBMT, and 14.79 for an oracle that selects the better model for each firm. CBMT wins for 718 firms and single-task forecasting for 248, leaving a 0.69-point gap between always CBMT and the infeasible oracle. Across 100 re-randomized 70/30 splits, mean held-out SMAPE is 15.441 for always CBMT, compared with 15.448 for logistic routing, 15.444 for decision-tree routing, and 15.457 for gradient-boosted routing. The three learned rules route an average of only 0.24, 0.34, and 0.66 of the 290 held-out firms, respectively, to single-task forecasting, with a median of zero for each rule; they beat always CBMT in 2, 0, and 4 of the 100 splits. Thus, although single-task forecasting wins ex post for a substantial minority of firms, the calibration-period observables do not identify them reliably enough to improve mean out-of-sample accuracy. Managers should default to CBMT when choosing between these two models, while using the mechanism and forecasting-difficulty analyses to calibrate expectations and governance rather than as a validated switching rule.

\section{Conclusion}
\label{discussion}

The analyses document when jointly forecasting customer-base primitives is associated with better downstream revenue projections, show how accurately those forecasts recover the sources of a sales change, and identify settings in which the forecasts warrant greater caution. We close by drawing out what this coordinated customer-base view offers managers, how it informs forecast governance and resource allocation, how the resulting error magnitudes compare with a familiar external yardstick, how it bears on the financing and valuation of customer relationships, and the scope conditions that bound the evidence.

The managerial value of joint customer-base forecasting extends beyond improving aggregate sales projections. A topline forecast indicates how much revenue is expected to change but not where the change will come from. The sales bridge answers that question, attributing the change to repeat-order volume, repeat spend per order, and net customer-base replenishment. CBMT has lower source MAE in 23 of 24 benchmark-by-outcome comparisons, and the one benchmark edge is not statistically distinguishable from zero.

The relevant managerial response differs across these patterns. When acquisition forecasts weaken while repeat purchasing and spending remain stable, managers may prioritize acquisition budgeting, channel allocation, campaign timing, or distribution. When ROPC declines, the more relevant questions concern retention, reactivation, onboarding, and customer engagement. When AOV weakens, managers may instead revisit pricing, promotional depth, product mix, or merchandising. These conditions also differ in their implications for future revenue. An acquisition shortfall propagates into later periods through smaller cohorts, while order-value weakness often reflects pricing and promotion levers the firm can at least partially influence. The decomposition is also useful when primitives move in offsetting directions. Rising acquisition accompanied by weaker repeat purchasing or spending can flag questions about the quality and durability of customer growth even when near-term aggregate sales remain stable. When several primitives weaken together, managers may place greater weight on scenario analysis, demand planning, and conservative revenue guidance. Cohort-level forecasts therefore provide a structured input into customer-base unit economics, cohort financing, and valuation as well as marketing resource allocation.

Our forecasting-difficulty analyses identify settings in which customer-base projections should be used with greater caution. Two observable characteristics explain much of the variation in absolute forecast accuracy: scale and volatility. Larger customer bases provide more stable signals, while highly volatile customer bases are more difficult for all models to predict. CBMT is the modal winner in all nine scale-by-volatility bands, but its relative advantage narrows as volatility rises; LSTM is the second most frequent winner in the 24-firm high-scale, high-volatility cell. Practitioners in these environments may therefore pair CBMT with challenger-model comparisons, more frequent updating, and more conservative planning ranges. We also examined whether calibration-period characteristics could flag firms for which switching to single-task forecasting would help. Although single-task forecasting wins ex post for 248 of 966 firms, the logistic, decision-tree, and gradient-boosted routing rules all have higher mean held-out total-sales SMAPE than always CBMT across 100 re-randomized splits. The heterogeneity is therefore best used to calibrate expectations and governance rather than as a validated switching rule.

To place these error magnitudes on a more familiar external yardstick, we compare CBMT's proportional forecast errors with sell-side analyst revenue-forecast errors for 223 publicly traded firms in our sample. The two forecasts target different revenue measures and draw on different information sets, so the exercise calibrates the scale of CBMT's errors rather than ranking the methods. Across the first three fiscal quarters following the forecast origin, CBMT's SMAPE against panel-observed transaction sales ranges from 4.6\% to 5.6\%, while analyst SMAPE against reported firm revenue ranges from 6.4\% to 7.4\%. Forecasts built from structured transaction histories alone thus operate at proportional error levels comparable to those managers and investors already work with, which supports their use as one input to planning and marketing--finance discussions. Web Appendix~\ref{wa:AnalystBenchmark} reports the matching procedure, full results, and scaling assumptions.

While we apply the multi-task learning framework in the context of customer base analysis and customer-based corporate valuation, the approach may extend to other marketing domains characterized by hierarchical dependencies between upstream behaviors and downstream outcomes. Marketing researchers frequently model interdependent processes that collectively determine aggregate performance, including email campaigns (send volumes, open rates, conversions), video engagement (viewers, viewing frequency, watch time), and multi-category purchasing. These contexts share the same structure: upstream behaviors are interdependent, temporal dynamics vary across processes, and accurate downstream predictions benefit from structured joint modeling. The same ingredients carry over: a shared representation captures cross-process regularities, task-specific heads preserve specialization, and an alignment regularizer ties upstream forecasts to the downstream outcome.

The modular architecture facilitates practical deployment across diverse organizational contexts. Marketing teams can substitute domain-specific models for individual behavioral processes while retaining the multi-task structure---for instance, incorporating a more sophisticated model for customer purchasing or a simpler model for AOV. Likewise, as new time-series models emerge, they can be integrated without restructuring the framework. The framework accommodates both deterministic and stochastic covariates, enabling integration of external signals (competitive actions, macroeconomic indicators) alongside internal signals. Together these choices let the framework adapt to each implementation context---a firm's business model, data granularity, and operating context---without building a new model from scratch for each one.

Beyond methodological considerations, our findings have implications for corporate finance, where customer relationships increasingly serve as collateralizable assets. A growing ecosystem of specialty finance firms, including General Catalyst's Customer Value Fund \citep{reuters2025grammarly} and platforms like Pipe \citep{ft2025pipe}, now advance capital directly against projected customer cohort revenues, addressing a fundamental timing mismatch between customer acquisition investments and their eventual payoffs. Forecast accuracy is relevant in this context because uncertainty about cohort revenues may affect how lenders and managers evaluate acquisition investments. More accurate cohort-level revenue projections could therefore inform financing and acquisition-budget discussions, although quantifying effects on financing terms or investment decisions lies outside the scope of this study.

Several scope conditions are important for interpreting and implementing the results. First, the forecasts are evaluated against transaction-panel sales rather than population-level firm revenue. This provides a consistent environment for comparing methods, while firm-specific deployment would require attention to panel coverage, representativeness, and scaling. Second, our April 2019 to February 2020 holdout period ends before the COVID-19 disruption. The results should therefore be interpreted as evidence from a pre-COVID forecasting environment rather than as a stress test under abrupt regime change. Future research should examine whether joint customer-base forecasting remains advantageous in such settings. Third, CBMT forecasts cohort-level rather than individual-level outcomes. This level of aggregation reduces sparsity and aligns naturally with revenue guidance, cohort-based unit economics, and resource allocation, but it is not designed for individualized targeting or treatment assignment. Finally, implementation requires firm-specific choices about data construction, updating frequency, and model governance.

Two extensions follow naturally from the framework's structure. We train a separate model for each firm rather than meta-learning across companies, because cross-firm learning offered little empirical benefit given the richness of within-firm data and the heterogeneity of business models and customer bases; for nascent companies or new entrants that lack extensive transaction histories, cross-company transfer may prove more valuable. The framework could likewise be extended to model competitive dynamics, in which one firm's customer behaviors shape rivals' outcomes. We have prioritized methods that firms can run on their own transaction data, or that investors can apply to a single company's metrics, rather than approaches that require industry-wide data rarely available in practice. Both directions are promising where the necessary data can be obtained, and we leave them to future research.

In conclusion, this study provides large-scale evidence on when jointly forecasting customer-base primitives improves downstream revenue projections, how accurately the forecasts recover the sources of a sales change, and when the forecasts warrant greater caution. The scenario-3 selected-family comparisons and co-movement results are consistent with shared temporal information and revenue alignment contributing to forecast performance, while remaining diagnostic rather than causal. The resulting forecasts give managers both a revenue projection and an auditable view of existing-base repeat-order volume, repeat spend per order, and net customer-base replenishment, while the difficulty estimates indicate when those forecasts can anchor routine planning and when they call for closer governance. Customer-base forecasting can thus serve as a disciplined diagnostic for the acquisition, retention, pricing, planning, financing, and valuation decisions that turn on where revenue is headed.


\clearpage

\newpage
\printbibliography[
title={REFERENCES}
]

\clearpage
\theendnotes





 

\makeatletter
\renewcommand{\thesection}{\Alph{section}}
\renewcommand{\thesubsection}{\thesection.\arabic{subsection}}
\renewcommand{\theHsection}{WA.\Alph{section}}
\renewcommand{\theHsubsection}{\theHsection.\arabic{subsection}}
\renewcommand{\theHfigure}{WA.\arabic{figure}}
\renewcommand{\theHtable}{WA.\arabic{table}}

\renewcommand{\@seccntformat}[1]{\csname the#1\endcsname.\quad}
\makeatother

\makeatletter
\def\WA@ext{watoc}

\newcommand{\WAstarttoc}{%
  \let\WA@orig@addcontentsline\addcontentsline
  \renewcommand{\addcontentsline}[3]{%
    \def\WA@file{##1}\def\WA@toc{toc}%
    \ifx\WA@file\WA@toc
      \WA@orig@addcontentsline{\WA@ext}{##2}{##3}%
    \else
      \WA@orig@addcontentsline{##1}{##2}{##3}%
    \fi
  }%
}

\newcommand{\WAstoptoc}{%
  \let\addcontentsline\WA@orig@addcontentsline
}

\newcommand{\WAtableofcontents}{%
  \@starttoc{\WA@ext}%
}
\makeatother

\renewcommand{\thefigure}{W\arabic{figure}}
\renewcommand{\thesubfigure}{\alph{subfigure}}
\renewcommand{\thetable}{W\arabic{table}}

\clearpage
\thispagestyle{empty}

\begin{center}
  \Large\bfseries
  Forecasting Revenue with Its Customer-Base Drivers: When and Why Coordination Helps\\[6pt]
  \large Web Appendix
\end{center}

\vspace{1em}

\begin{center}
  Kyeongbin Kim \quad\textbullet\quad Daniel M.~McCarthy \quad\textbullet\quad Dokyun Lee
\end{center}

\footnotetext[1]{This appendix provides supplementary material for the paper.}

\hrule

\setcounter{section}{0}
\setcounter{table}{0}
\setcounter{figure}{0}
\setcounter{page}{1}

\begin{singlespace}
\begingroup
\setcounter{tocdepth}{2}
\renewcommand\contentsname{Table of Contents}
\hypersetup{linktoc=all}
\WAtableofcontents               
\endgroup
\clearpage
\end{singlespace}

\clearpage

\WAstarttoc   

\section{Illustrative Companies by Industry}\label{append:examplecompany}

The sample contains transaction data from 966 companies across 25 industries and 120 subcategories. The list below gives selected company--subcategory examples.

{\small
\begin{singlespace}
\setlength{\columnsep}{1.5em}
\begin{multicols}{2}
\setlength{\parindent}{0pt}
\setlength{\parskip}{0.35\baselineskip}
\raggedcolumns
\raggedright

\textbf{Apparel \& Accessories (109):}
Lululemon (Active \& Athleisure);
Stitch Fix (Apparel Subscriptions);
Carter's (Children's Apparel);
Forever 21 (Fast Fashion).

\textbf{Auto (20):}
Jiffy Lube (Auto Parts \& Services);
CarMax (Auto Sales).

\textbf{Charitable Giving (4):}
GoFundMe (Fundraising Platforms);
World Vision (Religious Organizations).

\textbf{Department Stores (21):}
Nordstrom Full Price (High-End Department Stores);
Macy's (Mid-Tier Department Stores);
T.J. Maxx (Off-Price Department Stores).

\textbf{Digital Services (15):}
Canva (Community Platforms);
McAfee (Internet Security).

\textbf{Electronics (4):}
Apple (Computers \& Tablets).

\textbf{Events \& Attractions (25):}
Eventbrite (Booking Platforms);
Topgolf (Indoor Entertainment Centers);
AMC (Movie Theaters);
Six Flags (Theme Parks).

\textbf{Finance (49):}
Robinhood (Broker-Dealers);
Progressive Leasing (Financing);
Western Union (Money Transfer);
Square (Payment Facilitators).

\textbf{Fitness (13):}
LA Fitness (Gyms);
Peloton (Home Fitness);
Orangetheory (Workout Classes).

\textbf{General Merchandise (89):}
Target (Big Box Retailers);
Dollar Tree (Discount \& Dollar Stores);
Shell (Gas Stations);
Office Depot (Office Supplies).

\textbf{Gig Economy Income (2):}
Uber Driver Pay (Driver Pay).

\textbf{Grocers (111):}
Aldi (Discount Grocers);
Hello Fresh (Meal Kits);
Instacart (Online Grocers);
Whole Foods Market (Supermarkets).

\textbf{Health \& Beauty (36):}
Bath \& Body Works (Body Care \& Fragrances);
GNC (Diet Plans \& Supplements);
Sunglass Hut (Eyewear);
Sephora (General Cosmetics).

\textbf{Healthcare \& Insurance (13):}
Aetna (Health Insurance);
State Farm (Home \& Auto Insurance).

\textbf{Hobbies \& Toys (10):}
Hobby Lobby (Arts \& Crafts);
Toys ``R'' Us (Toys).

\textbf{Home (52):}
Rural King (Garden \& Outdoor);
Wayfair (Home Furnishings);
Lowe's (Home Improvement);
ADT (Home Security);
Terminix (Home Services).

\textbf{Home Entertainment (63):}
Barnes \& Noble (Book Retailers);
Kindle (E-Books);
Twitch (Gaming);
Spotify (Music Streaming \& Audio);
HBO (Video Streaming).

\textbf{Occasion \& Gifts (14):}
Yankee Candle Company (Flowers \& Gift Baskets);
Shutterfly (Photos \& Printing);
Party City (Seasonal \& Occasion).

\textbf{Pets (9):}
Banfield (Pet Care);
PetSmart (Pet Supplies).

\textbf{Restaurants (192):}
First Watch (Casual Dining);
DoorDash (Delivery Aggregators);
Panera Bread (Fast Casual);
Domino's Pizza (QSR).

\textbf{Specialty (7):}
USPS (Shipping).

\textbf{Specialty Food \& Beverage (6):}
Total Wine \& More (Alcohol);
Nespresso (Beverages).

\textbf{Sporting Goods (15):}
Cabela's (Hunting \& Fishing);
Dick's Sporting Goods (Sports Gear).

\textbf{Telecommunication (22):}
Comcast (Cable Providers).

\textbf{Travel \& Transportation (65):}
American Airlines (Air Travel);
Airbnb (Lodging \& Accommodation);
Metrocard (Mass Transit);
Expedia (Online Travel Agency).

\end{multicols}
\end{singlespace}
}

\section{Detailed Specification of the Model Framework}\label{wa:model_details}

This appendix details the data construction, forecast granularity, model implementation, training objective, and walk-forward inference procedure for the joint customer-base prediction framework.

\subsection{Input Data Construction}\label{wa:input_data}

\paragraph{Cohort-week primitives.}
For each firm, calendar weeks are indexed by $t\in\mathcal T$. 
Acquisition cohorts are indexed by $i\in\mathcal I$, and $t_0(i)$ denotes the calendar week in which cohort $i$ enters the customer base. 
The size of cohort $i$ is denoted by $A_i$. 
Equivalently, if $A_t$ is the number of newly acquired customers in calendar week $t$, then $A_i=A_{t_0(i)}$ for the cohort born in week $t_0(i)$. 
For each cohort-week $(i,t)$ with $t\geq t_0(i)$, we observe repeat orders per customer, $R_{i,t}$, average order value, $V_{i,t}$, and cohort-week sales, $Y_{i,t}$. 
Orders per customer are defined as
$ O_{i,t}=\mathbbm{1}\{t=t_0(i)\}+R_{i,t}.$
Thus, cohort-week sales and aggregate weekly sales satisfy
\begin{equation}\label{eq:wa_sales_eq} 
Y_{i,t}=A_iO_{i,t}V_{i,t},
\qquad
S_t=\sum_{i\in\mathcal I_t}Y_{i,t},
\end{equation}
where $\mathcal I_t$ is the set of cohorts active in week $t$.

\paragraph{Prediction samples and look-back windows.}
For each cohort-conditioned sample $(i,t)$, the model receives a look-back window of length $p+1$. 
In the empirical implementation we set $p=19$, corresponding to a 20-week window. 
The input sample is
\begin{equation}\label{eq:wa_input_window}
D_{i,t}
=
\left\{
\left(A_\tau,R_{i,\tau},V_{i,\tau},S_\tau^{\mathrm{aux}},C_{i,\tau}\right)
:
\tau=t-p,\ldots,t
\right\},
\end{equation}
where $C_{i,\tau}$ denotes deterministic covariates known at forecast time. 
This notation makes explicit that acquisition and aggregate sales, $A_\tau$ and $S_\tau^{\mathrm{aux}}$, are calendar-week quantities, whereas ROPC and AOV, $R_{i,\tau}$ and $V_{i,\tau}$, are cohort-week quantities. 
In implementation, the week-level series are replicated across cohort-conditioned samples for a given calendar week so that all four outcomes can be used in a common multi-output prediction framework.

\paragraph{Deterministic covariates.}
The deterministic covariates $C_{i,\tau}$ capture calendar, cohort, and tenure information known at the time forecasts are made. 
These include week-of-year indicators to capture seasonality; linear and quadratic calendar trends to capture gradual shifts in the baseline; holiday indicators for recurring events such as Black Friday; acquisition-month indicators to capture cohort-entry timing; cohort identifiers for cohorts observed before the prediction period; and linear and quadratic tenure terms, where tenure is the number of weeks since acquisition. 
The tenure terms encode the prior knowledge that repeat purchasing commonly varies with cohort age, while the calendar variables allow the model to capture recurring patterns without requiring a longer historical window.

\paragraph{Stochastic covariates (Auxiliary aggregate-sales).}
Aggregate sales $S_\tau^{\mathrm{aux}}$ are included as an auxiliary historical information channel because total sales summarize the realized downstream implication of all active cohorts. 
Unlike deterministic covariates, future aggregate sales are not known at forecast time. 
The model therefore treats aggregate sales as a stochastic covariate that is forecast jointly and recursively. 
This auxiliary forecast is used only to update the aggregate-sales input channel during walk-forward inference; the focal revenue forecast remains the component-implied sales forecast constructed from predicted acquisition, ROPC, and AOV. 
Including this channel allows the shared representation to learn from both local cohort-level behavior and the global sales trajectory of the firm, while avoiding the use of future sales information during holdout forecasting.

\paragraph{Zero padding and cold-start cohorts.}
For cohorts that have not yet been acquired at a given historical week, ROPC and AOV histories are padded with zeros. 
In this setting, zero padding corresponds to no observed purchase activity before the cohort is born. 
This construction allows the model to learn transitions from inactivity to first purchase and then to repeat purchasing. 
For cohorts that enter during the holdout period, the model uses zero histories and deterministic covariates until the cohort's predicted acquisition week. 
The predicted acquisition in that week becomes the forecasted cohort size used to compute the cohort's downstream revenue contribution during walk-forward inference.

\subsection{Output Data and Forecast Granularity}\label{wa:output_data}

Each sample $D_{i,t}$ is used to form one-step-ahead forecasts for week $t+1$:
\begin{equation}\label{eq:wa_outputs}
\left(
\widehat A^{(i)}_{t+1},
\widehat R_{i,t+1},
\widehat V_{i,t+1},
\widehat S^{(i)}_{t+1}
\right)
\end{equation}
The outputs $\widehat R_{i,t+1}$ and $\widehat V_{i,t+1}$ are cohort-week forecasts. 
The outputs $\widehat A^{(i)}_{t+1}$ and $\widehat S^{\mathrm{aux} (i)}_{t+1}$ are calendar-week forecasts generated from cohort $i$'s input window. 
We call them cohort-conditioned forecasts because the prediction is conditioned on cohort $i$'s recent history, even though the target is a week-level outcome shared by all cohorts in that calendar week. The superscript $(i)$ therefore indexes the input window, not a cohort-specific acquisition or sales outcome.
Let $\mathcal J_t$ denote the set of cohort-conditioned samples used to generate forecasts for week $t+1$. 
When we ultimately generate a single week-level forecast for acquisition, we average the cohort-conditioned forecasts within calendar week:
$
\widehat A_{t+1}
=
\frac{1}{|\mathcal J_t|}
\sum_{i\in\mathcal J_t}
\widehat A^{(i)}_{t+1}. 
$

\paragraph{Mixed granularity of the acquisition target.}
The distinction between cohort-level and calendar-week quantities is central to our implementation. 
In the exact cohort-level accounting identity in Equation~\eqref{eq:wa_sales_eq}, the acquisition term is the historical cohort size $A_i=A_{t_0(i)}$. 
Once cohort $i$ has entered the customer base, $A_i$ is observed and fixed across its subsequent cohort-week history. 
Using $A_i$ as the acquisition series inside a cohort-specific 20-week sequence would therefore turn acquisition into a constant-prediction task rather than a forecast of future customer inflows. 
It would also fail to produce the calendar-week acquisition forecast, $A_{t+1}$, needed to initialize new cohorts during walk-forward inference. 
For this reason, acquisition is modeled as a calendar-week outcome, while ROPC and AOV are modeled as cohort-week outcomes. 
This mixed granularity motivates the revenue-alignment regularizer described in Web Appendix~\ref{wa:training}.

\subsection{Implementation Details of the Model}\label{wa:transformer}

\begin{figure}[!t]
    \centering
    \caption{Customer-Based Multi-task Transformer (CBMT) Architecture}
    \includegraphics[width=0.8\textwidth]{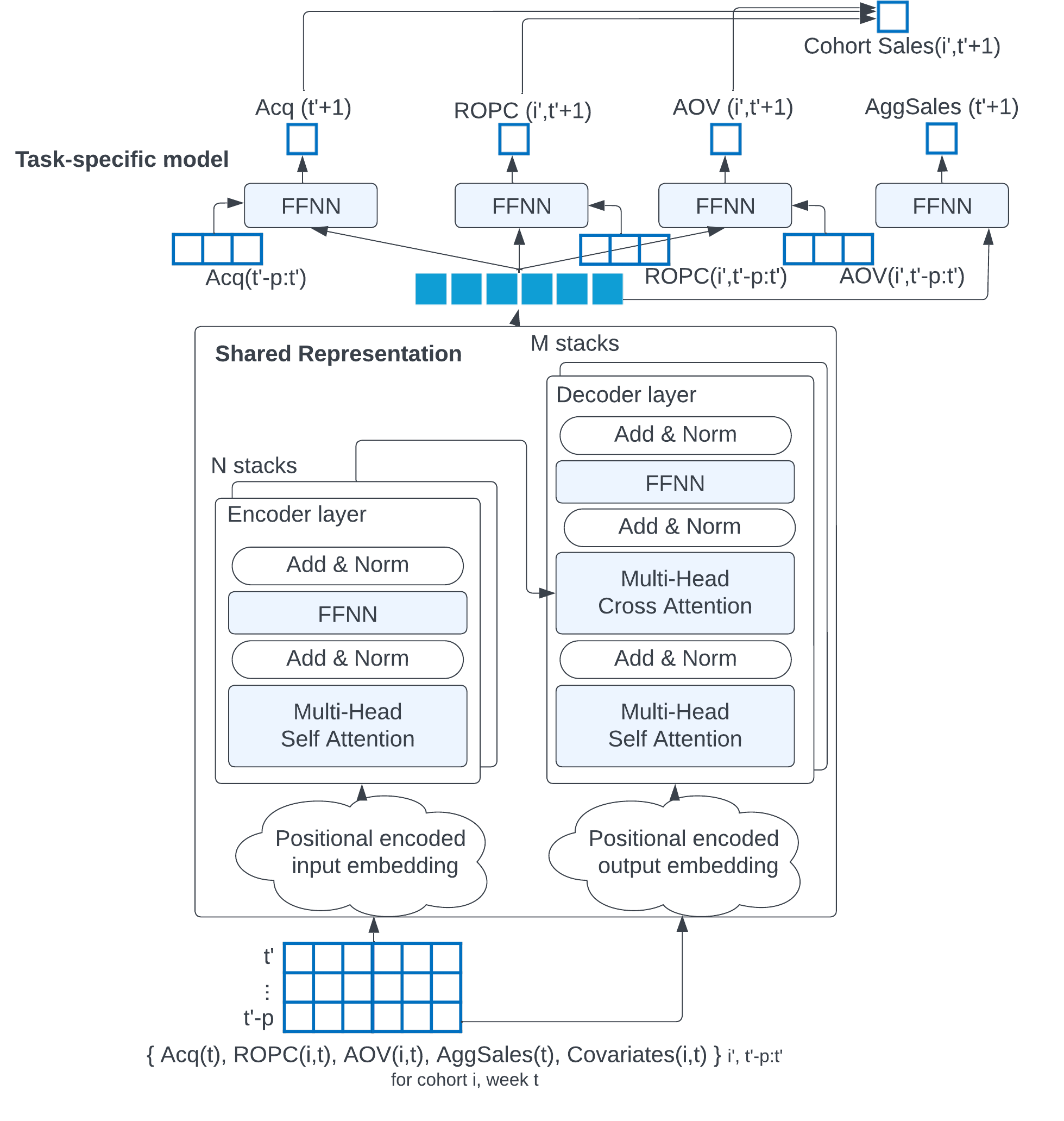}
    \label{fig:MTL}
    \begin{minipage}{\textwidth}
        \footnotesize \textit{Note:} The model employs a shared representation layer with task-specific heads for acquisition (Acq), repeat orders per customer (ROPC), average order value (AOV), and aggregate sales (AggSales). 
        The input window for cohort $i$ at week $t$ is scaled, embedded, and positionally encoded before being processed by the shared Transformer layer. 
        Task-specific heads transform the shared representation into one-step-ahead predictions. 
        During training, an auxiliary revenue-alignment objective links upstream primitive forecasts to downstream revenue, as described in Web Appendix~\ref{wa:training}.
    \end{minipage}
\end{figure}

This section describes the technical implementation of the shared representation and task-specific layers. 
The main paper presents these components at a higher level to emphasize the core forecasting problem: learning from multiple customer-base primitives while aligning component forecasts with downstream revenue. 
The detailed structure of the model is shown in Figure~\ref{fig:MTL}.

\paragraph{Input scaling and embeddings.}
The variables in $D_{i,t}$ differ substantially in scale. 
ROPC values are often less than one, AOV values are measured in currency units, and acquisition and sales can vary by several orders of magnitude across firms. 
We therefore apply feature-specific min-max normalization using parameters estimated on the calibration data. 
The same calibration-period scalers are used during validation and holdout forecasting. 
Categorical variables such as week-of-year, acquisition month, and cohort identifier are represented using entity embeddings. 
The embedding dimension for each categorical feature is set to the integer approximation of the square root of the number of levels, which provides a compact representation relative to high-dimensional dummy variables.

\paragraph{Shared layer.}
The scaled numerical variables and categorical embeddings are concatenated at each week of the look-back window and projected to a $d$-dimensional input representation. 
In our implementation, $d=32$, so each 20-week input window is represented as a $20\times 32$ matrix. 
We add sinusoidal positional encodings to preserve the order of observations in the window. 
Positional encodings allow the model to distinguish, for example, a repeat-purchase drop that occurred last week from one that occurred several months ago.

The shared layer follows the Transformer architecture. 
The central operation is scaled dot-product attention:
$\mathrm{Attention}(Q,K,V) = \mathrm{softmax}\left(\frac{QK^{\top}}{\sqrt{d_k}}\right)V$,
where $Q$, $K$, and $V$ are query, key, and value matrices and $d_k$ is the dimension of the key vectors. 
Intuitively, attention allows the model to learn which historical observations are most relevant for a particular forecast. 
For instance, a recent acquisition spike may be relevant for forecasting near-future repeat purchasing, while a recurring holiday week may be relevant for forecasting AOV.

We use multi-head attention with four heads. 
Each head can focus on different relationships within the same input window, such as within-behavior temporal persistence, contemporaneous co-movement across behaviors, or lagged relationships between acquisition and later repeat purchasing. 
The outputs of the attention heads are concatenated and passed through layer normalization, residual connections, and a feed-forward network with hidden size 64. 
We stack two encoder layers. 
The decoder receives the most recent week of the encoded input as its starting point and uses cross-attention over the encoder output to generate the final shared representation. 
The decoder output is a 32-dimensional vector $h_{i,t}$ that summarizes the recent customer-base history for cohort $i$ at week $t$.

Formally, the shared representation is
\begin{equation}\label{eq:wa_shared_representation}
h_{i,t}=f_{\psi}\left(e(D_{i,t})\right),
\end{equation}
where $e(\cdot)$ denotes the scaling, embedding, projection, and positional-encoding steps and $f_{\psi}$ denotes the Transformer backbone with parameters $\psi$.

\paragraph{Task-specific output layers.}
Each forecasting task has its own output layer. 
For task $k\in\{A,R,V,S\}$, the task-specific head maps the shared representation and the corresponding lagged history into a one-step-ahead forecast:
\begin{equation}\label{eq:wa_task_head}
\widehat x_{k,i,t+1}
=
g_{\phi_k}\left(h_{i,t},x_{k,i,t-p:t}\right).
\end{equation}
For notational convenience, $x_{A,i,\tau}=A_\tau$ and $x_{S,i,\tau}=S^{\mathrm{aux}}_\tau$ for week-level tasks, while $x_{R,i,\tau}=R_{i,\tau}$ and $x_{V,i,\tau}=V_{i,\tau}$ for cohort-week tasks. 
The task-specific heads are three-layer fully connected feed-forward neural networks with ReLU activations. 
To ease optimization, we reintroduce the corresponding lagged history $x_{k,i,t-p:t}$ into each head through a residual-style shortcut. 
This design allows the head to combine the deep shared representation with task-specific autoregressive information.

\subsection{Illustration of Dependency Learning in the Shared Layer}\label{append:dependencylearning}

The shared layer can capture three relationships common in customer-base data. These relationships motivate joint forecasting but are not imposed as fixed parametric restrictions.

First, the model can learn contemporaneous cross-behavior associations. A promotional event may simultaneously increase acquisition, repeat orders from existing customers, and the value of the orders placed. When such associations recur in the calibration data, the shared representation can use one behavior as a signal about the others.

Second, the model can learn within-behavior temporal associations. Repeat purchasing often changes with cohort tenure, and AOV may display persistence driven by category mix or customer composition. The attention mechanism can assign higher weight to historical observations that are informative for the same behavior in the next period.

Third, the model can learn lagged cross-behavior associations. A surge in acquisition may be followed several weeks later by changes in repeat purchasing as the newly acquired cohort matures. Similarly, high AOV in a particular period may signal product-mix or promotion conditions that affect future purchase frequency. Multi-head attention allows different heads to focus on different temporal and cross-behavioral patterns.

\subsection{Objective Function and Optimization Details}\label{wa:training}

\paragraph{Overall objective.}
Model parameters are estimated by minimizing a weighted objective with task-specific forecast losses and an auxiliary revenue-alignment loss:
\begin{equation}\label{eq:wa_loss}
\min_{\psi,\{\phi_k\}}
\;
\sum_{k\in\{A,R,V,S\}}
\omega_k L_k(\psi,\phi_k)
+
\omega_C L_C(\psi,\phi_A,\phi_R,\phi_V).
\end{equation}
For each task $k$, $L_k$ is the mean squared error for one-step-ahead prediction over a mini-batch $\mathcal B$:
$
L_k =
\frac{1}{|\mathcal B|}
\sum_{(i,t)\in\mathcal B}
\left[
s_k\left(\widehat x_{k,i,t+1}\right)
-
s_k\left(x_{k,i,t+1}\right)
\right]^2,
$
where $s_k(\cdot)$ is the calibration-period scaler for task $k$. 
The four task losses correspond to acquisition, ROPC, AOV, and aggregate sales. 
For the week-level tasks, the target is replicated across the cohort-conditioned samples associated with the same calendar week. 
Mini-batches sample cohort-week windows $(i,t)$ from the calibration period.

\paragraph{Auxiliary revenue-alignment regularizer.}
The auxiliary loss connects the upstream behavior forecasts to downstream revenue. 
Given the mixed granularity described in Web Appendix~\ref{wa:output_data}, this term is implemented as a soft alignment device rather than as a literal reconstruction of cohort-week sales. 
For each cohort-conditioned sample, define
\begin{equation}\label{eq:wa_aux_sales}
\widehat O_{i,t+1}
=
\mathbbm{1}\{t+1=t_0(i)\}
+
\widehat R_{i,t+1}; \quad
\widehat Y^{\mathrm{aux}}_{i,t+1}
=
\widehat A^{(i)}_{t+1}
\widehat O_{i,t+1}
\widehat V_{i,t+1}.
\end{equation}
We use $\widehat Y^{\mathrm{aux}}_{i,t+1}$, rather than $\widehat Y_{i,t+1}$, to emphasize that this is an auxiliary regularization object. 
The exact cohort-level accounting identity in Equation~\eqref{eq:wa_sales_eq} uses the historical cohort size $A_i$, whereas Equation~\eqref{eq:wa_aux_sales} uses a forecast of calendar-week acquisition, $\widehat A^{(i)}_{t+1}$. 
The term therefore links week-level acquisition, cohort-level repeat purchasing, cohort-level order value, and aggregate sales within a common training objective, but it is not a literal forecast of cohort-week sales.

The auxiliary revenue-alignment regularization loss is
\begin{equation}\label{eq:wa_coherence_loss}
L_C
=
\frac{1}{|\mathcal B|}
\sum_{(i,t)\in\mathcal B}
\left[
s_Y\left(\widehat Y^{\mathrm{aux}}_{i,t+1}\right)
-
s_Y\left(Y_{i,t+1}\right)
\right]^2,
\end{equation}
where $s_Y(\cdot)$ is the min-max scaler for cohort-week sales estimated from the calibration data. 
Operationally, the model first maps the predicted acquisition, ROPC, and AOV values back to their original scales, combines them through Equation~\eqref{eq:wa_aux_sales}, and then rescales the auxiliary revenue value before computing the squared loss. 
This procedure maintains numerical stability while allowing gradients from the auxiliary loss to update the acquisition, ROPC, AOV, and shared-layer parameters.

\paragraph{Interpretation of the auxiliary regularizer.}
We interpret $L_C$ as coherence-oriented predictive regularization rather than a literal accounting identity because acquisition is modeled as a calendar-week forecast rather than a fixed cohort-size attribute. The loss may still help the shared representation learn common business forces, discourage primitive forecasts that imply implausible revenue magnitudes, and combine multiple cohort histories as views of the same week-level demand environment. Gains from this term therefore indicate that structured alignment is predictively useful in this setting, not that the component forecasts satisfy an exact accounting identity.

\paragraph{Auxiliary aggregate-sales loss.}
The model includes aggregate sales as an auxiliary stochastic covariate, and the corresponding loss $L_S$ trains the sales head to predict the next value of this input channel. 
This prediction is used during walk-forward inference only to update the aggregate-sales history without using realized future sales. 
It is not the focal revenue forecast reported in the analysis.
$L_S$ provides a global training signal and a leakage-free recursive update for the aggregate-sales input channel, while the main revenue forecast remains decomposition-based.

\paragraph{Loss weights.}
The weights $\omega_k$ and $\omega_C$ balance the task-specific objectives and the auxiliary revenue-alignment objective. 
In the implementation, each weight is chosen from a small grid of low and high values based on validation-period performance as a part of hyperparameter optimization.

\paragraph{Optimization.}
Training uses AdamW. 
The shared backbone receives a lower learning rate than the task-specific heads, allowing the backbone to learn stable representations while permitting the heads to adapt to task-specific scales and dynamics. 
Because AOV is relatively noisy and sensitive to product mix, we apply stronger weight decay to the AOV head. 
Early stopping monitors validation performance and terminates training when validation error no longer improves. 
All normalization parameters, embedding dictionaries, and selected hyperparameters are estimated using calibration data only and then held fixed during holdout forecasting.

\subsection{Walk-Forward Inference Algorithm}\label{wa:inference}

The holdout forecasts are generated recursively using only information available at each forecast origin.

For inference, cohort revenue contributions are constructed using the relevant cohort size. 
For cohorts observed before the forecast horizon, the cohort size is the observed historical size $A_i$. 
For cohorts born during the forecast horizon, the predicted acquisition in the birth week defines the forecasted cohort size. 
Let
\begin{equation}\label{eq:wa_inference_cohort_size}
\widehat A^{\mathrm{coh}}_i
=
\begin{cases}
A_i, & \text{if cohort } i \text{ is observed before the forecast horizon},\\
\widehat A_{t_0(i)}, & \text{if cohort } i \text{ is born during the forecast horizon}.
\end{cases}
\end{equation}
Then the inference-time cohort revenue contribution is
\begin{equation}\label{eq:wa_inference_sales}
\widehat Y_{i,t+1}
=
\widehat A^{\mathrm{coh}}_i
\left[
\mathbbm{1}\{t+1=t_0(i)\}
+
\widehat R_{i,t+1}
\right]
\widehat V_{i,t+1}.
\end{equation}
This inference-time quantity is distinct from $\widehat Y^{\mathrm{aux}}_{i,t+1}$. 
The former is used to compute forecasted cohort revenue contributions during walk-forward prediction, while the latter is used as a training regularizer.

Component-implied aggregate sales are obtained by summing inference-time cohort contributions: $\widehat S_{t+1} = \sum_{i\in\mathcal I_{t+1}} \widehat Y_{i,t+1}.$

For week-level tasks, cohort-conditioned forecasts are averaged before the predicted values are appended to the history. Deterministic covariates are known rather than forecast, and no realized holdout outcomes enter the recursion. The same walk-forward procedure is used for CBMT and applicable sequence-model benchmarks.

\begin{algorithm}[!t]
\caption{Walk-forward forecasting with customer-base primitives}
\label{alg:walk_forward_jm}
\begin{algorithmic}[1]
\State \textbf{Inputs:} trained model $\{f_{\psi},g_{\phi_A},g_{\phi_R},g_{\phi_V},g_{\phi_S}\}$, calibration-period scalers, look-back length $p$, forecast horizon $T_0+1,\ldots,T_1$.
\State Initialize histories for acquisition, ROPC, AOV, and auxiliary aggregate sales using the final $p+1$ calibration weeks.
\For{$t=T_0$ to $T_1-1$}
    \State Construct input windows $D_{i,t}$ for cohorts active or potentially active in week $t+1$.
    \State Compute shared representations $h_{i,t}=f_{\psi}(e(D_{i,t}))$.
    \State Generate forecasts $\widehat A^{(i)}_{t+1}$, $\widehat R_{i,t+1}$, $\widehat V_{i,t+1}$, and auxiliary sales update $\widetilde S^{(i),\mathrm{aux}}_{t+1}$.
    \State Average cohort-conditioned acquisition forecasts to obtain $\widehat A_{t+1}$.
    \State Set cohort size $\widehat A^{\mathrm{coh}}_i=A_i$ for existing cohorts and $\widehat A^{\mathrm{coh}}_i=\widehat A_{t_0(i)}$ for cohorts born during the forecast horizon.
    \State Compute component-implied cohort revenue $\widehat Y_{i,t+1}$.
    \State Aggregate component-implied sales:
    $\widehat S_{t+1}=\sum_{i\in\mathcal I_{t+1}}\widehat Y_{i,t+1}$.
    \State Append predicted acquisition, ROPC, AOV, and auxiliary aggregate-sales update to the histories.
\EndFor
\State \textbf{Outputs:} forecasts for acquisition, ROPC, AOV, cohort-level revenue contributions, and component-implied aggregate sales.
\end{algorithmic}
\end{algorithm}

\section{Benchmark Model Implementation Details} \label{sec:wa_benchmark_selection}

This appendix documents the benchmark variants and the selection of main-text representatives. The model families are defined in the main paper's Benchmark Models subsection and summarized by outcome in Table \ref{tab:cbcv_benchmark}. Briefly, GLS, LMP, and SSW denote the classical CBCV specifications of \citet{gupta2004valuing}, \citet{libai2009diffusion}, and \citet{schulze2012linking}; WG-PNBD combines Weibull-Gamma acquisition, Pareto/NBD repeat purchasing, and regression AOV, while WG-PNBD-E adds covariates; long short-term memory (LSTM) is the recurrent sequence benchmark; and RF (Random Forest) and XGB (XGBoost) are tree-based learners. The definitions below concern the implementation suffixes V1--V3 and FP/PP shown in Table \ref{tab:wa_model_selection}.

\subsection{Variants of Probabilistic Benchmark Models}
We test three versions of each CBCV benchmark along two dimensions: the inclusion of time-varying covariates and the granularity of the input data. The set covers both the original aggregate specifications and variants that use the cohort-time panel.

\textbf{Version 1 (V1):} Original implementation using aggregate data (summing across cohorts each week to create a time series) without covariates, following the specifications in the original papers.

\textbf{Version 2 (V2):} Aggregate data with month dummy covariates to capture temporal dynamics while maintaining the aggregate structure assumed in the original papers.

\textbf{Version 3 (V3):} Granular cohort-time panel data with month dummy covariates, representing the most data-rich implementation.

Note that LMP does not allow covariates in its original formulation, so we implement only two versions: one using aggregate data without covariates and one using cohort-time data without covariates.

For the WG-PNBD and WG-PNBD-E models, we explored two approaches to handling cohort heterogeneity in the ROPC component:

\textbf{Fully Pooled (FP):} A single PNBD (or PNBD-E) model fitted to the entire set of cohorts, with cohort heterogeneity captured through a covariate for cohort number (where 1 corresponds to the first cohort and increments thereafter). This approach assumes cohorts improve or degrade monotonically over time.

\textbf{Partially Pooled (PP):} Separate PNBD (or PNBD-E) models fitted to subsets of cohorts grouped by acquisition period (e.g., 2016, Q1 2017, Q2 2017, ..., Q2 2018 and all subsequent cohorts). This approach allows for more flexible cohort heterogeneity patterns, with each subset's parameters reflecting distinct behavioral dynamics.

\subsection{Model Selection Results}

Table \ref{tab:wa_model_selection} presents the full comparison of out-of-sample fit for weekly aggregated total sales, measured by SMAPE and MASE across all model variants. Considering predictive performance jointly across multiple customer-base primitives and sales outcomes, we selected GLS V2 (`GLS') as the representative benchmark from the classical CBCV model family, WG-PNBD-E PP (`WG-PNBD-E') from the probabilistic CRM model family, and Random Forest (`RF') from the traditional machine-learning model family. We also report XGBoost (`XGB') as an additional tree-based benchmark.

\begin{table}[!htbp]
\centering
\caption{Benchmark Model Comparison and Main-Text Representatives}
\label{tab:wa_model_selection}
\begin{tabular}{llrrrr}
\toprule
& & \multicolumn{2}{c}{SMAPE} & \multicolumn{2}{c}{MASE} \\
Models & N & Mean & Median & Mean & Median \\
\midrule
GLS V1 & 966 & 45.13 & 41.92 & 4.47 & 3.29 \\
GLS V2 & 966 & 42.36 & 36.41 & 4.12 & 2.94 \\
GLS V3 & 966 & 142.12 & 151.40 & 10.41 & 8.10 \\
LMP V1 & 966 & 27.93 & 19.07 & 2.96 & 1.70 \\
LMP V2 & 966 & 24.15 & 16.26 & 2.32 & 1.55 \\
SSW V1 & 966 & 23.76 & 17.42 & 2.28 & 1.71 \\
SSW V2 & 966 & 29.67 & 19.30 & 3.06 & 1.84 \\
SSW V3 & 966 & 83.06 & 64.76 & 6.22 & 4.44 \\
WG-PNBD FP & 966 & 29.75 & 22.37 & 3.03 & 1.91 \\
WG-PNBD PP & 966 & 28.83 & 21.38 & 2.88 & 1.90 \\
WG-PNBD-E FP & 966 & 23.10 & 15.00 & 2.10 & 1.39 \\
WG-PNBD-E PP & 966 & 22.11 & 14.83 & 2.12 & 1.41 \\
XGB & 966 & 24.13 & 15.66 & 2.74 & 1.65 \\
RF & 966 & 23.57 & 16.05 & 2.51 & 1.74 \\
LSTM & 966 & 28.15 & 18.85 & 3.34 & 1.66 \\
\bottomrule
\end{tabular}
\end{table}

Aggregate data with covariates generally performs best among the classical CBCV variants, whereas their cohort-time implementations perform worse. Partial pooling often improves the probabilistic CRM models, and adding covariates lowers SMAPE by approximately 23\% for the partially pooled WG-PNBD-E specification relative to WG-PNBD. RF has lower median SMAPE and MASE than XGB in this comparison. These patterns inform the representative benchmarks reported in the main text.

\subsection{Direct Aggregate-Sales Benchmark}
\label{wa:direct_sales_benchmark}

The established benchmarks above forecast the customer-base primitives and combine those forecasts into revenue. A natural complementary question is how CBMT compares with a flexible model trained only to forecast aggregate sales: if a manager's sole objective were a topline weekly-sales forecast, how well would a Transformer trained directly on aggregate-sales history and calendar covariates perform? This appendix reports that practical benchmark.

\paragraph{Model and inputs.}
The direct model is a company-specific Transformer that forecasts weekly company-level sales. The target is weekly spend aggregated to the company-week level. Each training example pairs a rolling window of scaled weekly sales with aligned calendar covariates: a one-week-ahead holiday indicator, linear and quadratic time trends, and a week-of-year index passed through a learned embedding. The PyTorch architecture uses sinusoidal positional encoding, a learned input projection, multi-head self-attention, feed-forward layers, and a final linear layer for the one-week-ahead sales forecast. Forecasts over the reported test period are generated recursively and evaluated on the original sales scale. The model does not forecast acquisition, repeat purchasing, or order value.

\paragraph{Tuning design.}
The reported optimization evaluates eight Transformer configurations separately for each of 966 companies, yielding 7{,}728 completed candidate runs and no reported failures. The reported design includes a 13-week validation window and a 47-week test window. As shown in Table~\ref{tab:wa_direct_sales_grid}, the grid varies the input-history length, learning rate, model width, feed-forward dimension, encoder and decoder depth, and dropout. All candidates use four attention heads, batch size 64, gradient clipping at 1, ReLU activation, and seed 0.

\begin{table}[!htbp]
\centering
\caption{Direct Aggregate-Sales Transformer Candidate Grid}
\label{tab:wa_direct_sales_grid}
\begin{tabular}{@{}lrrrrrrr@{}}
\toprule
Candidate & Input weeks & Learning rate & $d_{\mathrm{model}}$ & FF dim. & Enc. & Dec. & Dropout \\ \midrule
C01 & 20 & 0.0005 & 32 & 64  & 2 & 2 & 0.00 \\
C02 & 20 & 0.0001 & 32 & 64  & 2 & 2 & 0.10 \\
C03 & 20 & 0.0010 & 32 & 64  & 2 & 2 & 0.05 \\
C04 & 20 & 0.0005 & 16 & 32  & 1 & 1 & 0.10 \\
C05 & 20 & 0.0005 & 64 & 128 & 2 & 2 & 0.10 \\
C06 & 20 & 0.0003 & 32 & 128 & 3 & 2 & 0.10 \\
C07 & 12 & 0.0005 & 32 & 64  & 2 & 2 & 0.05 \\
C08 & 24 & 0.0003 & 32 & 64  & 2 & 2 & 0.10 \\
\bottomrule
\end{tabular}
\begin{minipage}{\textwidth}
\small
\noindent\textit{Note: } Enc and Dec denote the number of encoder and decoder layers, FF dim. is the feed-forward width, and $d_{\mathrm{model}}$ is the model dimension. The fixed settings are reported in the text.
\end{minipage}
\end{table}

\paragraph{Results.}
Table~\ref{tab:wa_direct_sales_results} reports summaries for the tuned direct model and CBMT total sales. The company identifiers align one-to-one across all 966 observations, allowing paired inference. CBMT's mean SMAPE is 15.48 versus 15.90 for the tuned direct model, a 2.65\% reduction. The paired direct-minus-CBMT mean difference is 0.42 SMAPE points and is not statistically distinguishable from zero (paired $t$-test $p=.222$). CBMT also has lower cross-company SMAPE dispersion (19.14 versus 22.59) and lower mean MASE (1.28 versus 1.40), whereas the direct model has the lower median SMAPE (9.58 versus 10.48) and median MASE (0.93 versus 0.95).

\begin{table}[!htbp]
\centering
\caption{Weekly Total-Sales Accuracy: CBMT versus Tuned Direct Model}
\label{tab:wa_direct_sales_results}
\begin{tabular}{@{}lrrrrrr@{}}
\toprule
          &      & \multicolumn{3}{c}{SMAPE} & \multicolumn{2}{c}{MASE} \\
\cmidrule(l){3-5} \cmidrule(l){6-7}
Strategy & N & \multicolumn{1}{c}{Mean} & \multicolumn{1}{c}{SD} & \multicolumn{1}{c}{Median} & \multicolumn{1}{c}{Mean} & \multicolumn{1}{c}{Median} \\
\midrule
CBMT (customer-base decomposition)       & 966 & \textbf{15.48} & \textbf{19.14} & 10.48 & \textbf{1.28} & 0.95 \\
Tuned direct aggregate-sales Transformer & 966 & 15.90 & 22.59 & 9.58 & 1.40 & 0.93 \\
\bottomrule
\end{tabular}
\begin{minipage}{\textwidth}
\small
\noindent\textit{Note: } Lower mean values and the lower SMAPE standard deviation are bolded. Paired inference uses the matched company-level SMAPE observations reported for the two models.
\end{minipage}
\end{table}

\section{Construction and MAE Evaluation of the Sources of Sales Change}
\label{wa:source_sales}

This appendix documents the construction and evidence behind the main paper's sales-bridge comparison. The bridge is an accounting decomposition that reconciles exactly: it measures how close forecasts are to the realized change in total sales and to the three sources of that change. It describes which customer behaviors changed, not why they changed.

\subsection{Endpoint Construction, Cohort Cutoff, and Reconciliation}

The bridge uses two seasonally aligned 47-week windows. The actual baseline runs from April 8, 2018 through February 24, 2019, and the target window runs from April 7, 2019 through February 23, 2020. February 24, 2019 is the fixed cohort cutoff. Both the realized and predicted comparisons use the same actual baseline. Realized endpoints retain the existing source-bridge actual exports and raw cohort histories; the scenario-specific actual exports used for SMAPE validation are not substituted into this analysis. The CBMT predicted endpoint is sales reconstructed from the final scenario-3 acquisition, repeat orders per customer (ROPC), and average order value (AOV) exports, with observed acquisition for cohorts born before the holdout and final predicted acquisition held fixed over the lifetime of cohorts born during the holdout.

The immutable transaction history separates initial and repeat activity as follows. In a cohort's birth week, initial orders equal total orders minus repeat orders and initial sales equal total spend minus repeat spend. Repeat orders and repeat sales are read directly from the corresponding repeat fields. All target-window activity from cohorts acquired by the cutoff is repeat activity by construction.

Let $Q_{0r}$, $S_{0r}$, and $A_{0r}=S_{0r}/Q_{0r}$ denote baseline repeat orders, repeat sales, and repeat spend per order. Let $Q_{1r}$, $S_{1r}$, and $A_{1r}=S_{1r}/Q_{1r}$ denote the corresponding target-window quantities for cohorts acquired by the cutoff. Let $G_1$ denote target-window sales from cohorts acquired after the cutoff and let $I_0$ denote baseline initial-order sales. The three displayed contributions are
\begin{align*}
C^{\mathrm{vol}} &= (Q_{1r}-Q_{0r})\frac{A_{0r}+A_{1r}}{2},\\
C^{\mathrm{spend}} &= (A_{1r}-A_{0r})\frac{Q_{0r}+Q_{1r}}{2},\\
C^{\mathrm{net\,repl}} &= G_1-I_0.
\end{align*}
The first two terms satisfy $C^{\mathrm{vol}}+C^{\mathrm{spend}}=S_{1r}-S_{0r}$. Adding net customer-base replenishment yields target sales minus baseline sales. Gross post-baseline cohort sales and baseline initial-order sales are construction fields, not additional headline contributions.

Primary eligibility requires finite repeat orders and sales at all three repeat endpoints, strictly positive non-near-zero aggregate repeat orders, a finite repeat-spend-per-order denominator, and successful raw-history and outer-endpoint reconciliation. An aggregate is near zero when it is nonzero and its absolute sum divided by its sum of absolute row values is no greater than $10^{-12}$. Zero or negative aggregate repeat sales and repeat spend per order remain eligible when the denominator is valid; source values are preserved without clipping. Babies ``R'' Us is the sole primary exclusion because its reconstructed-predicted target-existing repeat orders and sales are both exactly zero, leaving repeat spend per order undefined. The primary sample therefore contains 965 of the inherited 966 companies.

Every repeat subtotal, net-replenishment definition, and outer bridge must reconcile in memory and after 17-significant-digit serialization within $10^{-6}$. The maximum observed repeat-subtotal and outer errors are $4.18\times10^{-7}$ and $3.58\times10^{-7}$, respectively. The strict prespecified sensitivity additionally requires positive repeat sales and repeat spend per order and a cancellation ratio $|\sum x|/\sum|x|$ of at least 0.95 at every repeat endpoint. It contains 954 companies.

\subsection{Benchmark Eligibility and Identical Common Samples}

Benchmark eligibility was decided before inspecting bridge performance. An eligible method needed immutable source provenance, all required company identifiers, compatible cohort-week support, matching realized fields, finite acquisition/ROPC/AOV primitives, and a defensible acquisition value fixed over each holdout-born cohort's lifetime. Table~\ref{tab:wa_source_eligibility} records the disposition of all 15 benchmark variants.

\begin{table}[!htbp]
\centering
\scriptsize
\caption{Composition Eligibility of the 15 Benchmark Variants}
\label{tab:wa_source_eligibility}
\begin{tabularx}{\textwidth}{@{}p{0.22\textwidth}p{0.10\textwidth}p{0.16\textwidth}X@{}}
\toprule
Method(s) & Panel & Disposition & Reason \\
\midrule
WG-PNBD-E PP & Primary & Eligible & All 965 frozen-primary firms pass coverage, key, realized-field, fixed-acquisition, and finite-primitive checks. \\
RF & Primary & Eligible & All 965 firms pass after carrying the calendar-week acquisition forecast unchanged over each holdout-born cohort's lifetime. \\
XGB & Primary & Eligible & All 965 firms pass the same fixed-cohort acquisition and primitive checks as RF. \\
GLS V2 & Primary & Ineligible & Fourteen primary firms lack a finite recoverable required primitive, specifically repeat spend per order. \\
LSTM & Primary & Ineligible & Repeat exports omit pre-training-start cohorts needed for target-existing repeat orders and sales; censored spend cannot recover repeat orders or repeat spend per order. \\
WG-PNBD FP; WG-PNBD PP; WG-PNBD-E FP & Secondary & Eligible & Each variant has complete source coverage and all 965 firms pass the frozen primitive and fixed-acquisition gates. \\
GLS V1; GLS V3; LMP V1; LMP V2; SSW V1; SSW V2 & Secondary & Ineligible & Fourteen primary firms lack a finite recoverable required primitive, specifically repeat spend per order. \\
SSW V3 & Secondary & Ineligible & Ninety-four required sample exports are missing, and the remaining files do not provide a complete fixed-endpoint, finite-primitive path. \\
\bottomrule
\end{tabularx}
\begin{minipage}{\textwidth}
\footnotesize
\noindent\textit{Note:} RF denotes Random Forest and XGB denotes XGBoost. ``Primary'' and ``secondary'' describe the prespecified reference panels, not performance. Ineligibility is substantive and performance-blind; no comparison is made with an ineligible or unrun method.
\end{minipage}
\end{table}

The primary common sample contains CBMT, WG-PNBD-E PP, RF, and XGB. The secondary common sample contains CBMT, WG-PNBD FP, WG-PNBD PP, and WG-PNBD-E FP. Both use the identical 965 companies and exclude only Babies ``R'' Us under the frozen CBMT denominator rule. Every method uses the same actual baseline, target dates, cohort cutoff, realized outcomes, contribution formulas, and $10^{-6}$ reconciliation tolerance. For RF and XGB, the corrected stored acquisition and sales columns independently reproduce the already-frozen bridge endpoint; they do not alter the results reported here.

\subsection{Full MAE Results and Paired Uncertainty}

For method $m$ and outcome $k$, mean absolute error is
\begin{equation*}
\operatorname{MAE}_{mk}=\frac{1}{965}\sum_{i=1}^{965}\left|\widehat C_{imk}-C_{ik}\right|.
\end{equation*}
The percentage-point scale divides each company-level contribution error by that company's actual baseline sales and multiplies by 100. The main text reports this scaled version because it places firms of different sizes on a common scale. Lower MAE means the forecast is closer, on average, to the realized outcome. Table~\ref{tab:wa_source_mae_full} reports the complete point estimates in percentage points and in millions of dollars. For CBMT, the four percentage-point MAEs are 7.91, 6.43, 5.91, and 5.80, and their dollar-scale counterparts are \$4.18, \$3.61, \$4.05, and \$2.69 million.

\begin{table}[!htbp]
\centering
\scriptsize
\caption{Full Mean Absolute Error Results on the Identical 965-Company Sample}
\label{tab:wa_source_mae_full}
\resizebox{\textwidth}{!}{%
\begin{tabular}{lrrrrrrr}
\toprule
Outcome & CBMT & WG-PNBD-E PP & RF & XGB & WG-PNBD FP & WG-PNBD PP & WG-PNBD-E FP \\
\midrule
\multicolumn{8}{l}{\textit{Panel A: Percentage points}} \\
Total sales change & 7.91 & 18.17 & 22.60 & 28.87 & 22.70 & 22.13 & 18.62 \\
Existing-base repeat-order volume & 6.43 & 9.60 & 11.11 & 11.18 & 9.98 & 11.07 & 9.81 \\
Existing-base repeat spend per order & 5.91 & 8.27 & 8.70 & 12.65 & 7.83 & 8.22 & 5.43 \\
Net customer-base replenishment & 5.80 & 6.93 & 10.10 & 13.01 & 10.58 & 8.39 & 8.49 \\
\multicolumn{8}{l}{\textit{Panel B: Millions of dollars}} \\
Total sales change & 4.18 & 10.12 & 17.49 & 20.16 & 14.52 & 14.77 & 9.45 \\
Existing-base repeat-order volume & 3.61 & 8.34 & 4.02 & 7.89 & 8.53 & 9.99 & 8.30 \\
Existing-base repeat spend per order & 4.05 & 5.64 & 12.47 & 9.78 & 4.00 & 5.59 & 4.20 \\
Net customer-base replenishment & 2.69 & 2.46 & 4.45 & 6.15 & 4.67 & 3.31 & 3.00 \\
\bottomrule
\end{tabular}%
}
\begin{minipage}{\textwidth}
\footnotesize
\noindent\textit{Note:} Every entry uses the same 965 companies. Panel A supplies the standardized paper comparison; Panel B reports the corresponding dollar scale.
\end{minipage}
\end{table}

Table~\ref{tab:wa_source_mae_paired} reports paired benchmark-minus-CBMT differences in absolute error. Each interval comes from 10{,}000 paired-company percentile-bootstrap resamples using seed 20260626. Positive differences favor CBMT. CBMT has lower point-estimate MAE in 23 of 24 cells. The only negative point estimate is the repeat-spend-per-order comparison with WG-PNBD-E FP: its benchmark-minus-CBMT difference is $-0.47$ percentage points, with a 95\% interval of [$-1.01$, 0.06], so the exception is not statistically distinguishable from zero.

\begin{table}[!htbp]
\centering
\scriptsize
\caption{Paired Differences in MAE Relative to CBMT}
\label{tab:wa_source_mae_paired}
\resizebox{\textwidth}{!}{%
\begin{tabular}{lrrrr}
\toprule
Benchmark & Total sales change & Repeat-order volume & Repeat spend per order & Net replenishment \\
\midrule
WG-PNBD-E PP & 10.26 [9.14, 11.49] & 3.18 [2.41, 3.94] & 2.36 [1.74, 2.98] & 1.13 [0.48, 1.86] \\
RF & 14.69 [11.97, 17.74] & 4.68 [3.93, 5.46] & 2.79 [1.55, 4.25] & 4.30 [2.82, 6.17] \\
XGB & 20.96 [15.40, 27.97] & 4.75 [3.27, 6.75] & 6.74 [3.25, 11.35] & 7.21 [4.58, 10.72] \\
WG-PNBD FP & 14.79 [13.51, 16.15] & 3.55 [2.83, 4.31] & 1.92 [1.35, 2.49] & 4.77 [4.02, 5.61] \\
WG-PNBD PP & 14.23 [12.96, 15.56] & 4.64 [3.87, 5.45] & 2.32 [1.69, 2.95] & 2.59 [1.92, 3.36] \\
WG-PNBD-E FP & 10.72 [9.45, 12.08] & 3.39 [2.64, 4.18] & -0.47 [-1.01, 0.06] & 2.69 [2.03, 3.45] \\
\bottomrule
\end{tabular}%
}
\begin{minipage}{\textwidth}
\footnotesize
\noindent\textit{Note:} Entries are benchmark-minus-CBMT differences in company-level absolute error, in percentage points of baseline sales, with paired 95\% bootstrap intervals in brackets. Positive values mean lower error for CBMT. All comparisons use $n=965$ identical companies. The WG-PNBD-E FP repeat-spend-per-order difference is $-0.47$ [$-1.01$, 0.06] and is statistically indistinguishable from zero.
\end{minipage}
\end{table}

\subsection{Prespecified MAE Sensitivity}

The strict sensitivity removes the 11 otherwise primary-eligible firms that fail the positive, low-cancellation repeat-endpoint rule, leaving 954 companies. Table~\ref{tab:wa_source_mae_sensitivity} reports the directly relevant percentage-point MAEs. CBMT's four MAEs are 7.78, 6.37, 5.70, and 5.68. The point-estimate pattern remains 23 of 24 benchmark-by-outcome comparisons in CBMT's favor, with WG-PNBD-E FP lower only for repeat spend per order. The paired inferential statement is based on the primary 965-company sample in Table~\ref{tab:wa_source_mae_paired}; no new interval is inferred from this sensitivity table.

\begin{table}[!htbp]
\centering
\scriptsize
\caption{Prespecified 954-Company Sensitivity: Mean Absolute Error}
\label{tab:wa_source_mae_sensitivity}
\resizebox{\textwidth}{!}{%
\begin{tabular}{lrrrrrrr}
\toprule
Outcome & CBMT & WG-PNBD-E PP & RF & XGB & WG-PNBD FP & WG-PNBD PP & WG-PNBD-E FP \\
\midrule
Total sales change & 7.78 & 17.89 & 20.80 & 24.75 & 22.46 & 21.89 & 18.23 \\
Existing-base repeat-order volume & 6.37 & 9.31 & 10.85 & 10.13 & 9.69 & 10.79 & 9.50 \\
Existing-base repeat spend per order & 5.70 & 8.14 & 7.75 & 10.81 & 7.72 & 8.06 & 5.31 \\
Net customer-base replenishment & 5.68 & 6.85 & 9.44 & 11.23 & 10.50 & 8.36 & 8.42 \\
\bottomrule
\end{tabular}%
}
\begin{minipage}{\textwidth}
\footnotesize
\noindent\textit{Note:} Entries are MAEs in percentage points of baseline sales on the fixed 954-company strict sensitivity. Lower values mean forecasts are closer to realized outcomes.
\end{minipage}
\end{table}

The bridge and comparisons remain descriptive and noncausal. Repeat spend per order is not a pure price effect, and net customer-base replenishment is not an incremental acquisition effect. No result in this section supports a claim about a benchmark that failed the eligibility gate or was not run.

\section{Sensitivity to an Alternative Error Metric}
\label{append:main_robustness}

We report mean absolute scaled error (MASE) alongside SMAPE to determine whether the performance ranking depends on the error metric.

\subsubsection*{MASE Definition}
\begin{equation*}
\text{MASE} = \frac{\frac{1}{n}\sum_{t=1}^{n}|F_t - X_t|}{\frac{1}{n-1}\sum_{t=2}^{n}|X_t - X_{t-1}|}
\end{equation*}
where $F_t$ is the forecast and $X_t$ is the actual value in period $t$. MASE scales absolute error by the error from a naive one-step-ahead forecast; values below 1 indicate improvement over that benchmark.

\begin{table}[!htbp]
\centering
\small
\caption{Mean Forecast Accuracy Under SMAPE and MASE}
\label{tab:eval_all}
\begin{tabular}{lllrrrrr}
\toprule
& & & CBMT & GLS & WG-PNBD-E & LSTM & RF \\
\midrule
\multicolumn{8}{l}{Acquisition} \\
 & SMAPE & ($\mu$) & 19.30 & 27.71 & 23.31 & 26.03 & 26.02 \\
 &  &  &  & (+30\%) & (+17\%) & (+26\%) & (+26\%) \\
 &  & ($\sigma$) & 20.12 & 23.86 & 23.66 & 29.09 & 20.68 \\
 & MASE & ($\mu$) & 1.42 & 1.76 & 1.48 & 1.94 & 1.46 \\
 &  &  &  & (+20\%) & (+4\%) & (+27\%) & (+3\%) \\
 &  & ($\sigma$) & 1.60 & 1.89 & 1.92 & 4.28 & 2.53 \\
\multicolumn{8}{l}{ROPC} \\
 & SMAPE & ($\mu$) & 35.56 & 178.64 & 50.85 & 42.56 & 37.26 \\
 &  &  &  & (+80\%) & (+30\%) & (+16\%) & (+5\%) \\
 &  & ($\sigma$) & 24.68 & 17.70 & 27.94 & 25.90 & 24.30 \\
 & MASE & ($\mu$) & 0.76 & 5.20 & 1.49 & 1.08 & 0.79 \\
 &  &  &  & (+85\%) & (+49\%) & (+29\%) & (+3\%) \\
 &  & ($\sigma$) & 0.57 & 4.71 & 1.32 & 0.98 & 0.68 \\
\multicolumn{8}{l}{AOV} \\
 & SMAPE & ($\mu$) & 35.94 & 39.37 & 38.59 & 41.05 & 38.62 \\
 &  &  &  & (+9\%) & (+7\%) & (+12\%) & (+7\%) \\
 &  & ($\sigma$) & 25.00 & 29.92 & 29.33 & 31.05 & 31.39 \\
 & MASE & ($\mu$) & 0.96 & 0.95 & 0.95 & 1.06 & 0.93 \\
 &  &  &  & (-2\%) & (-2\%) & (+9\%) & (-4\%) \\
 &  & ($\sigma$) & 0.58 & 0.47 & 0.54 & 0.80 & 0.59 \\
\bottomrule
\end{tabular}
\end{table}

\begin{table}[!htbp]
\centering
\small
\begin{tabular}{lllrrrrr}
\toprule
& & & CBMT & GLS & WG-PNBD-E & LSTM & RF \\
\midrule
\multicolumn{8}{l}{Cohort-Week Sales} \\
 & SMAPE & ($\mu$) & 53.91 & 170.93 & 64.47 & 61.06 & 59.29 \\
 &  &  &  & (+68\%) & (+16\%) & (+12\%) & (+9\%) \\
 &  & ($\sigma$) & 30.40 & 21.41 & 32.44 & 31.06 & 33.38 \\
 & MASE & ($\mu$) & 0.51 & 2.16 & 0.81 & 0.81 & 0.53 \\
 &  &  &  & (+76\%) & (+36\%) & (+37\%) & (+2\%) \\
 &  & ($\sigma$) & 0.30 & 2.13 & 0.63 & 0.94 & 0.47 \\
\multicolumn{8}{l}{Total Sales} \\
 & SMAPE & ($\mu$) & 15.48 & 42.36 & 22.11 & 28.15 & 23.57 \\
 &  &  &  & (+63\%) & (+30\%) & (+45\%) & (+34\%) \\
 &  & ($\sigma$) & 19.14 & 27.82 & 23.60 & 28.77 & 24.94 \\
 & MASE & ($\mu$) & 1.28 & 4.12 & 2.12 & 3.34 & 2.51 \\
 &  &  &  & (+69\%) & (+39\%) & (+62\%) & (+49\%) \\
 &  & ($\sigma$) & 1.45 & 4.04 & 2.81 & 5.88 & 3.18 \\
\bottomrule
\end{tabular}
\end{table}

\begin{table}[!htbp]
\centering
\small
\caption{Median Forecast Accuracy Under SMAPE and MASE}
\label{tab:median}
\begin{tabular}{lllrrrrr}
\toprule
& & & CBMT & GLS & WG-PNBD-E & LSTM & RF \\
\midrule
\multicolumn{8}{l}{Acquisition} \\
 & SMAPE & & 13.23 & 21.16 & 16.50 & 17.08 & 21.08 \\
 &  & &  & (+37\%) & (+20\%) & (+23\%) & (+37\%) \\
 & MASE & & 0.99 & 1.28 & 1.03 & 0.97 & 1.07 \\
 &  & &  & (+22\%) & (+4\%) & (-3\%) & (+8\%) \\
\multicolumn{8}{l}{ROPC} \\
 & SMAPE & & 29.60 & 185.19 & 44.01 & 36.58 & 32.08 \\
 &  & &  & (+84\%) & (+33\%) & (+19\%) & (+8\%) \\
 & MASE & & 0.67 & 3.73 & 1.02 & 0.80 & 0.70 \\
 &  & &  & (+82\%) & (+34\%) & (+16\%) & (+5\%) \\
\multicolumn{8}{l}{AOV} \\
 & SMAPE & & 29.68 & 30.80 & 30.50 & 32.22 & 29.93 \\
 &  & &  & (+4\%) & (+3\%) & (+8\%) & (+1\%) \\
 & MASE & & 0.82 & 0.83 & 0.81 & 0.85 & 0.80 \\
 &  & &  & (+2\%) & (-1\%) & (+3\%) & (-2\%) \\
\multicolumn{8}{l}{Cohort-Week Sales} \\
 & SMAPE & & 47.56 & 179.45 & 56.46 & 53.92 & 51.87 \\
 &  & &  & (+73\%) & (+16\%) & (+12\%) & (+8\%) \\
 & MASE & & 0.45 & 1.40 & 0.59 & 0.54 & 0.41 \\
 &  & &  & (+68\%) & (+23\%) & (+16\%) & (-9\%) \\
\multicolumn{8}{l}{Total Sales} \\
 & SMAPE & & 10.48 & 36.41 & 14.83 & 18.85 & 16.05 \\
 &  & &  & (+71\%) & (+29\%) & (+44\%) & (+35\%) \\
 & MASE & & 0.95 & 2.94 & 1.41 & 1.66 & 1.74 \\
 &  & &  & (+68\%) & (+33\%) & (+43\%) & (+45\%) \\
\bottomrule
\end{tabular}
\end{table}

\section{Alternative Performance Measures Across All Model Variants}\label{append:relative_perform}

The main paper emphasizes a small set of representative benchmarks. Tables~\ref{tab:append_avg_rank} and \ref{tab:append_p_rank1} instead compare CBMT with all 15 benchmark variants using average rank and the probability of ranking first across firms.
\begin{table}[!htbp]
\centering
\caption{Average Forecast-Accuracy Rank Across All Model Variants}
\label{tab:append_avg_rank}
\begin{threeparttable}
\begin{tabular}{@{}lrrrrr@{}}
\toprule
Models & Acquisition & ROPC & AOV & Cohort-Week Sales & Total Sales \\
\midrule
\textbf{CBMT} & \textbf{3.50} & \textbf{2.12} & \textbf{6.07} & \textbf{2.15} & \textbf{3.06} \\
LSTM & 5.30 & 3.85 & 9.37 & 4.43 & 7.80 \\
RF & 6.74 & 3.16 & 6.31 & 4.15 & 7.01 \\
XGB & 6.31 & 2.83 & 6.97 & 4.10 & 6.55 \\
WG-PNBD-E FP & 4.92 & 5.14 & 8.44 & 4.95 & 6.02 \\
WG-PNBD-E PP & 4.83 & 6.06 & 7.00 & 5.56 & 6.17 \\
WG-PNBD FP & 7.68 & 6.41 & 7.52 & 5.38 & 9.22 \\
WG-PNBD PP & 8.68 & 7.07 & 8.46 & 6.48 & 8.74 \\
GLS V1 & 8.26 & 12.75 & 7.61 & 10.37 & 11.91 \\
GLS V2 & 6.36 & 12.27 & 8.28 & 10.43 & 11.02 \\
GLS V3 & 8.52 & 15.20 & 9.52 & 14.11 & 15.39 \\
LMP V1 & 12.36 & 12.19 & 9.64 & 12.73 & 8.05 \\
LMP V2 & 12.19 & 11.70 & 10.23 & 12.62 & 6.61 \\
SSW V1 & 12.47 & 10.48 & 9.20 & 11.74 & 7.64 \\
SSW V2 & 12.60 & 10.37 & 10.31 & 11.88 & 8.55 \\
SSW V3 & 15.29 & 14.39 & 11.07 & 14.92 & 12.27 \\
\bottomrule
\end{tabular}
\begin{tablenotes}
\footnotesize
\item \textit{Note:} The rank of a method for a given outcome is the average of company-specific ranks for that method. Lower ranks indicate better performance, with 1 being the best and 16 being the worst. Each company-outcome pair is ranked by SMAPE across all 16 models.
\end{tablenotes}
\end{threeparttable}
\end{table}

\begin{table}[!htbp]
\centering
\caption{Probability of Ranking First Across All Model Variants}
\label{tab:append_p_rank1}
\begin{threeparttable}
\begin{tabular}{@{}lrrrrr@{}}
\toprule
Models & Acquisition & ROPC & AOV & Cohort-Week Sales & Total Sales \\
\midrule
\textbf{CBMT} & \textbf{0.347} & \textbf{0.495} & \textbf{0.220} & \textbf{0.545} & \textbf{0.441} \\
LSTM & 0.236 & 0.095 & 0.090 & 0.099 & 0.072 \\
RF & 0.046 & 0.186 & 0.166 & 0.114 & 0.060 \\
XGB & 0.055 & 0.195 & 0.157 & 0.105 & 0.079 \\
WG-PNBD-E FP & 0.065 & 0.012 & 0.034 & 0.037 & 0.057 \\
WG-PNBD-E PP & 0.087 & 0.004 & 0.064 & 0.014 & 0.060 \\
WG-PNBD FP & 0.038 & 0.004 & 0.079 & 0.053 & 0.016 \\
WG-PNBD PP & 0.001 & 0.000 & 0.048 & 0.011 & 0.030 \\
GLS V1 & 0.037 & 0.000 & 0.014 & 0.000 & 0.012 \\
GLS V2 & 0.043 & 0.000 & 0.030 & 0.000 & 0.025 \\
GLS V3 & 0.002 & 0.005 & 0.029 & 0.008 & 0.006 \\
LMP V1 & 0.017 & 0.000 & 0.004 & 0.000 & 0.028 \\
LMP V2 & 0.013 & 0.000 & 0.005 & 0.002 & 0.043 \\
SSW V1 & 0.003 & 0.000 & 0.002 & 0.001 & 0.022 \\
SSW V2 & 0.006 & 0.000 & 0.020 & 0.000 & 0.026 \\
SSW V3 & 0.003 & 0.003 & 0.037 & 0.010 & 0.023 \\
\bottomrule
\end{tabular}
\begin{tablenotes}
\footnotesize
\item \textit{Note:} Each value is the proportion of selected companies where the model achieves rank 1 among all 16 models. Higher proportions indicate better performance.
\end{tablenotes}
\end{threeparttable}
\end{table}

\section{Explaining Prediction Performance with Meta-Model Analysis}\label{wa:FeatureImportance}

To explain cross-firm variation in prediction performance, we construct 21 calibration-period contextual features in 10 groups (Table \ref{tab:feature_blocks}). Aggregated cohort-level statistics are weighted by cohort size.

\begin{table}[!htbp]
\centering
\caption{List of Contextual Features for Meta-model Analysis}
\label{tab:feature_blocks}
\renewcommand{\arraystretch}{1.15}
\setlength{\tabcolsep}{4.5pt}
\small
\begin{tabular}{p{0.02\linewidth} p{0.33\linewidth} p{0.57\linewidth}}
\hline
 & Group & Contextual Factors \\ \hline
A. & Scale &
Total Sales; Total Orders \\
B. & Temporal Concentration &
Sales Herfindahl-Hirschman Index (HHI); Cohort-Orders HHI; Acquisition HHI \\
C. & Volatility &
Cohort-Sales Coefficient of Variation (CV); Median Absolute Deviation (MAD) in Sales \\
D. & Trend Dynamics &
Sales Growth; Orders Growth; Sales Growth Acceleration \\
E. & Acquisition vs. Repeat Mix &
Acquisition Intensity Proportion \\
F. & Retention \& Repeat Intensity &
Decay Rate of Active Customers; Average Orders per Active Customer \\
G. & Dependence &
Autocorrelation (ACF) at Lag 1 for Sales; Partial Autocorrelation (PACF) at Lag 4 for Sales\\
H. & Regularity \& Simple-Model Fit &
Cubic Polynomial Trend $R^2$ for Cohort-Orders; Cubic Polynomial Trend + Seasonality $R^2$ for Orders; Linear Trend + Seasonality $R^2$ for Cohort-AOV \\
I. & Inter-Cohort Heterogeneity &
Gini Coefficient of Cohort Sizes \\
J. & Irregular Shocks &
Spectral Entropy of Sales; Number of Changepoints in Sales\\
\hline
\end{tabular}
\end{table}

The feature groups are constructed as follows:

\begin{enumerate}[label=\Alph*.]
    \item \textbf{Scale} is total sales and total orders accumulated over the calibration window.
    \item \textbf{Temporal Concentration} uses time-based Herfindahl--Hirschman Indices (HHI). Sales HHI is
    \[
        \sum_{t} \left( \frac{S_t}{\sum_{t'} S_{t'}} \right)^{2},
    \]
    with analogous measures for acquisition and cohort-size-weighted cohort orders.
    \item \textbf{Volatility} includes the cohort-size-weighted coefficient of variation of weekly cohort sales and the median absolute change in total sales between consecutive weeks.
    \item \textbf{Trend Dynamics} compares total sales and orders over the last twelve months (LTM) and first twelve months (FTM) to compute annualized growth and acceleration.
    \item \textbf{Acquisition vs. Repeat Mix} is the ratio of revenue from cohorts in their initial month to total LTM revenue.
    \item \textbf{Retention and Repeat Intensity} includes the cohort-size-weighted share of active customers in month 12 relative to acquisition size for cohorts at least one year old, plus average orders per active customer.
    \item \textbf{Dependence} includes lag-1 autocorrelation and lag-4 partial autocorrelation of detrended weekly total sales.
    \item \textbf{Regularity and Simple-Model Fit} uses $R^2$ from low-order trend and seasonal models. For total orders, we fit
    \[
        \log(Orders_t+1)
        = \beta_0 + \beta_1 t + \beta_2 t^2 + \beta_3 t^3
        + \sum_m \beta_m \mathbf{1}\{M=m(t)\}.
    \]
    Cohort-order and cohort-AOV $R^2$ values are averaged using cohort-size weights.
    \item \textbf{Inter-Cohort Heterogeneity} is measured by the cohort-size Gini coefficient,
    \[
        \frac{\sum_{i=1}^{C}\sum_{j=1}^{C} \lvert N_i - N_j \rvert}
        {2C \sum_{j=1}^{C} N_j},
    \]
    where $N_i$ is the initial size of cohort $i$ and $C$ is the number of cohorts.
    \item \textbf{Irregular Shocks} includes spectral entropy of weekly total sales,
    \[
        -\frac{1}{\log K} \sum_{k=1}^{K} p_k \log p_k,
    \]
    and the number of changepoints from a piecewise-constant mean model.
\end{enumerate}

\subsection{Additional Meta-Model Diagnostics}
\label{subsec:metamodel_diagnostics}

We use two complementary meta-models: gradient boosting regression (GBR) and elastic net (ENet). Table~\ref{tab:cv_r2} reports the cross-validated $R^2$ for each specification, or the share of cross-firm variation in holdout SMAPE explained by the 21 contextual features.

\begin{table}[htbp]
\centering
\caption{Cross-Validated $R^2$ by Process, Meta-Model, and Prediction Model}
\label{tab:cv_r2}
\begin{tabular}{llcccc}
\toprule
\textbf{Process} & \textbf{Meta-Model} & \textbf{CBMT} & \textbf{LSTM} & \textbf{WG-PNBD-E} & \textbf{WG-PNBD} \\
\midrule
\multirow{2}{*}{Acquisition} 
    & GBR & 0.249 & 0.411 & 0.465 & 0.319 \\
    & ENet & 0.160 & 0.259 & 0.310 & $-0.006$ \\
\addlinespace
\multirow{2}{*}{\parbox{3cm}{Repeat Orders\\per Customer}} 
    & GBR & 0.405 & 0.437 & 0.528 & 0.466 \\
    & ENet & $-0.009$ & 0.066 & 0.281 & 0.174 \\
\addlinespace
\multirow{2}{*}{\parbox{3cm}{Average Order\\Value}} 
    & GBR & 0.627 & 0.458 & 0.579 & 0.588 \\
    & ENet & 0.465 & 0.407 & 0.466 & 0.457 \\
\addlinespace
\multirow{2}{*}{\parbox{3cm}{Cohort-week\\Sales}} 
    & GBR & 0.495 & 0.503 & 0.610 & 0.459 \\
    & ENet & 0.256 & 0.277 & 0.397 & 0.238 \\
\addlinespace
\multirow{2}{*}{Total Sales} 
    & GBR & 0.407 & 0.477 & 0.396 & 0.397 \\
    & ENet & 0.314 & 0.103 & 0.309 & 0.019 \\
\bottomrule
\end{tabular}
\begin{tablenotes}
\small
\item \textit{Note:} Each entry is the average cross-validated $R^2$ from five-fold cross-validation. Higher values indicate that the contextual features explain more cross-company variation in prediction accuracy. CBMT uses the scenario-3 Full results; ROPC and AOV use cohort-size-weighted SMAPE for all forecasting models, and LSTM values use the fixed rerun snapshot.
\end{tablenotes}
\end{table}

\begin{figure}[!htbp]
    \centering
    \caption{Permutation Importance of Contextual Features for CBMT Forecast Error}
    \label{fig:top_features_GBR}
    \includegraphics[width=0.7\textwidth]{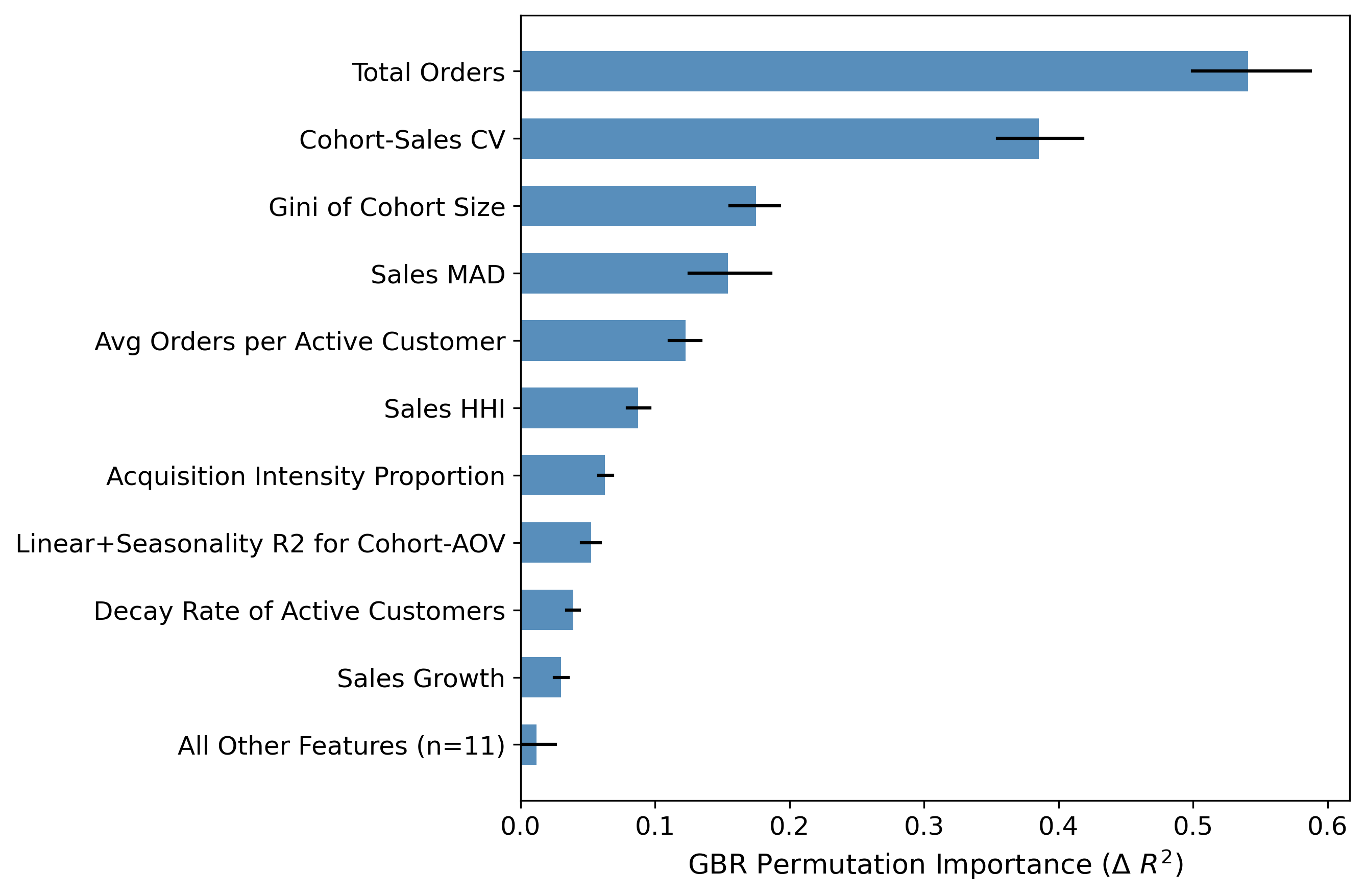}
    \footnotesize
    \begin{minipage}{\textwidth}{\textit{Note:} Bars report the mean decrease in $R^2$ when each feature is permuted; black lines are 95\% confidence intervals from 200 repetitions. The ten leading features are shown separately and the remainder are grouped as ``All Other Features.''}\end{minipage}
\end{figure}

Figure~\ref{fig:top_features_GBR} reports permutation importance for the GBR meta-model of CBMT's cohort-week sales SMAPE. Importance is the decline in $R^2$ after permuting each feature across companies. Total Orders, Cohort-Sales CV, Gini of Cohort Size, and Sales MAD have the largest estimated contributions.

\subsection{Relative Performance Across Scale and Volatility}
\label{subsec:scale_volatility}

\begin{figure}[!htbp]
    \centering
    \caption{Model Win Rates Across Firm Scale and Cohort-Sales Volatility}
    \label{fig:winrate_tercile_2drivers}
    \includegraphics[width=\textwidth]{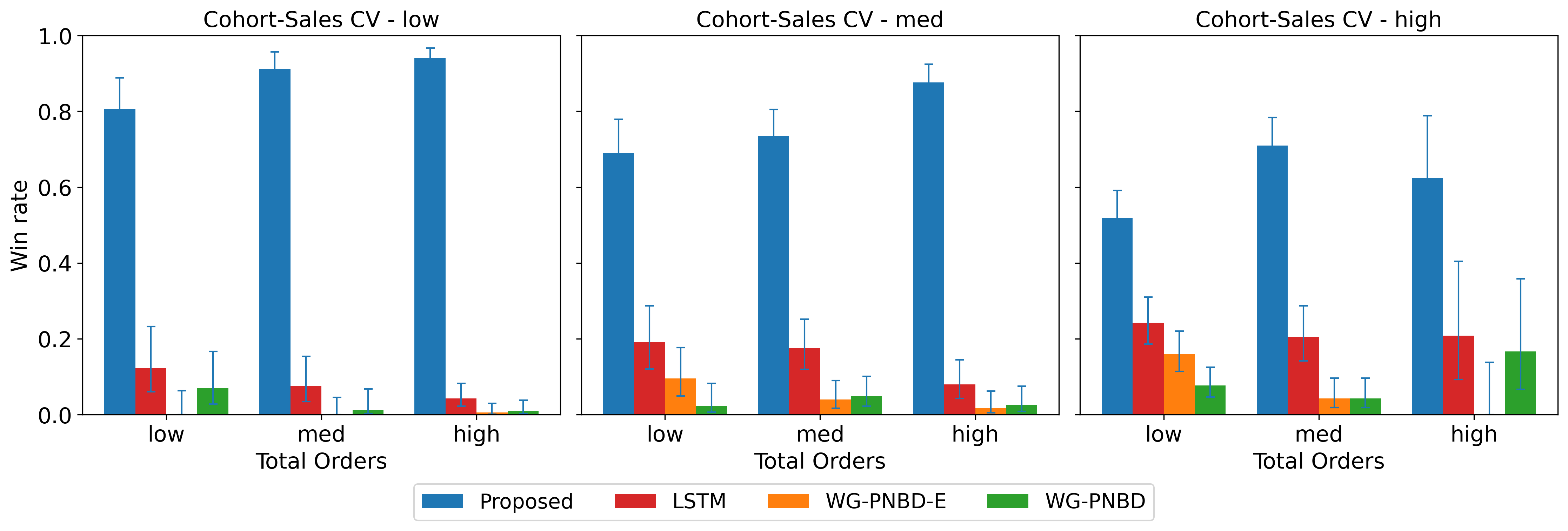}
    \footnotesize
    \begin{minipage}{\textwidth}{\textit{Note:} The 3-by-3 grid is defined by terciles of Total Orders and Cohort-Sales CV. Bars report the share of firms in each cell for which a model has the lowest holdout SMAPE; error bars are Wilson 95\% intervals.}\end{minipage}
\end{figure}

To connect relative performance to structural drivers, we stratify firms into terciles on Total Orders (scale) and Cohort-Sales CV (volatility), producing a $3 \times 3$ grid. Figure~\ref{fig:winrate_tercile_2drivers} presents win rates (the fraction of companies for which each model achieves the lowest holdout SMAPE) within each cell, along with Wilson 95\% confidence intervals.

CBMT is the modal winner in all nine scale-by-volatility cells; the probabilistic baselines rarely win.

\section{Performance Heterogeneity Across Industries} \label{append:hetero_industry}

\begin{figure}[!htbp]
    \centering
    \caption{Average Total-Sales Forecast Error by Industry}
    \label{fig:hetero_industry_sales}
    \includegraphics[width=\textwidth]{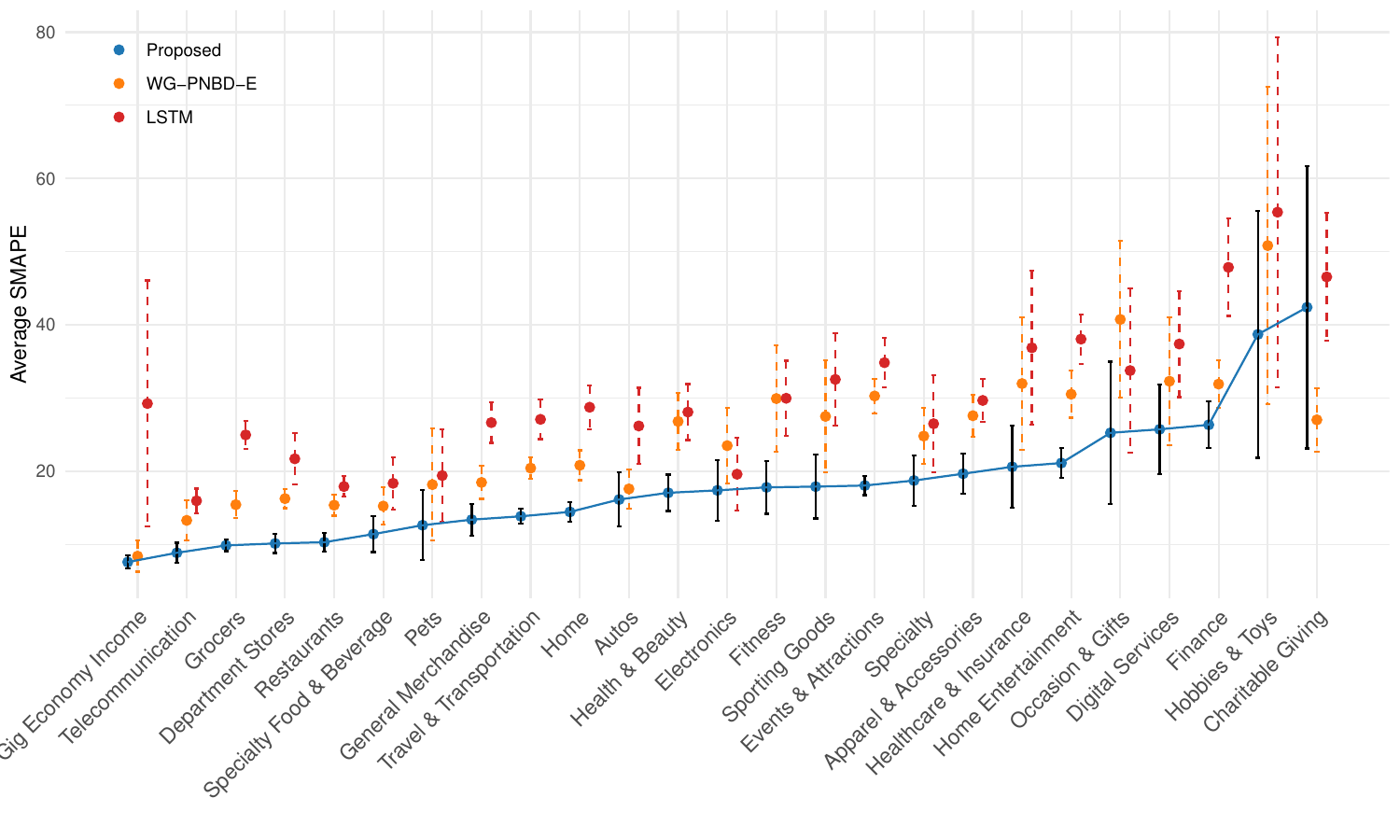}
    \footnotesize
    \begin{minipage}{\textwidth}{\textit{Note:} The horizontal axis lists the industries in Web Appendix~\ref{append:examplecompany}; the vertical axis reports mean holdout total-sales SMAPE within each industry. Points show the model-specific means identified in the legend. Intervals around the CBMT means denote standard errors.} \end{minipage}
\end{figure}

Figure \ref{fig:hetero_industry_sales} compares average holdout total-sales SMAPE for CBMT, WG-PNBD-E PP, and LSTM across the 25 industries, ordered by CBMT's SMAPE. CBMT has the lowest mean SMAPE in 24 of the 25 industries; WG-PNBD-E PP is lower only in Charitable Giving.

\section{External Calibration Using Analyst Revenue Forecasts}\label{wa:AnalystBenchmark}

This appendix reports the external calibration summarized in the main paper's Conclusion. Because CBMT and analysts forecast different revenue measures using different information sets, the comparison is intended only to place CBMT's proportional errors on a familiar scale. It is not a direct benchmark of relative forecasting performance.

\subsection{Data Construction}\label{wass:DataConstruction}

We build a matched firm--quarter dataset with common public-company coverage, fiscal-quarter timestamps, and a March 31, 2019 information cutoff. Of the 966 merchant entities forecast by CBMT, 380 map unambiguously to public parent companies. Merchants sharing a parent are collapsed (e.g., Avis, Budget, and Zipcar to Avis, ticker CAR), yielding 248 tickers. The matched sample is then trimmed by removing the 25 largest analyst--CBMT error-difference outliers, the top 10\%, leaving 223 firms. Weekly panel sales are summed to fiscal-quarter totals using SEC filing calendars.

Analyst forecasts come from IBES for the same 223 tickers and fiscal period-ends from April 2019 through February 2020. The stored analyst value averages eligible forecasts generated during the pre-cutoff calibration period. Both forecast sets therefore exclude information after the common March 31, 2019 cutoff.

\subsection{Empirical Approach}\label{wass:empapproach}

CBMT forecasts are evaluated against panel-observed transaction sales, while analyst forecasts are evaluated against reported firm revenue and may use management guidance, disclosures, industry knowledge, and other institutional resources. Because the targets and information sets differ, their absolute errors are not directly comparable. SMAPE provides a scale-free summary within each native target, so the side-by-side values serve only as a contextual reference.

We compute firm-quarter SMAPE and average it across covered tickers within each horizon. CBMT contributes one forecast per firm-quarter; the analyst value is the average of eligible forecasts generated during the pre-cutoff calibration period for that firm-quarter. Standard errors are obtained by resampling firms 1,000 times. Table~\ref{tab:analyst_comparison_main} reports the descriptive results.

\begin{table}[H]
\centering 
\caption{External Calibration Using Analyst Revenue Forecasts}
\label{tab:analyst_comparison_main}
\label{tab:analyst_eval}
\begin{tabular}{@{}lccc@{}}
\toprule
Horizon
  & \begin{tabular}[c]{@{}c@{}}CBMT Panel-Sales\\SMAPE\end{tabular}
  & \begin{tabular}[c]{@{}c@{}}Analyst Reported-Revenue\\SMAPE\end{tabular}
  & Firms \\ \midrule
Q1 & \begin{tabular}[c]{@{}c@{}}4.56\\ (0.29)\end{tabular}
  & \begin{tabular}[c]{@{}c@{}}6.36\\ (0.61)\end{tabular}
  & 223 \\ \midrule
Q2 & \begin{tabular}[c]{@{}c@{}}4.75\\ (0.42)\end{tabular}
  & \begin{tabular}[c]{@{}c@{}}7.16\\ (0.75)\end{tabular}
  & 219 \\ \midrule
Q3 & \begin{tabular}[c]{@{}c@{}}5.64\\ (0.49)\end{tabular}
  & \begin{tabular}[c]{@{}c@{}}7.35\\ (0.72)\end{tabular}
  & 216 \\ \bottomrule
\end{tabular}
\begin{minipage}[t]{\textwidth}
\vspace*{0.5em}
\footnotesize
\textit{Note.} Horizons are fiscal quarters after the March 31, 2019 forecast origin. CBMT is evaluated against panel sales; analysts are evaluated against reported revenue and may use broader information. The table is a descriptive, scale-free calibration across distinct targets, not a direct test of relative forecasting performance. The 248-ticker matched sample is trimmed by removing the top 10\% of analyst--CBMT error-difference outliers, leaving 223 firms at the first horizon. Bootstrapped standard errors ($B=1{,}000$) appear in parentheses. The analyst value averages eligible forecasts generated during the pre-cutoff calibration period, and Firms is the number of unique firms at each horizon.
\end{minipage}
\end{table}
Across the three horizons, both sets of errors are in the single-digit range on their respective targets. This places CBMT's errors on a familiar descriptive scale, but it does not establish that CBMT matches or outperforms analysts or measure the economic value of using the forecasts.

\subsection{List of Tickers}\label{wass:ListTickers}

{\small
\begin{singlespace}
\begin{multicols}{4}
\raggedright
\sloppy
AAL, AAN, AAP, AAPL, ADBE, ADS, ADT, AEO, ALGT, ALK, ALL, AMC, AMTD, AMZN, ANCUF, ANF, APRN, ARMK, ASCMA, ATUS, AZO, BBBY, BBW, BBY, BGFV, BIG, BJ, BJRI, BKE, BKNG, BKS, BLMN, BURL, CAKE, CAL, CAR, CASY, CBRL, CCL, CHH, CHS, CHTR, CHUY, CI, CMCSA, CMG, CMPR, CNK, COST, CPRI, CRI, CROX, CTL, CUBE, CVS, CVX, DAL, DAVE, DDS, DELL, DENN, DG, DGX, DIN, DIS, DISH, DKS, DLTH, DLTR, DNKN, DPZ, DRI, DUK, EAT, EB, EFX, ETSY, EVLV, EXC, EXPE, EXR, EYE, FB, FDX, FISV, FL, FLWS, FND, FRAN, FRGI, FTDR, FTR, GCO, GDDY, GME, GNC, GOGO, GOOGL, GPS, GRPN, GRUB, GSKY, GT, H, HD, HIBB, HLF, HLT, HOG, HOME, HTZ, HUD, HUM, JACK, JBLU, JWN, JYNT, KIRK, KMX, KR, KSS, LC, LE, LH, LOCO, LOW, LULU, LUV, LYV, M, MAR, MGI, MIK, MTCH, MUSA, NDLS, NGVC, NWL, NWSA, ODP, OLLI, ORLY, OSTK, PBPB, PCG, PGR, PIR, PLAY, PLCE, PLNT, PRTY, PSA, PSX, PZZA, QRTEA, QSR, RAD, RCII, RGS, RL, ROST, RRGB, RTW, S, SAVE, SBH, SBUX, SCHL, SCVL, SEAS, SFIX, SFLY, SFM, SHAK, SHOP, SIG, SIRI, SIX, SKX, SNA, SPTN, SPWH, SQ, SUN, T, TACO, TCS, TGT, TJX, TLYS, TMUS, TPR, TRU, TSCO, TSLA, TXRH, UAA, UAL, UHAL, ULTA, URBN, USM, VFC, VLO, VRA, VVV, VZ, W, WBA, WEN, WH, WIFI, WING, WMT, WOW, WSM, WTRH, WU, WWE, XOM, YUM, ZNGA, ZUMZ
\end{multicols}
\end{singlespace}
}

\section{Diagnostic Comparison of Scenario-3 Model Families}\label{wa:ablation_protocol}

This appendix describes the diagnostic comparison reported in the main text. It uses the selected scenario-3 results from three model families: independently estimated single-task models, a joint family without the revenue-alignment regularizer, and the full joint family with the regularizer. Because each family undergoes its own candidate search and selection, the comparison is not a same-hyperparameter causal ablation.

The single-task family estimates acquisition, ROPC, and AOV separately and constructs total sales compositionally from those selected forecasts. The joint family uses a shared representation without revenue alignment. The full family uses both the shared representation and the alignment objective described in Web Appendix~\ref{wa:training}. Each row in the reported comparison is the selected result from that family's scenario-3 candidate grid and selection strategy.

Accordingly, the reported differences are descriptive family-level contrasts. They summarize the performance of the selected implementations but do not isolate the causal effect of adding a shared representation or the revenue-alignment objective.

\section{Shared Customer-Base Drivers and Joint-Forecasting Gains}
\label{wa:why_mtl}

This appendix provides exploratory diagnostic evidence about when joint
forecasting helps. Acquisition, repeat purchasing, and spending can provide
multiple signals about the same business conditions. If the joint model
benefits from that shared structure, firms whose primitives co-move more
strongly during calibration should receive larger gains from joint forecasting
in the holdout period. The analysis does not identify the underlying business
shocks, establish a causal effect of co-movement, or isolate a single
mechanism.

\subsection{Measuring Cross-Metric Co-Movement}

For each firm, we construct weekly firm-level summaries of the three upstream
primitives during the calibration period:
\begin{align*}
A_t &= \text{new customers acquired in week } t, \\
R_t &= \text{repeat orders per cohort member in week } t
       \text{ (cohort-size-weighted)},\\
V_t &= \text{average order value in week } t
       \text{ (order-weighted)}.
\end{align*}
The calibration window spans January 1, 2017 through March 31, 2019. We apply a
$\log(1+x)$ transformation to each primitive and compute the co-movement
measures using the transformed weekly trajectories.

These firm-week summaries are used only as a diagnostic of cross-metric
co-movement. We do not assume that aggregate weekly revenue satisfies a
firm-week identity of the form \(S_t=A_tR_tV_t\). The model's revenue
construction is defined at the cohort-week level and aggregated by summing
cohort-week sales.

Our primary diagnostic is a \textit{Cross-Metric Co-Movement Score}, defined as
the first principal component of a broad battery of dependence measures. The
implementation generates 34 realized columns drawn from 28 conceptual families.
Before PCA, missing values are median-imputed, the decomposition-tightness
measure is winsorized at the 1st and 99th percentiles, and each input column is
standardized. The first component explains 41.1\% of the variance across the 34
columns. For interpretation, we express the resulting component score in
standard-deviation units in the analyses below. Table
\ref{tab:w1_comovement_measures} summarizes the measure families.

\begin{table}[!t]
\centering
\small
\caption{Cross-Metric Co-Movement Measures}
\label{tab:w1_comovement_measures}
\setlength{\tabcolsep}{3pt}
\begin{tabular}{
  p{0.15\linewidth}
  p{0.4\linewidth}
  p{0.35\linewidth}
}
\toprule
\textbf{Group} & \textbf{Measures} & \textbf{Interpretation} \\
\midrule

{\footnotesize \makecell[l]{Linear and rank \\ co-movement}} &
{\footnotesize (1) Pairwise Pearson correlation, (2) Spearman rank correlation, (3) partial correlation} &
{\footnotesize Do the primitives move together contemporaneously?} \\
\midrule

{\footnotesize \makecell[l]{Lead-lag \\ co-movement}} &
{\footnotesize (4) Max lead-lag correlation, (5) decay-weighted lag correlation, (6) acquisition-to-order/spend correlation} &
{\footnotesize Does movement in one primitive align with later movement in another?} \\
\midrule

{\footnotesize \makecell[l]{Cohort-level \\ co-movement}} &
{\footnotesize (7) Cohort-averaged directional dependence, (8) tenure-specific coupling} &
{\footnotesize Is co-movement visible in the cohort panel rather than only in firm-week aggregates?} \\
\midrule

{\footnotesize \makecell[l]{Nonlinear \\ co-movement}} &
{\footnotesize (9) Mutual information, (10) distance correlation, (11) kernel dependence, (12) tail co-exceedance} &
{\footnotesize Do the primitives exhibit nonlinear or co-extreme dependence?} \\
\midrule

{\footnotesize \makecell[l]{Predictive and \\ directed \\dependence}} &
{\footnotesize (13) Incremental predictive coupling, (14) Granger causality index, (15) transfer entropy, (16) forecast-error variance decomposition} &
{\footnotesize Does one primitive contain incremental forecasting information for another?} \\
\midrule

{\footnotesize \makecell[l]{Covariance and \\ latent-state \\ structure}} &
{\footnotesize (17) Shared covariance, (18) generalized variance reduction, (19) offsetting-error index, (20) first principal component share, (21) static factor share} &
{\footnotesize Are the primitives partly driven by a common factor?} \\
\midrule

{\footnotesize \makecell[l]{Regime, spectral, \\ and panel-aware\\ measures}} &
{\footnotesize (22) Spike coincidence, (23) spectral coherence, (24) rolling-window stability, (25) within-week cross-cohort order-spend coupling, (26) acquisition-dispersion link, (27) young-cohort correlation} &
{\footnotesize Do common shocks, periodicity, or cohort-panel patterns create shared movement?} \\
\midrule

{\footnotesize \makecell[l]{Decomposition \\ diagnostic}} &
{\footnotesize (28) Firm-week decomposition tightness} &
{\footnotesize How tightly a log-additive firm-week decomposition holds; diagnostic only} \\

\bottomrule
\end{tabular}

\begin{minipage}{0.95\linewidth}
\vspace{0.5em}
\footnotesize \textit{Note.} The main score uses calibration-period
\(\log(1+x)\)-transformed weekly primitives. Each of the 34 realized input
columns is standardized before PCA, and the resulting first-component score is
expressed in standard-deviation units for regression. 
\end{minipage}
\end{table}

\subsection{Specification and Main Results}

Our primary outcome is an indicator for whether the joint model improves
holdout sales forecasting relative to the corresponding single-task baseline:
\begin{align*}
\text{WinJoint}_i =
\mathbf{1}\left[
\text{SMAPE}_{\text{Joint},i} <
\text{SMAPE}_{\text{SingleTask},i}
\right].
\end{align*}
This indicator answers the practical question of whether joint forecasting
helps for firm \(i\). It is also less sensitive to extreme percentage errors
for firms with low sales levels.

For the main diagnostic, we test whether calibration-period co-movement
predicts holdout-period joint-model gains. The baseline linear-probability
specification is
\[
\text{WinJoint}_i =
\alpha+\beta\text{CoMovement}_i+\epsilon_i,
\]
where \(\text{CoMovement}_i\) is the Cross-Metric Co-Movement Score expressed
in standard-deviation units. We then add pre-holdout historical controls and
industry fixed effects:
\[
\begin{aligned}
\text{WinJoint}_i ={}&
\alpha+\beta\text{CoMovement}_i
+\gamma_1\text{HistoricalScale}_i \\
&+\gamma_2\text{HistoricalDispersion}_i
+\delta_{\text{Industry}(i)}+\epsilon_i.
\end{aligned}
\]
Historical scale is the log of one plus mean positive annual sales over
2016--2018. Historical dispersion is the coefficient of variation across those
annual sales totals. These controls are measured before the April 2019 holdout
begins.

We use five complementary tests: a two-sided comparison of the joint-model win
rate between the high and low co-movement terciles, the baseline
linear-probability model, the controlled linear-probability model, a controlled
logistic model reporting odds ratios, and a permutation placebo that shuffles
the firm-level co-movement scores 1,000 times and re-estimates the controlled
logit. Table \ref{tab:comovement_main_results} reports the results.

\begin{table}[!t]
\centering
\small
\caption{Cross-Metric Co-Movement and Joint-Model Performance}
\label{tab:comovement_main_results}
\setlength{\tabcolsep}{3pt}
\begin{tabular}{
  p{0.12\linewidth}
  p{0.18\linewidth}
  p{0.13\linewidth}
  p{0.15\linewidth}
  p{0.08\linewidth}
  p{0.22\linewidth}
}
\toprule
\textbf{Test} & \textbf{Specification} & \textbf{Estimate} & \textbf{Uncertainty} & \textbf{p-value} & \textbf{Interpretation} \\
\midrule

{\footnotesize \makecell[l]{Tercile \\ comparison}} &
{\footnotesize \makecell[l]{High vs. low \\ co-movement \\ tercile}} &
{\footnotesize +8.39 pp} &
{\footnotesize --} &
{\footnotesize .016} &
{\footnotesize High co-movement firms have a higher joint-model win rate} \\
\midrule

{\footnotesize \makecell[l]{Linear \\ probability}} &
{\footnotesize \makecell[l]{\(\text{WinJoint} \sim\) \\ \(\text{CoMovement}\)}} &
{\footnotesize \(\beta = 0.039\)} &
{\footnotesize SE = 0.014} &
{\footnotesize .006} &
{\footnotesize A one-SD increase predicts a 3.85 pp higher win probability} \\
\midrule

{\footnotesize \makecell[l]{Linear \\ probability \\ + controls}} &
{\footnotesize \makecell[l]{Historical controls \\ + industry FE}} &
{\footnotesize \(\beta = 0.050\)} &
{\footnotesize SE = 0.016} &
{\footnotesize .002} &
{\footnotesize A one-SD increase predicts a 5.03 pp higher win probability} \\
\midrule

{\footnotesize Logit} &
{\footnotesize \makecell[l]{Binary win model \\ with controls}} &
{\footnotesize OR = 1.310} &
{\footnotesize \makecell[l]{95\% CI \\ {[1.102, 1.557]}}} &
{\footnotesize .002} &
{\footnotesize A one-SD increase is associated with 31.0\% higher odds} \\
\midrule

{\footnotesize \makecell[l]{Permutation \\ placebo}} &
{\footnotesize \makecell[l]{1,000 score shuffles; \\ controlled logit}} &
{\footnotesize \makecell[l]{\(\beta = 0.270\) \\ \text{(obs.)}}} &
{\footnotesize \makecell[l]{SD = 0.078 \\ \text{(perm.)}}} &
{\footnotesize .000} &
{\footnotesize Observed association lies in the tail of the placebo
distribution} \\

\bottomrule
\end{tabular}

\begin{minipage}{0.95\linewidth}
\vspace{0.5em}
\footnotesize \textit{Note.} The dependent variable equals one if the joint
model has lower holdout SMAPE than the single-task baseline. Linear-probability
and logit models use HC1 standard errors. Controlled specifications include
pre-holdout annual-sales scale, cross-year annual-sales dispersion over
2016--2018, and industry fixed effects. The baseline model uses 966 firms. Five
firms lack the historical controls, so the controlled linear-probability model
uses 961 firms. The fixed-effects logit uses 943 firms because it additionally
excludes 18 firms in Electronics and Occasion \& Gifts, where every firm records a joint-model
win. The permutation placebo
randomly reassigns firm-level scores 1,000 times and re-estimates the controlled
logit.
\end{minipage}
\end{table}

Stronger cross-metric co-movement is associated with a higher probability that
joint forecasting outperforms the single-task baseline. The high-minus-low tercile difference is +8.39 percentage points. A one-standard-deviation increase in the co-movement score predicts a 5.03 percentage point increase in win probability in the controlled
linear-probability model and is associated with 31.0\% higher odds in the controlled logit.

\subsection{Additional Checks}
\label{wass:robustness-comovement}

The composite diagnostic is not driven by a single specialized dependence
measure. Across the 34 realized columns, 26 have positive associations with
joint-model win probability. Ten survive Benjamini-Hochberg
false-discovery-rate correction at \(q<.10\): nine positive columns and
transfer entropy, a directed-dependence measure, with a negative coefficient
(logit coefficient \(-0.43\), \(p=.001\), adjusted \(q=.041\)). The leading
positive results include contemporaneous and rank correlations, short-lag
co-movement, mutual information, distance correlation, and common-factor
share. Complete individual-column results are provided with the replication
output.

Taken together, these binary-outcome checks support a narrow interpretation: cross-metric co-movement is associated with where the joint model performs better. Because this is an exploratory diagnostic, the evidence does not by itself identify the underlying shared drivers or establish a causal mechanism.

\section{Model Selection for Deployment}\label{wa:deployment}

This appendix reports the model-selection evaluation described in the main text. The question is whether calibration-period observables can identify firms for which the single-task baseline forecasts total sales more accurately than the joint model.

The analysis has two distinct holdouts. First, CBMT and the single-task baseline generate forecasts for the April 2019 through February 2020 forecasting holdout. For each firm, these forecasts produce two total-sales SMAPE values; the single-task forecast is constructed by combining its separately forecast acquisition, ROPC, and AOV paths. We set $\text{WinJoint}_i = 1$ when CBMT has the lower forecasting-holdout SMAPE. Second, we create a cross-firm holdout by splitting the 966 firms into policy-training and policy-evaluation samples. Forecasting-holdout errors for the policy-training firms supply the routing labels and policy loss, whereas the corresponding errors for cross-firm held-out firms are not used until final policy evaluation.

Every predictor is constructed using data available no later than March 31, 2019. The logistic regression uses a Cross-Metric Co-Movement Score, log total orders over the calibration window, and cohort-sales volatility, measured as the cohort-size-weighted average of within-cohort weekly sales coefficients of variation. The depth-three decision tree and gradient-boosted classifier additionally use the 34 individual dependence measures underlying the co-movement score.

All learned preprocessing is nested within the cross-firm evaluation. In each inner cross-validation fold, missing dependence values are replaced using medians estimated from that fold's fitting firms, the 34 dependence measures are standardized using those firms, and the first principal component is estimated from the standardized fitting-firm data. The resulting transformation is then applied to the validation firms. The three inputs to the logistic regression are standardized within the same fitting fold. For the final outer-split model, these steps are re-estimated using all policy-training firms and then applied to the cross-firm held-out firms. Thus, neither the co-movement score nor any other learned transformation uses the cross-firm held-out data. 

For the primary evaluation, we split firms 70/30 into 676 training firms and 290 held-out firms, stratified by industry. Industries with fewer than 10 firms are pooled for stratification. The candidate-rule specifications and hyperparameters are held fixed across splits and are not selected using the outer held-out firms. Within the training sample, five-fold cross-validation produces out-of-fold predictions for each candidate rule, with preprocessing re-estimated in every fold.

The decision threshold minimizes the mean total-sales SMAPE that the resulting policy would have produced among the policy-training firms. The preprocessor and model are then refit on all policy-training firms and applied once to the cross-firm held-out firms. We benchmark each rule against always using single-task forecasts, always using CBMT, and a two-model oracle that selects the lower-SMAPE model for each firm with perfect foresight.

Table~\ref{tab:deployment_baserates} reports the base rates over all firms. Always-CBMT gives a mean total-sales SMAPE of 15.48 and always-single-task 20.54, while the two-model oracle gives 14.79. The 0.69-point gap between always-CBMT and the oracle bounds the gain available to a rule restricted to choosing between these two models under mean firm-level total-sales SMAPE. CBMT wins for 718 of the 966 firms (74.3\%), with a mean gain of 7.74 SMAPE points, and loses for the remaining 248 firms, with a mean shortfall of 2.69 points.

\begin{table}[!htbp]
\centering
\small
\caption{Base-Rate Policy Comparison Over All Firms}
\label{tab:deployment_baserates}
\begin{tabular}{lcc}
\hline
\textbf{Policy} & \textbf{Mean total-sales SMAPE} & \textbf{Firms to CBMT / single-task} \\
\hline
Always single-task & 20.54 & 0 / 966 \\
Always CBMT & 15.48 & 966 / 0 \\
Oracle (per-firm best) & 14.79 & 718 / 248 \\
\hline
\end{tabular}
\begin{minipage}{0.9\linewidth}
\vspace{0.5em}
\footnotesize \textit{Note.} Quantities are computed over all 966 firms. The oracle is restricted to choosing between CBMT and the compositional single-task model. It assigns each firm to the model with the lower holdout total-sales SMAPE and therefore sends to single-task models only the 248 firms where single-task forecasting wins.
\end{minipage}
\end{table}

On the primary split, the logistic rule routes 2 of the 290 held-out firms to single-task forecasting and attains mean SMAPE 17.171, compared with 17.189 for always-CBMT. The decision tree and gradient-boosted classifier route no firms to single-task and therefore reproduce the always-CBMT result. The oracle routes 82 firms to single-task and reaches 16.585.

Table~\ref{tab:deployment_resplit} shows the results across 100 re-randomized 70/30 splits. The logistic regression makes 24 single-task assignments across 11 splits, improving on always-CBMT in 2 splits and worsening it in 9. The decision tree makes 34 single-task assignments in one split and is worse in that split. The gradient-boosted classifier makes 66 single-task assignments across 22 splits, improving in 4 and worsening in 18. Averaged across all 100 splits, always-CBMT has mean SMAPE 15.441, compared with 15.448, 15.444, and 15.457 for the logistic regression, decision tree, and gradient-boosted classifier, respectively. Thus, none of the fitted routing rules improves on always-CBMT on average.

\begin{table}[!htbp]
\centering
\small
\caption{Out-of-Sample Policy Comparison Across 100 Re-Randomized Splits}
\label{tab:deployment_resplit}
\begin{tabular}{lcc}
\hline
\textbf{Policy} & \textbf{Mean holdout SMAPE} & \textbf{SD} \\
\hline
Always single-task & 20.486 & 1.251 \\
Always CBMT & 15.441 & 0.958 \\
Logistic regression & 15.448 & 0.959 \\
Decision tree & 15.444 & 0.958 \\
Gradient-boosted classifier & 15.457 & 0.961 \\
Oracle (two-model choice) & 14.738 & 0.925 \\
\hline
\end{tabular}
\begin{minipage}{0.9\linewidth}
\vspace{0.5em}
\footnotesize \textit{Note.} Mean and standard deviation of held-out total-sales SMAPE across 100 industry-stratified 70/30 re-splits. In every inner and outer split, imputation, standardization, principal-component estimation, model fitting, and threshold selection use only the corresponding fitting firms. The oracle is restricted to the CBMT-versus-single-task choice.
\end{minipage}
\end{table}

The deployment conclusion is correspondingly narrow. The tested calibration-period observables do not support a policy that improves average cross-firm held-out total-sales accuracy over always deploying CBMT. This does not mean the single-task model never wins: it has lower realized forecasting-holdout SMAPE for 248 firms. Rather, the fitted rules do not identify those firms reliably enough to improve on the 74.3\% always-CBMT base rate on average. The analysis is restricted to choosing between CBMT and its compositional single-task counterpart under mean firm-level total-sales SMAPE; it does not evaluate routing to LSTM or another challenger model.

\section{Selected Studies in Related Literature}\label{wa:literature_comparison}

Table~\ref{tab:literature_comparison} provides a descriptive comparison of selected studies discussed in the related-literature section. The table is included as a reference aid rather than as a feature-checklist claim about novelty.

\begin{table}[!htbp]
\centering
\caption{Selected Studies in Related Literature}
\label{tab:literature_comparison}
\footnotesize
\setlength{\tabcolsep}{4pt}
\renewcommand{\arraystretch}{1.15}
\begin{tabular*}{\textwidth}{@{\extracolsep{\fill}}>{\raggedright\arraybackslash}p{4.5cm}>{\raggedright\arraybackslash}p{2.3cm}>{\raggedright\arraybackslash}p{3.0cm}>{\raggedright\arraybackslash}p{1.6cm}>{\raggedright\arraybackslash}p{1.4cm}@{}}
\toprule
Study & Modeling Approach & Behaviors Modeled & Empirical Context & Up/ Downstream \\
\midrule
\multicolumn{5}{l}{\textit{Customer-to-Firm Valuation Literature}} \\
\citet{blattberg1996manage} & Strategic & Adoption, churn & Conceptual & No \\
\citet{rust2004return} & Strategic & Adoption, churn, spend & 4 firms & No \\
\citet{gupta2004valuing} & Parametric & Adoption, churn, spend & 5 firms & No \\
\citet{schulze2012linking} & Parametric & Adoption, churn, spend & 2 firms & No \\
\citet{mccarthy2017valuing} & Parametric & Adoption, churn, spend & 2 firms & No \\
\citet{mccarthy2018customer} & Parametric & Adoption, purchase, spend & 2 firms & No \\
\midrule
\multicolumn{5}{l}{\textit{Customer Acquisition / New-Product Adoption}} \\
\citet{bass1969new} & Parametric & Adoption & 11 products & No \\
\citet{norton1987diffusion} & Parametric & Adoption, substitution & 3 products & No \\
\citet{bass1994gbm} & Parametric & Adoption & 11 products & No \\
\midrule
\multicolumn{5}{l}{\textit{Purchase (Noncontractual)}} \\
\citet{schmittlein1987counting} & Parametric & Purchase & Conceptual & No \\
\citet{fader2005rfm} & Parametric & Purchase & 1 cohort & No \\
\citet{schweidel2013incorporating} & Parametric & Purchase & 1 cohort & No \\
\citet{braun2015transaction} & Parametric & Purchase & 4 cohorts & No \\
\citet{platzer2016ticking} & Parametric & Purchase & 6 cohorts & No \\
\citet{dew2018bayesian} & Nonparametric & Purchase & 2 cohorts & No \\
\citet{bachmann2021role} & Parametric & Purchase & 3 cohorts & No \\
\citet{valendin2022customer} & Nonparametric & Purchase & 8 cohorts & No \\
\midrule
\multicolumn{5}{l}{\textit{Spending}} \\
\citet{fader2005rfm} & Parametric & Spend & 1 cohort & No \\
\midrule
\multicolumn{5}{l}{\textit{Joint Modeling}} \\
\citet{schweidel2008bivariate} & Parametric & Adoption, churn & 1 firm & No \\
\citet{netzer2008hidden} & Parametric & Purchase, spend & 1 firm & No \\
\citet{singh2009bayesian} & Parametric & Purchase, spend & 1 cohort & No \\
\citet{ascarza2013joint} & Parametric & Usage, churn & 1 cohort & No \\
\midrule
Our Paper (CBMT) & Nonparametric & Adoption, purchase, spend & 966 firms & Yes \\
\bottomrule
\end{tabular*}
\begin{minipage}{0.98\linewidth}\vspace{2pt}\footnotesize
\textit{Note.} `Up/Downstream' indicates whether the study explicitly incorporates both upstream customer behaviors and downstream revenue outcomes in estimation. In \citet{schweidel2008bivariate} and \citet{netzer2008hidden}, the authors estimate models across adoption/graduation cohorts with cross-cohort differences captured through covariates and heterogeneity terms. \citet{braun2015transaction} model four distinct acquisition cohorts with cohort-specific covariates to account for cross-cohort differences.
\end{minipage}
\end{table}

\clearpage
\newpage
\printbibliography[
title={REFERENCES}
]







\WAstoptoc

\end{document}